\documentclass[%
 aip,
 jmp,
 amsmath,amssymb,
 preprint,
]{revtex4-1}

\usepackage{graphicx}
\usepackage{dcolumn}
\usepackage{bm}
\usepackage[off]{auto-pst-pdf}
\usepackage{psfrag}
\usepackage{bm}
\usepackage{subcaption}
\usepackage{tabularx}
\usepackage{array}
\newcolumntype{x}[1]{>{\centering\arraybackslash\hspace{0pt}}p{#1}}
\newcolumntype{Y}{>{\centering\arraybackslash}X}

\makeatletter
\renewcommand\@makecaption[2]{%
  \par
  \vskip\abovecaptionskip
  \begingroup
   \small\rmfamily
    \begingroup
     \samepage
     \flushing
     \let\footnote\@footnotemark@gobble
     \@make@capt@title{#1}{#2}\par
    \endgroup
  \endgroup
  \vskip\belowcaptionskip
}
\makeatother

\newcommand*{\citen}[1]{%
  \begingroup
    \romannumeral-`\x 
    \setcitestyle{numbers}%
    \cite{#1}%
  \endgroup
}

\begin{document}

\title[]{Evolution of $K$-means solution landscapes with the addition of dataset outliers and a robust clustering comparison measure for their analysis}

\author{L. Dicks}
\affiliation{ 
Yusuf Hamied Department of Chemistry, Lensfield Road, Cambridge CB2 1EW, United Kingdom
}
\affiliation{
IBM Research Europe, Hartree Centre, Sci-Tech Daresbury, United Kingdom
}
\author{D. J. Wales}%
 \email{dw34@cam.ac.uk}
\affiliation{ 
Yusuf Hamied Department of Chemistry, Lensfield Road, Cambridge CB2 1EW, United Kingdom
}

\date{\today}

\begin{abstract}
The $K$-means algorithm remains one of the most widely-used clustering methods
due to its simplicity and general utility. The performance of $K$-means depends
upon location of minima low in cost function, amongst a potentially vast number
of solutions. Here, we use the energy landscape approach to map the change in
$K$-means solution space as a result of increasing dataset outliers and show
that the cost function surface becomes more funnelled. Kinetic analysis reveals
that in all cases the overall funnel is composed of shallow locally-funnelled
regions, each of which are separated by areas that do not support any
clustering solutions. These shallow regions correspond to different types of
clustering solution and their increasing number with outliers leads to longer
pathways within the funnel and a reduced correlation between accuracy and cost
function. Finally, we propose that the rates obtained from kinetic analysis
provide a novel measure of clustering similarity that incorporates information
about the paths between them. This measure is robust to outliers and we
illustrate the application to datasets containing multiple outliers.
\end{abstract}


\maketitle

\section{\label{sec:level1} Introduction}

Clustering is ubiquitous in data analysis and, consequently, many diverse
clustering methods have been developed. Amongst the most popular is the hard,
partitional $K$-means algorithm.\cite{Lloyd1982} The popularity of $K$-means
stems from its efficiency, algorithmic simplicity, and its ability to produce
clusterings in agreement with more complex and expensive
algorithms.\cite{Jain2004, deSouto2008} $K$-means clusters a given dataset into
a prespecified number of clusters, $K$. For some given initial cluster
positions, $K$-means produces a valid clustering solution by
minimising the sum-of-squares cost function until a local minimum is attained.
The cost function surface is non-convex and can support many local minima,
which correspond to clusterings of varying quality. The number of local minima
for the $K$-means cost function can be vast, and enumeration is impossible for
all but trivial cases.\cite{Mahajan2012}

Many schemes for initialising cluster centres have been proposed, which are
usually repeatedly applied and, after minimisation of each initial position,
the best clustering solution is retained. Many popular initialisation schemes
use uniform random samples of the data.\cite{Steinley2003, Bradley1998,
Faber1994} Such methods have a low computational cost, allowing many
uncorrelated initial cluster positions to be generated. More complex
initialisation schemes draw non-uniform data samples,\cite{Arthur2007,
Mohammad2009, Bachem2016} or pair with other algorithms, such as hierarchical
clustering,\cite{Milligan1980} or principal component analysis.\cite{Su2004}
These schemes generally have a greater computational cost, but produce local
minima of higher quality, allowing fewer distinct initialisations to be
considered. Additionally, it is possible to use known solutions to guide future
searching, rather than generating uncorrelated trials, and $K$-means has been
paired with genetic algorithms,\cite{Krishna1999, Franti2000, Marghny2014}
Monte Carlo-style schemes,\cite{Hand2005} and various other optimisation
algorithms.\cite{Hatamlou2012, Yao2016, Das2018, Manju2018, Xie2019}

The relative performance of all the above methods depends heavily on the dataset
structure and the number of clusters.\cite{Steinley2007, Filippone2008,
Rai2010, Celebi2013} The effect of dataset structure on $K$-means performance
is encoded by the cost function topography, as the ability of initialisation
schemes to locate low-valued $K$-means minima depends upon the organisation of
the solution space. The cost function can be coarse-grained
in terms of the local minima and the transition states that connect them, as in
the energy landscape framework.\cite{Wales03} The structure, thermodynamics, and kinetic properties of physical systems can be described from
these stationary points, and we have recently extended this approach to explore the structure of the $K$-means cost function.\cite{Dicks2022, Wu2023} 

The dependence of clustering accuracy on dataset structure means that datasets
are frequently modified prior to clustering to change the structure to favour
higher accuracy.\cite{Luai2006} This preprocessing commonly includes removal of
outliers, defined as data points that differ significantly from the rest, which
can badly degrade the quality of clustering that is
achievable.\cite{GarciaEscudero1999, Zhang2003} Hence care is often taken to
identify, and remove, outliers prior to clustering. However, there are various
methods to make $K$-means more robust to outliers.\cite{Mohammad2009, Xu2016,
Gupta2017, Huang2018, Brodinova2019, Zhang2020} Clustering itself can reveal
outliers, and there are many examples of outlier detection during
clustering.\cite{AlZoubi2009, Jiang2001, Marghny2014, Zhang2021} The effect of
outliers on clustering accuracy is well studied, but the effect they have on
the organisation of the solution space is unknown, and this is the question we address
in this contribution.

We first describe the algorithms of the energy landscape
approach, and the extensions required to explore cost function surfaces. Using
this framework we analyse the change in the $K$-means solution space that
results from the presence of dataset outliers, and 
for two datasets we find that the landscape becomes more funnelled.
Subsequently, we determine the properties of clustering
solutions using kinetic analysis to further examine the nature of the
funnelling. Lastly, we propose the use of metrics based upon rates as a robust cluster
comparison measure that will be especially useful in describing global
optimisation properties.

\section{Methods}

\subsection{Local minima}

The $K$-means algorithm aims to minimise the sum-of-squares cost function,
\begin{equation} \label{cost}
J(\bm{\mu}) = \sum\limits^{N}_{i=1} \sum\limits^{K}_{k=1} r_{ik} \| \mathbf{x}_i - \mathbf{\bm{\mu}}_{k} \|^2.
\end{equation}
$\mathbf{x}$ is an ($N_{\mathrm{f}} \times N$)-dimensional matrix that contains
the $N$ data points, each of $N_{\mathrm{f}}$ features. $\bm{\mu}$ contains the
coordinates of all $K$ clusters. Each individual data point and cluster are
denoted by the $N_\mathrm{f}$-dimensional vectors $\mathbf{x}_i$ and
$\bm{\mu}_{k}$, respectively. For any given $\bm{\mu}$, data points are
assigned to the nearest cluster centre in Euclidean distance according to
\begin{equation}
r_{ik} = \begin{cases}
1, & \text{if } k = \displaystyle\min\limits_j\| \mathbf{x}_i - \bm{\mu}_{j} \|^2,\\
0, & \text{if otherwise.}
\end{cases}
\end{equation}
$r_{ik}$ is the element of the cluster assignment matrix $\mathbf{r}$ that denotes the assignment of data point $i$ to cluster $k$.

The cost function, from any given cluster positions, can be locally minimised
using Lloyd's algorithm.\cite{Lloyd1982} Initial cluster coordinates were
generated using uniform random points drawn within the range of the dataset in
each feature. This inefficient initialisation scheme was chosen to capture both
poor and good clustering solutions, as we aim to explore the complete solution
space. $10^6$ initial cluster positions
were generated, and subsequently minimised. Repeated minima were removed, along
with those containing empty clusters.

There are a variety of metrics for evaluating the clusterings generated by
$K$-means. Without known data point assignments, as is common in exploratory
data analysis, internal metrics can assess clustering quality. Internal metrics
use properties of the resulting clusters, such as their compactness and
separation. The central internal metric within $K$-means is the cost function,
which gives the within-cluster variance. However, various other internal
metrics have been proposed.\cite{Dunn1974, Davies1979, Rousseeuw1987}

In contrast, external metrics measure the similarity of a clustering to a given
reference. For labelled datasets we refer to the known partitioning as the
ground truth clustering. These ground truth labels are a common reference, in
which case we measure the clustering accuracy. Popular external metrics can be
based upon cluster matchings,\cite{Rand1971, Hubert1985, Fowlkes1983}
information theory,\cite{Meila2007} or meta-clusters.\cite{Cazals2019} We use
the adjusted Rand index (ARI)\cite{Rand1971, Hubert1985} throughout this work,
which is described in detail in the SI. The adjusted Rand index ranges from $-1$
to 1, with a value of 1 indicating perfect agreement of two clusterings, and 0
indicating the similarity expected between two random clusterings. Values less
than 0 indicate a lower similarity than expected for random clusterings.

\subsection{Transition states}

For a detailed description of the methodology we employ to explore the
$K$-means cost function topography we refer readers to Ref.~\citen{Dicks2022}.
A brief summary is presented here.

Transition states provide the lowest barrier in the intermediate region of a
surface between one minimum and another and, therefore, provide valuable
information about the organisation. There are no transition
states, defined as Hessian index one saddles,\cite{Murrell1968} on $K$-means
cost function surfaces.\cite{Dicks2022} However, minimum-energy crossing points
(MECPs), which lie at the minimum of an intersection seam between two
quadratics defined by different cluster assignments, can be used as  transition
states in the present context. At all points on a seam the Hessian is
undefined, but an MECP can have the correct curvature as a maximum in the
direction orthogonal to the intersection seam, and a minimum with respect to
all directions along the intersection seam.

Most transition state location algorithms\cite{Munro1999, Henkelman2000,
Peters2004} require a function with smooth derivatives, which is not the case
for $K$-means. Instead, we employ a penalty-constrained MECP optimisation
algorithm.\cite{Levine2008} Location of an MECP between a pair of cluster
assignments, $\bm{r}^{(1)}$ and $\bm{r}^{(2)}$, which differ in a single data
point assignment, can be achieved through minimisation of a surrogate function,
\begin{equation} \label{eq:1}
F_+(\bm{\mu}, \bm{r}^{(1)}, \bm{r}^{(2)}, \sigma, \alpha) = \frac{1}{2}\Big( J^{(1)}+J^{(2)}\Big) + \sigma \left( \frac{\Big(J^{(1)}- J^{(2)} \Big)^2}{\|J^{(1)}-J^{(2)} \| + \alpha} \right).
\end{equation}
$J^{(j)} = J(\bm{\mu}, \bm{r}^{(j)})$ is the $K$-means cost function, evaluated
at cluster position $\bm{\mu}$, assuming a fixed cluster assignment
$\bm{r}^{(j)}$. $\sigma$ and $\alpha$ are parameters that determine the
increase in cost function away from the intersection seam, and smooth the
surface to avoid a discontinuity there, respectively. $\alpha=0.02$ and
$\sigma=30$ were used throughout this work, which produced a difference between
the cost function evaluated under cluster assignments $\bm{r}^{(1)}$ and
$\bm{r}^{(2)}$ of $\mathcal{O}(10^{-4})$. This difference was sufficiently
small for our application, and permitted $\mathcal{O}(10^{5})$ transition state
location attempts, but the intersection seam can be found to progressively
higher accuracy by reminimising $F_+$ with small increments applied to
$\sigma$. Minimisations were performed using a customised L-BFGS algorithm.\cite{Nocedal1980, Liu1989}

The minimum of $F_+$, our approximation to the MECP, was then evaluated on $J$
to give the cost function at the transition state between clusterings
$\bm{r}^{(1)}$ and $\bm{r}^{(2)}$. Transition states have two connected minima,
found by following (approximate) steepest-descent paths beginning in the
forwards and backwards direction of the eigenvector corresponding to the unique
negative Hessian eigenvalue. For $K$-means transition states the required
eigenvector must be determined from an alternative function. Here we use
Eq.~(\ref{eq:1}) with subtraction, rather than addition, of the second term.
The function, denoted $F_-$, penalises the two quadratics when they are similar in
value and has the desired negative curvature in the direction perpendicular to
the intersection seam. We evaluate the Hessian on the $F_-$ surface, at the
minimum of $F_+$, to find the negative eigenvalue and its corresponding
eigenvector.

\subsection{Building a stationary point network}

We have summarised the methodology for finding transition states between
neighbouring minima, but to connect a large number of minima we need a scheme
to select appropriate pairs for transition state searches. Pairs are selected using a distance-based criterion,\cite{RoederW18a} which finds the lowest-valued minimum not currently
connected to the global minimum by any sequence of transition states and connected
minima, and selects the nearest clustering solution in Euclidean distance that
is connected to the global minimum.

For each chosen pair of minima, we do not directly attempt a transition state
search, as they may be distant in cluster assignment. Instead, we construct a
linear path composed of discrete, equally-spaced cluster positions, referred to
as images. The number of images is adapted such that the cluster assignment
between any two adjacent images differs by only a single data point, and at
such a change in cluster assignment, a transition state search is attempted.
$F_+$ is constructed using the differing cluster assignments of the two images,
and a minimisation is started from the first image. If a valid MECP is found we
compute the Hessian on $F_-$ and identify the connected minima by LBFGS
minimisation. The process is repeated for all differing pairs of images, and
the valid transition states are added to a stationary point database, along
with their connectivity.

The aim of successive transition state searches is to produce a fully connected
set of minima, where it is possible to move between any pair via a series of
transition states and connected minima (a discrete path\cite{Wales2002,Wales2004}). The repeated application of the
distance-based criterion metric to select minima leads to efficient growth of
the global minimum connections until all minima are connected. The resulting
database of minima and transition states can be represented as a weighted
graph, in which minima are nodes, and all pairs of minima connected by a
transition state are joined by an edge weighted by a function of the barrier
height. For physical systems, this database corresponds to a kinetic transition
network,\cite{Noe2008,Carr2009,Prada2009} which we visualise using
disconnectivity graphs.\cite{Becker1997, Wales1998} Disconnectivity graphs
illustrate the minima and the barriers separating them for functions in
arbitrary dimensions, as described in Sec.~SI.

We construct the analogue of a kinetic transition network for the $K$-means
cost function surface to calculate transition rates between groups of minima.
Such rates have no physical analogue, unlike molecular systems, but they
nevertheless provide a useful estimate of the difficulty in moving between
different regions of cluster solution space. The rates account for all
the intervening barriers in the cost function and all the possible pathways
via any number of intervening local minima.
These properties are important in understanding the $K$-means
solution space, and contain important additional information that is not present in
distance calculations of either cluster positions or assignments.

We compute the rate constant between two minima that are directly connected by
a transition state, using the analogue of unimolecular rate
theory.\cite{Eyring1935,Evans1935} The vibrational density of states that
appears in this elementary rate theory is set to unity for our $K$-means applications. The
description of global kinetics for a complete network is based upon the
combination of the rates of these individual steps, and the graph
transformation algorithm\cite{TrygubenkoW06b, Wales2009} was used to calculate
all the overall rates reported in this work. We also extract the fastest path, when
intervening minima are considered in steady state, using Dijkstra's
algorithm\cite{Dijkstra1959} with suitable edge weights.\cite{Wales2002}

The temperature parameter strongly affects the observed pathways and rates.
Higher temperatures lead to faster rates, and pathways that contain larger
barriers. Each cost function landscape was analysed at a fictitious temperature
that produces observed rates in the range between $\mathcal{O}(10^{-6}) -
\mathcal{O}(10^{-3})$, which ensures that we favour pathways with small
barriers, and see a clearer separation in rates.
The cost function landscape itself is independent of temperature.

\subsection{Datasets} \label{DataPreparation}

We construct $K$-means landscapes for variations of two labelled datasets:
Fisher's \textit{Iris} dataset\cite{Fisher1936} and the glass identification
dataset,\cite{Evett1989} both taken from the UCI machine learning data
repository.\cite{Dua2019} Fisher's \textit{Iris} dataset contains 150 data
points of four features each. There are fifty measurements of each of the three
\textit{Iris} species: \textit{Virginica}, \textit{Versicolor} and
\textit{Setosa}. The \textit{Setosa} data points form a clearly-separable
cluster, but measurements of the \textit{Versicolor} and \textit{Virginica}
species overlap, and cannot be separated by the $K$-means algorithm.

The glass dataset contains 214 data points separated into six clusters of
uneven populations. The dataset has nine features, which exhibit uneven ranges
that vary by two orders of magnitude. The ground truth clusters have
significant overlap, and the accuracy achievable by $K$-means is limited for
this dataset.

Outliers were sequentially added to both datasets, four to the \textit{Iris}
dataset and three to the glass identification dataset. The addition of each
successive outlier produced a distinct dataset, giving four \textit{Iris}
datasets with an increasing number of outliers. Outliers were placed distant
from both the original data points and any existing outliers. The distance
between each outlier and the mean of the overall data was not constrained, but
is similar for all outliers. The coordinates were selected at random, and are
given in Tables~SI and SII.

\section{Results}

Outliers modify $K$-means clustering to produce two alternative types of solutions:
\begin{enumerate}
\item{The outlier belongs to a cluster containing no data points from the original dataset.}

There remain only $K-O$ clusters to represent the original dataset, where $O$
is the number of outlier clusters. $K$-means performance is degraded due to
poorer partitioning of the original data, and the degradation is greater when
the original dataset is composed of $K$ well-separated clusters with little
overlap.
      
\item{The outlier belongs to a cluster containing some original data points.}

If an outlier does not belong to its own cluster then it must be accommodated
in a cluster containing original data points. The outlier may skew the
cluster centre a significant distance away from the mean of the original data
points, due to its remote position.

\end{enumerate}

\begin{figure}
    \centering
    \begin{subfigure}[t]{0.25\textwidth}
        \centering
        \includegraphics[width=0.8\textwidth]{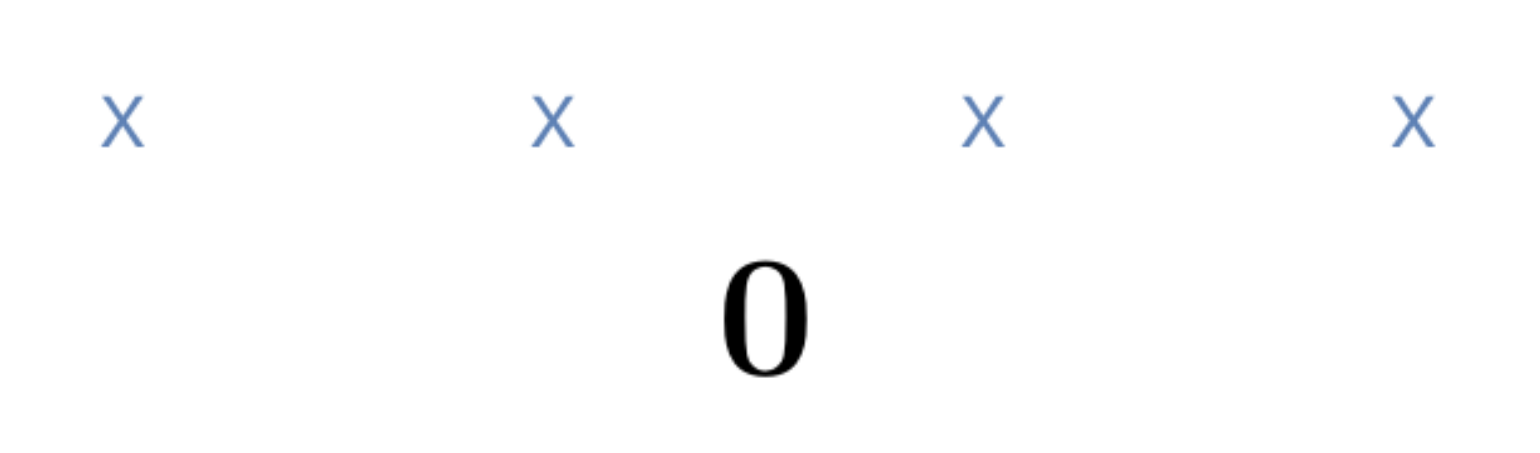}
    \end{subfigure}%
    \begin{subfigure}[t]{0.25\textwidth}
        \centering
        \includegraphics[width=0.8\textwidth]{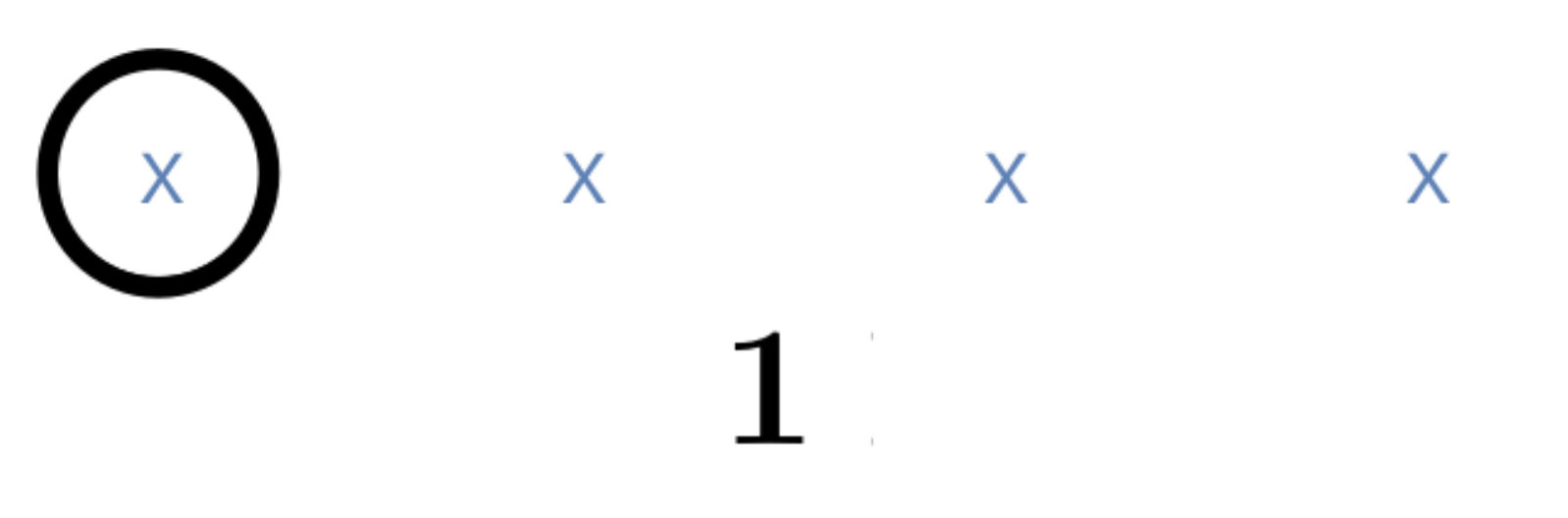}
    \end{subfigure}%
    \begin{subfigure}[t]{0.25\textwidth}
        \centering
        \includegraphics[width=0.8\textwidth]{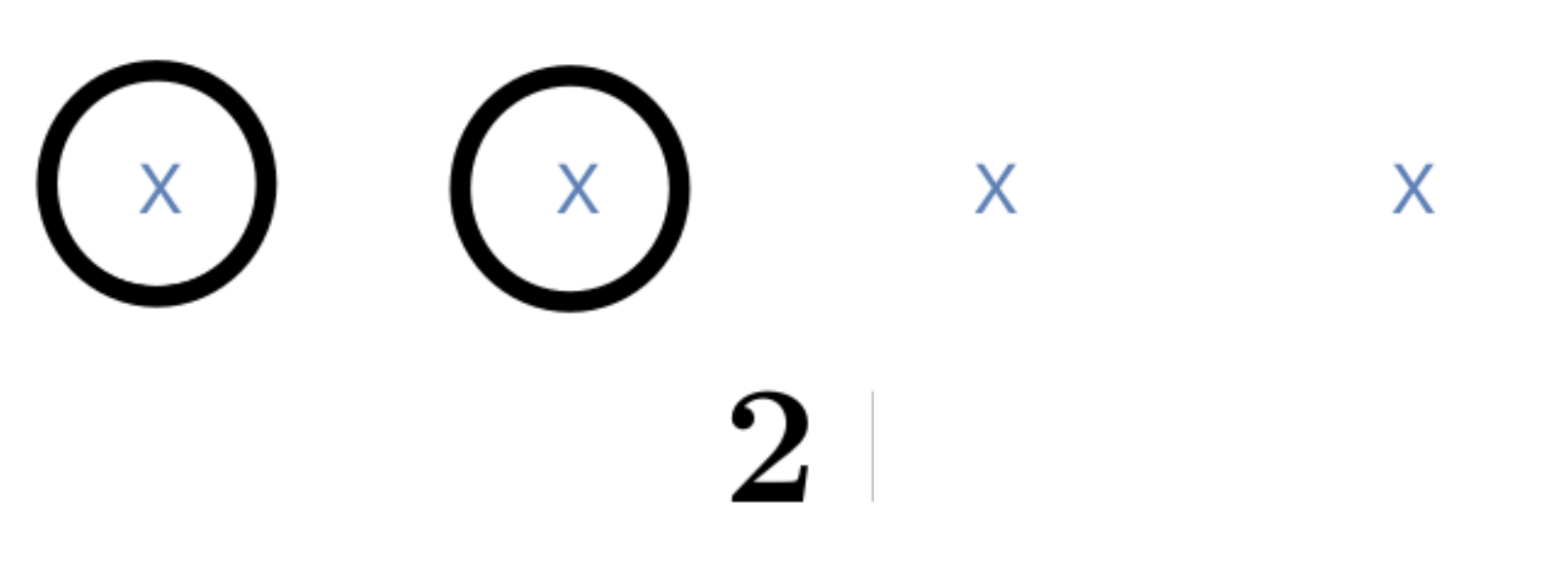}
    \end{subfigure}%
    \begin{subfigure}[t]{0.25\textwidth}
        \centering
        \includegraphics[width=0.8\textwidth]{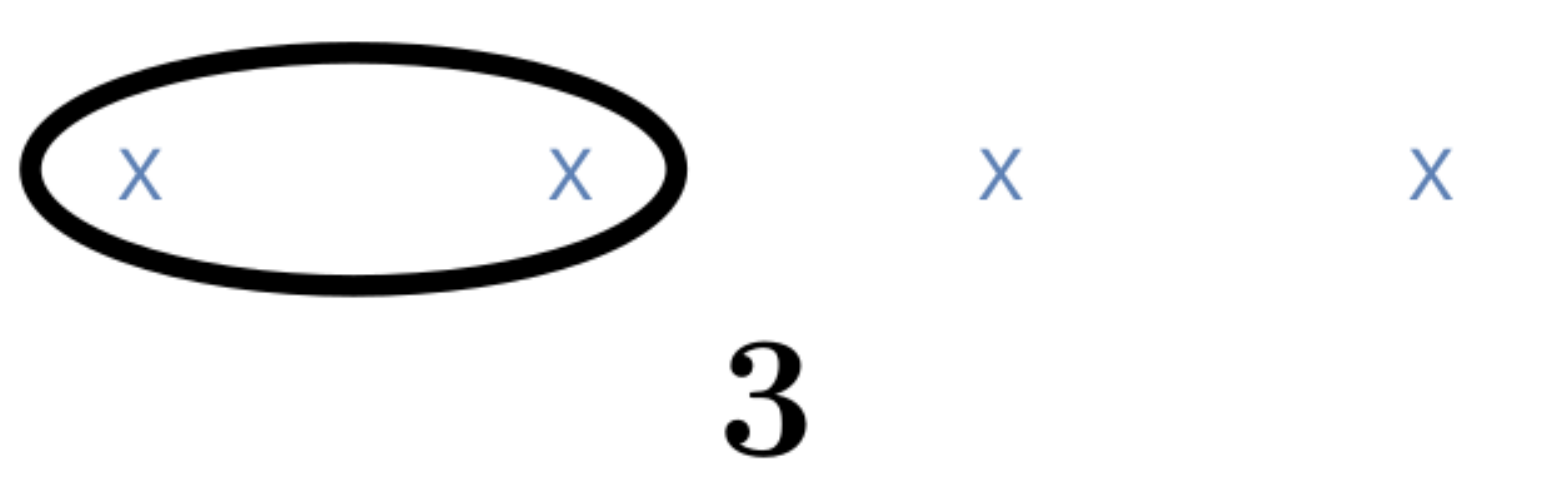}
    \end{subfigure} \\
    \vspace*{0.3cm}
    \begin{subfigure}[t]{0.25\textwidth}
        \centering
        \includegraphics[width=0.8\textwidth]{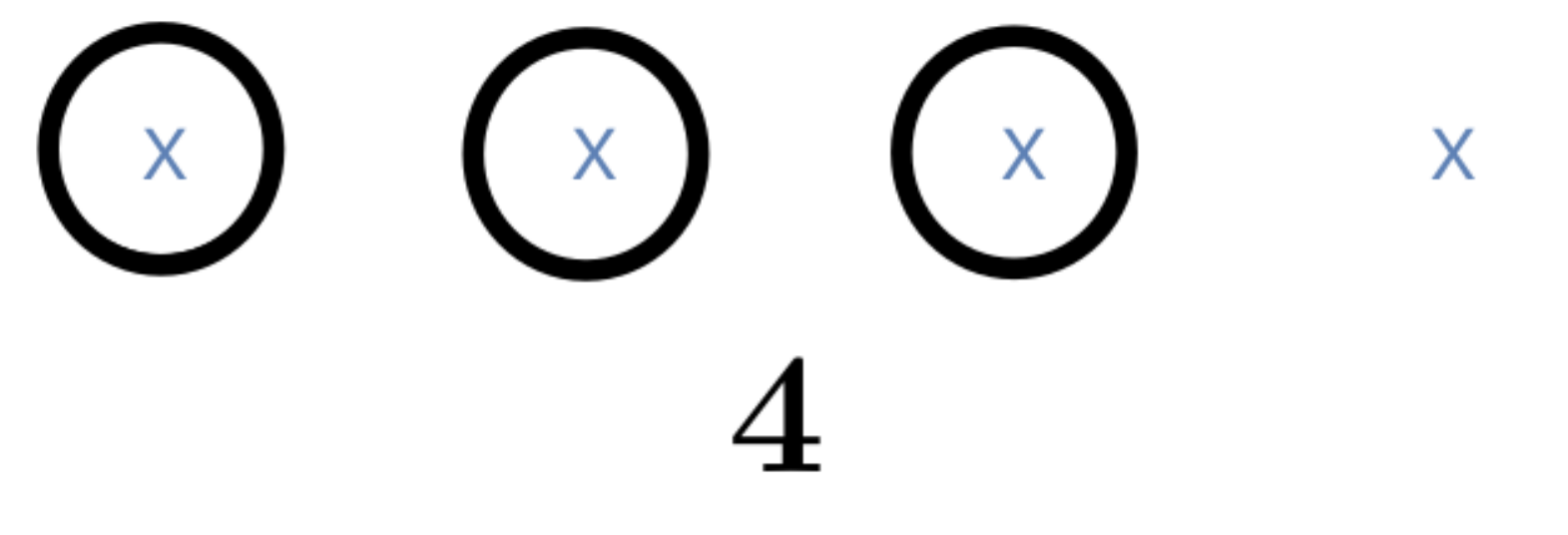}
    \end{subfigure}%
    \begin{subfigure}[t]{0.25\textwidth}
        \centering
        \includegraphics[width=0.8\textwidth]{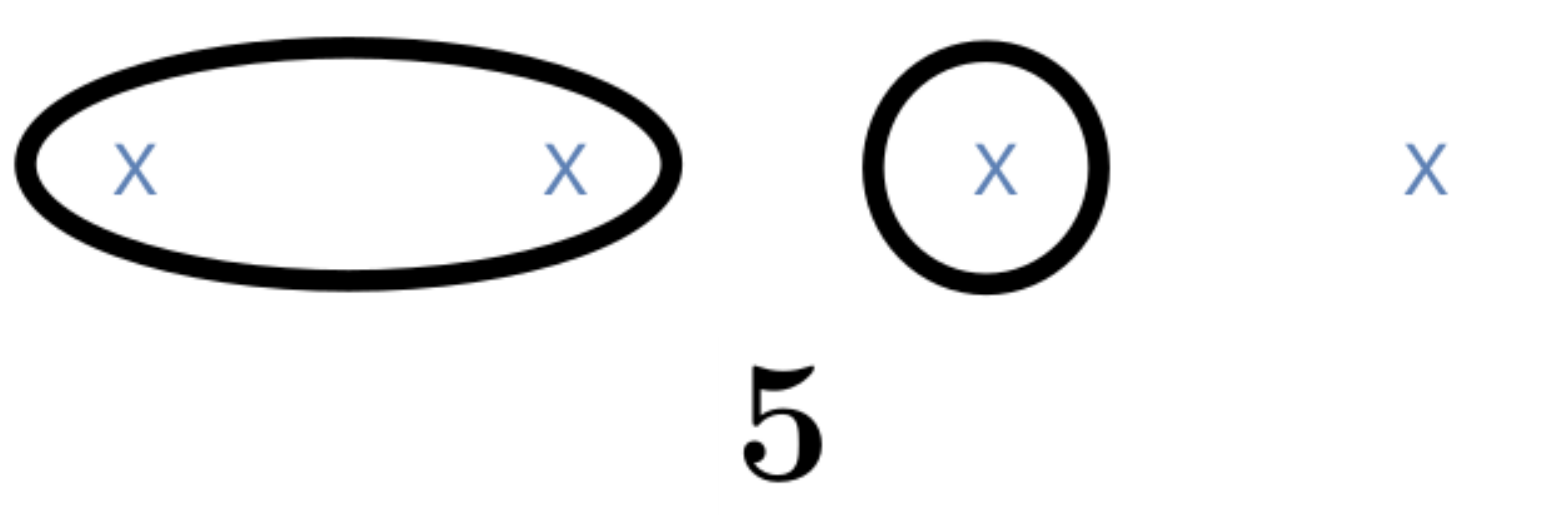}
    \end{subfigure}%
    \begin{subfigure}[t]{0.25\textwidth}
        \centering
        \includegraphics[width=0.8\textwidth]{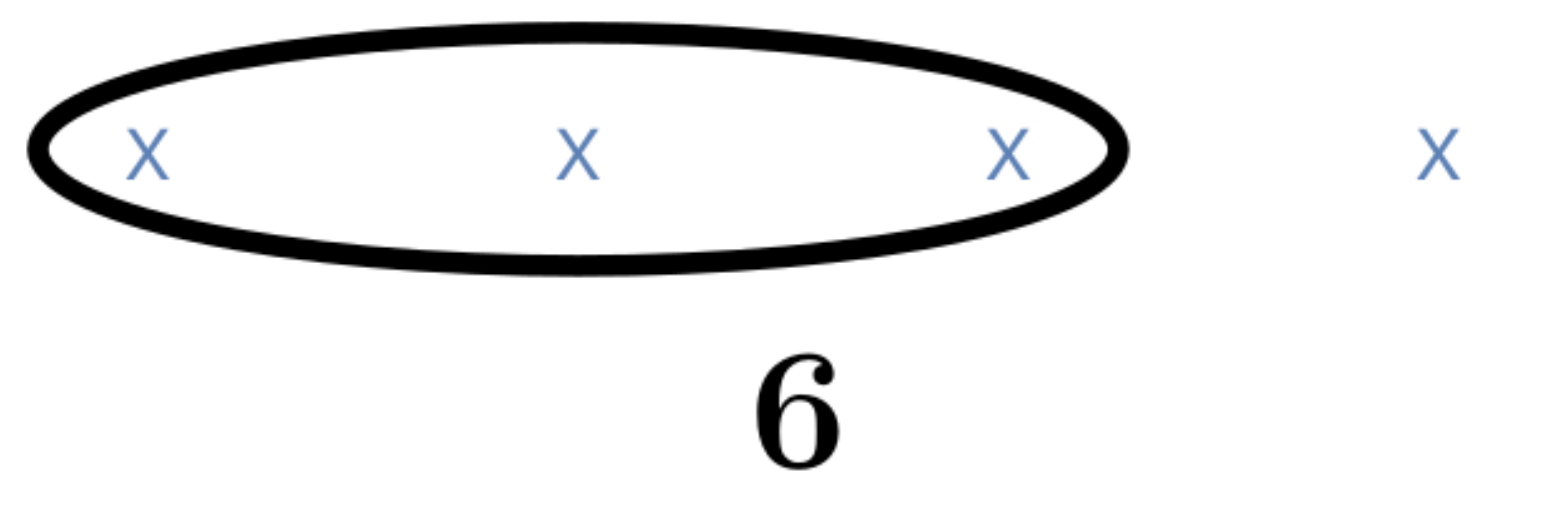}
    \end{subfigure}%
    \begin{subfigure}[t]{0.25\textwidth}
        \centering
        \includegraphics[width=0.8\textwidth]{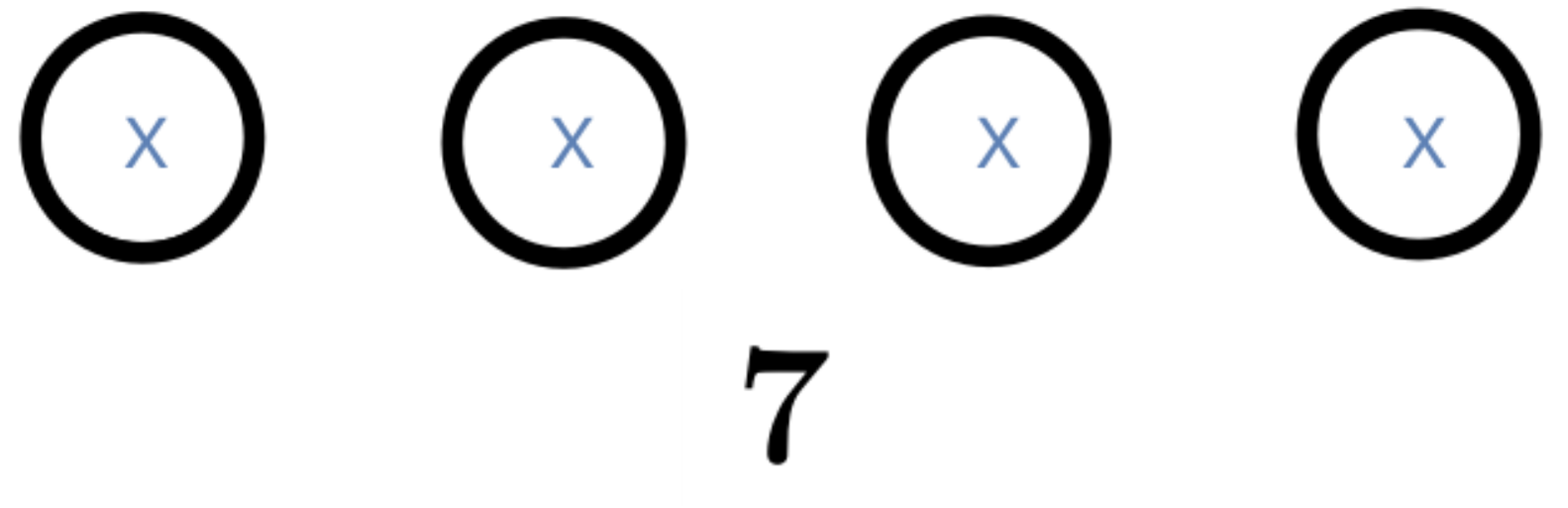}
    \end{subfigure}\\
    \vspace{0.3cm}
    \begin{subfigure}[t]{0.25\textwidth}
        \centering
        \includegraphics[width=0.8\textwidth]{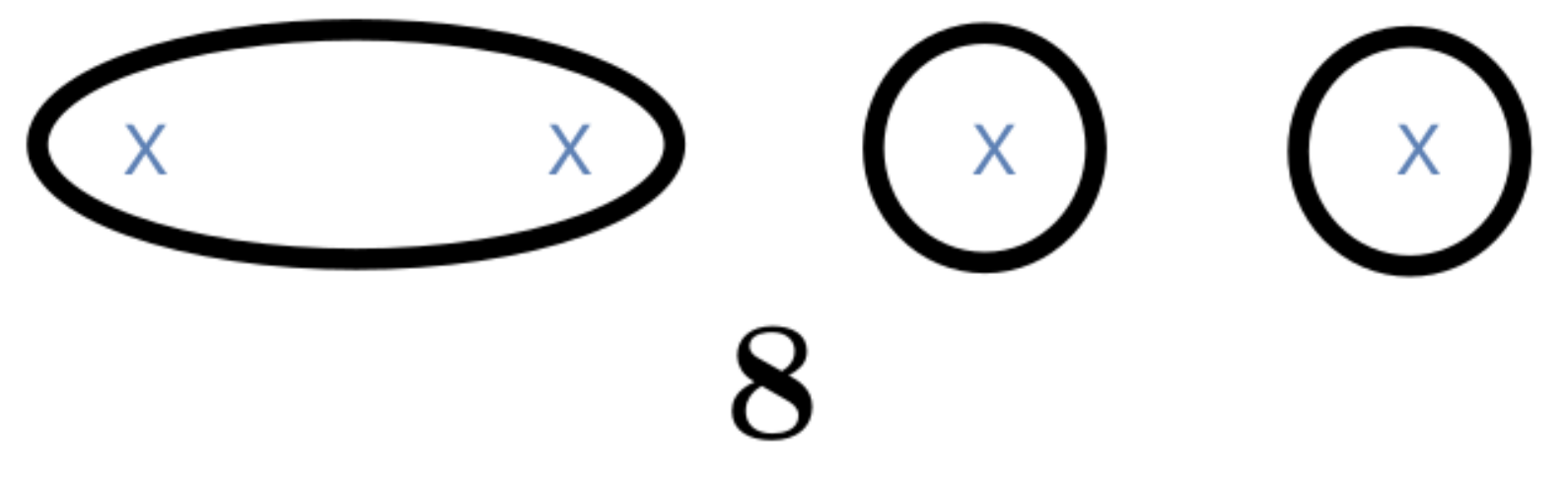}
    \end{subfigure}%
    \begin{subfigure}[t]{0.25\textwidth}
        \centering
        \includegraphics[width=0.8\textwidth]{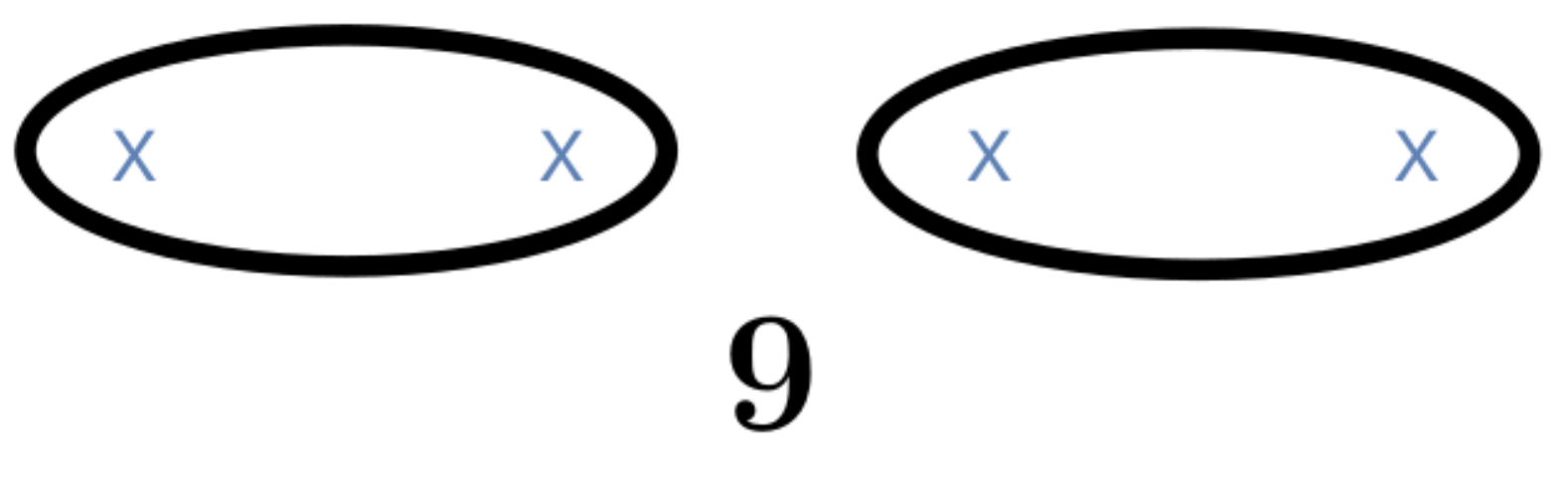}
    \end{subfigure}%
    \begin{subfigure}[t]{0.25\textwidth}
        \centering
        \includegraphics[width=0.8\textwidth]{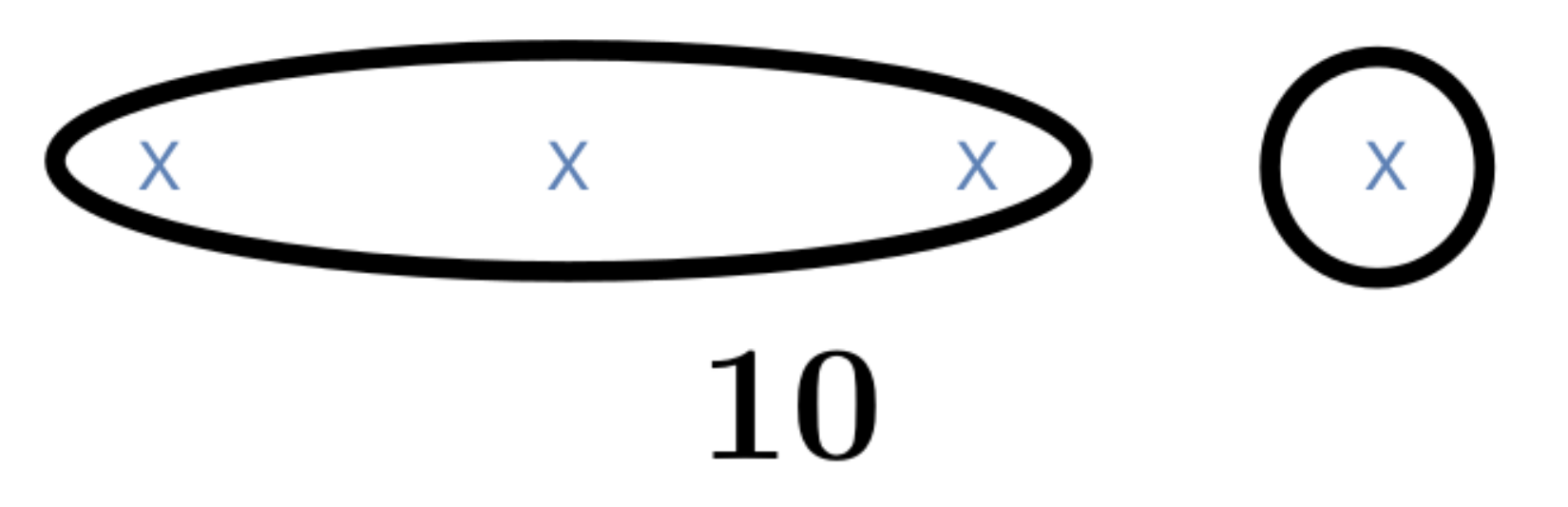}
    \end{subfigure}%
    \begin{subfigure}[t]{0.25\textwidth}
        \centering
        \includegraphics[width=0.8\textwidth]{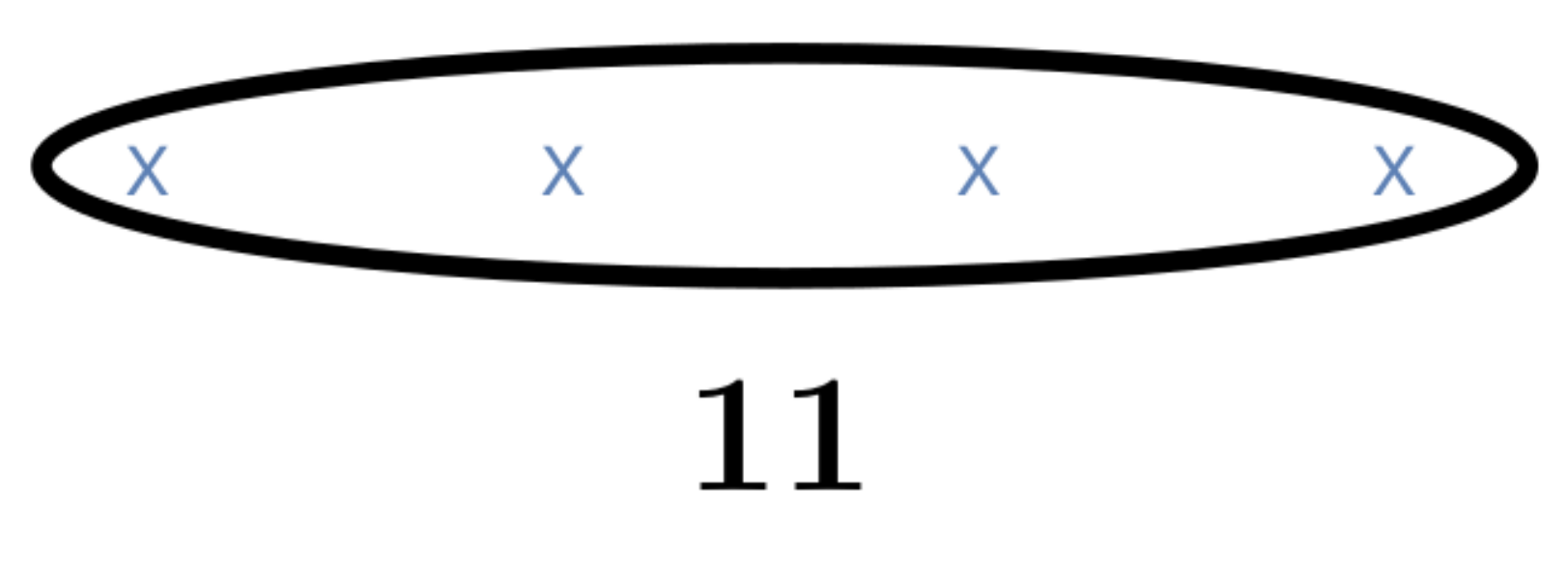}
    \end{subfigure} \\
    \caption{The structure type scheme used to distinguish clusterings
containing outliers. Each cross represents an outlier in the $K$-means
landscape, irrespective of position in space. Circles surrounding one or more
outliers indicate membership of a distinct cluster with no original data
points. All other outliers are incorporated into clusters with original data
points. The scheme is shown for four outliers, but the ordering remains the
same for datasets with fewer than four.}
    \label{OutliersScheme}
\end{figure}

The two outlier solutions distinguished above can be used to understand the
structure of the clustering solutions within $K$-means landscapes. For datasets
with multiple outliers, each can correspond to one of the two cases, and we
provide a classification of the possible structures in
Fig.~\ref{OutliersScheme}.

The structure type, in our specified scheme, is independent of the outlier
performing the function. For example, we do not consider which outlier
corresponds to its own cluster, giving four options for structure type 1 in a
dataset containing four outliers. Each different type 1 solution will have
different values of the cost function depending upon where the outlier lies in
relation to the original data. However, the scheme remains valuable for
understanding the relevant structure of the clusterings.

\subsection{Cost function topography}

The topography of the solution space is central to global optimisation of all
machine learning algorithms, and for $K$-means it determines the ease of
locating the optimal clustering. First, we visualise the global structure of
the solution space, and describe its properties with reference to global
optimisation. Subsequently, we use kinetic analysis to further explore the
intermediate regions of solution space, and finally address the properties of
the clustering solutions.

\subsubsection{Disconnectivity graphs}

\begin{figure}
    \centering
    \begin{subfigure}[t]{0.28\textwidth}
        \centering
        \psfrag{epsilon}{\small{25}}
        \psfrag{a}{\scriptsize{0}}
        \psfrag{b}{\scriptsize{1}}
        \includegraphics[width=1.00\textwidth]{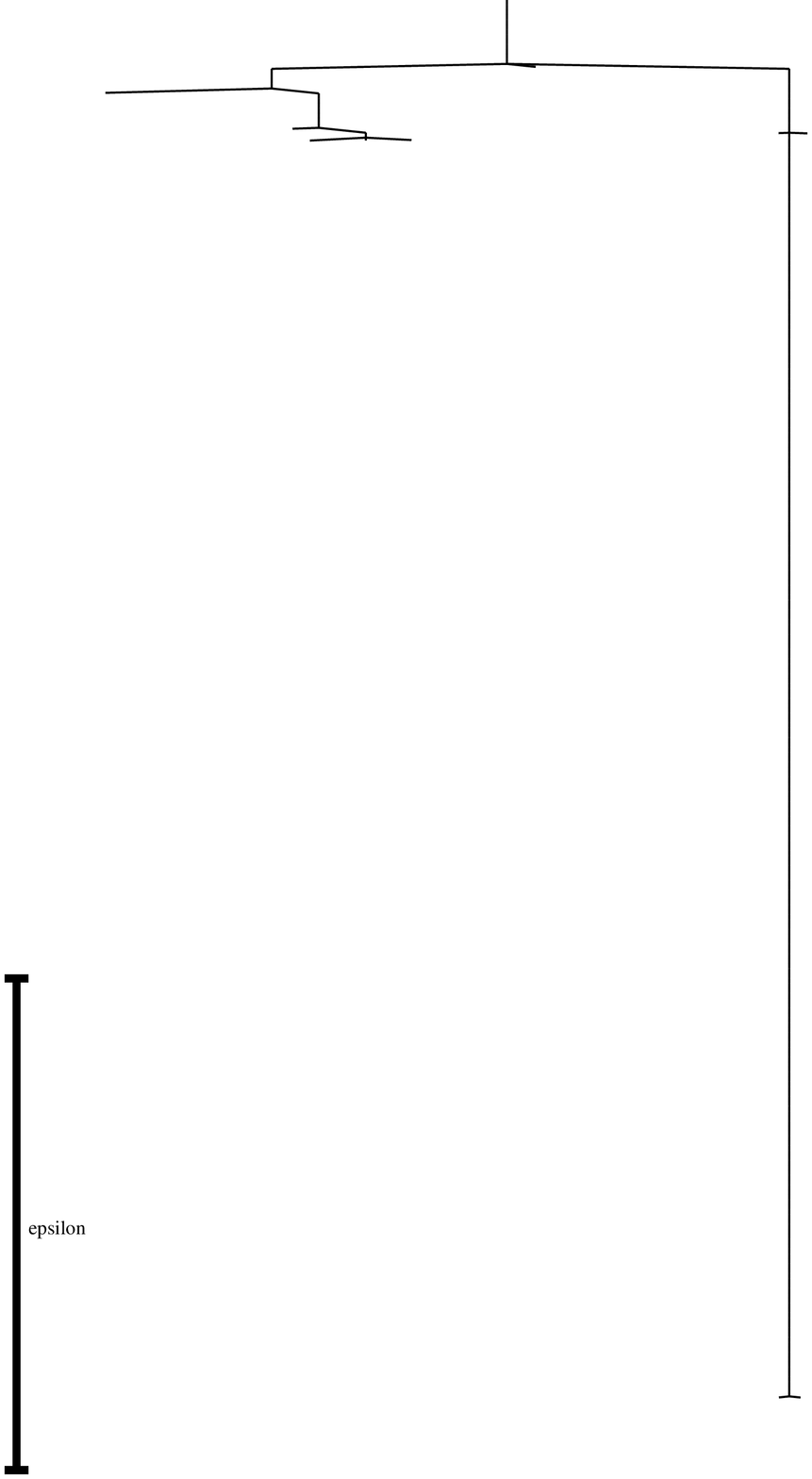}
        \caption{Original}
    \end{subfigure}%
    \begin{subfigure}[t]{0.28\textwidth}
        \centering
        \psfrag{epsilon}{\small{}}
        \psfrag{a}{\scriptsize{0}}
        \psfrag{b}{\scriptsize{1}}
        \includegraphics[width=1.00\textwidth]{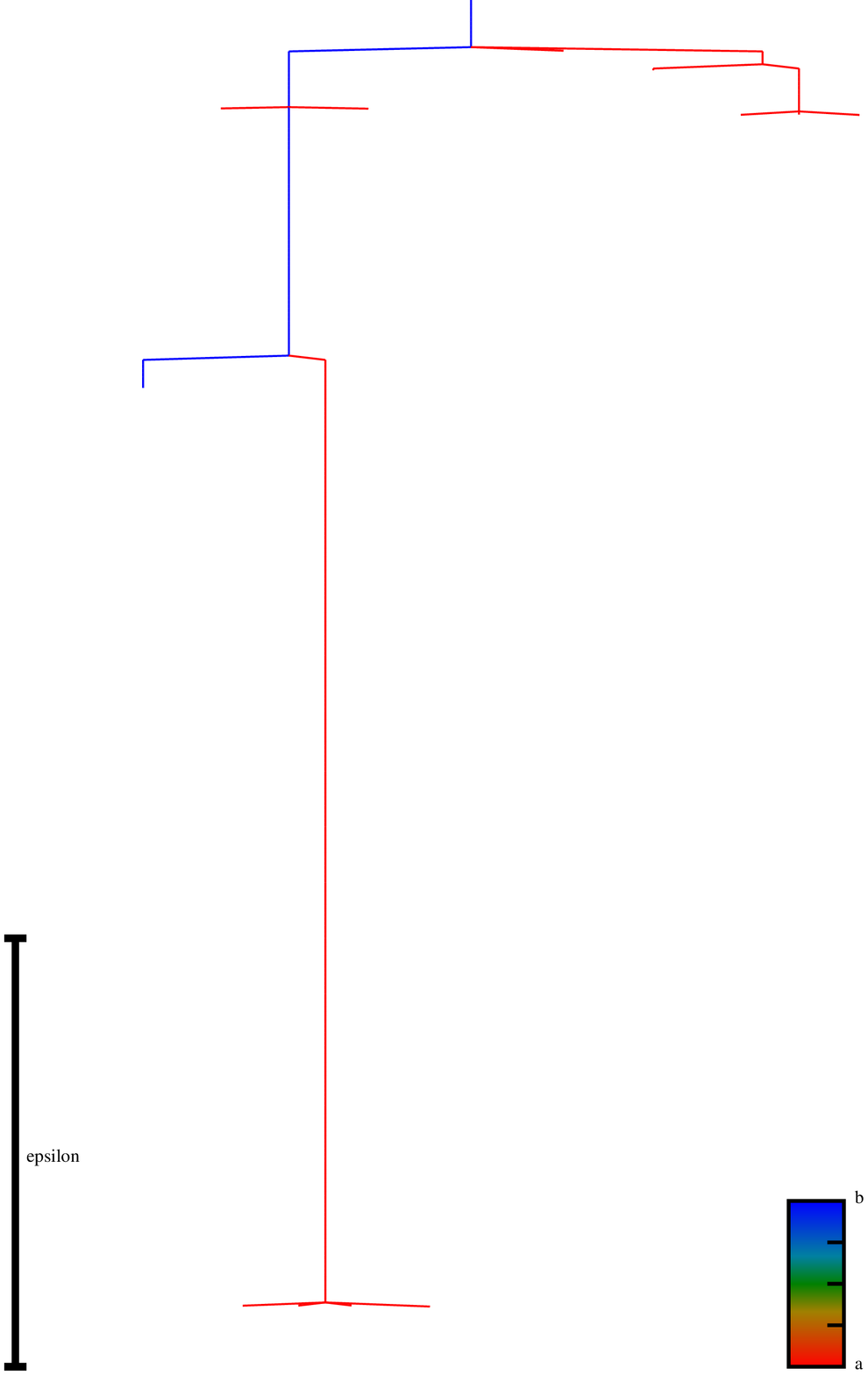}
        \caption{Outlier 1}
    \end{subfigure}%
    \begin{subfigure}[t]{0.28\textwidth}
        \centering
        \psfrag{epsilon}{\small{}}
        \psfrag{a}{\scriptsize{0}}
        \psfrag{b}{\scriptsize{3}}
        \includegraphics[width=1.00\textwidth]{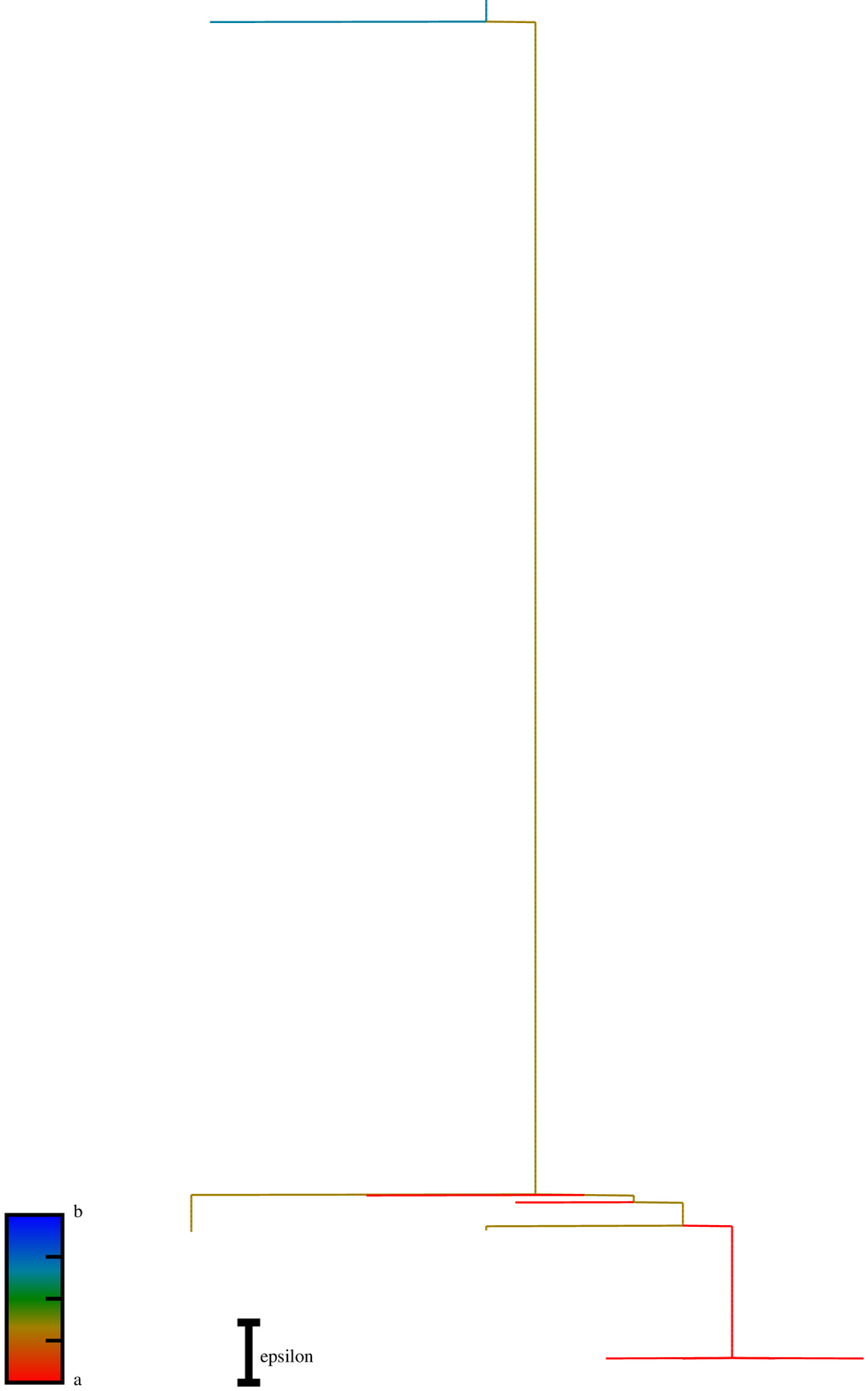}
        \caption{Outlier 2}
    \end{subfigure} \\
    \begin{subfigure}[t]{0.28\textwidth}
        \centering
        \psfrag{epsilon}{\small{}}
        \psfrag{a}{\scriptsize{0}}
        \psfrag{b}{\scriptsize{5}}
        \includegraphics[width=1.00\textwidth]{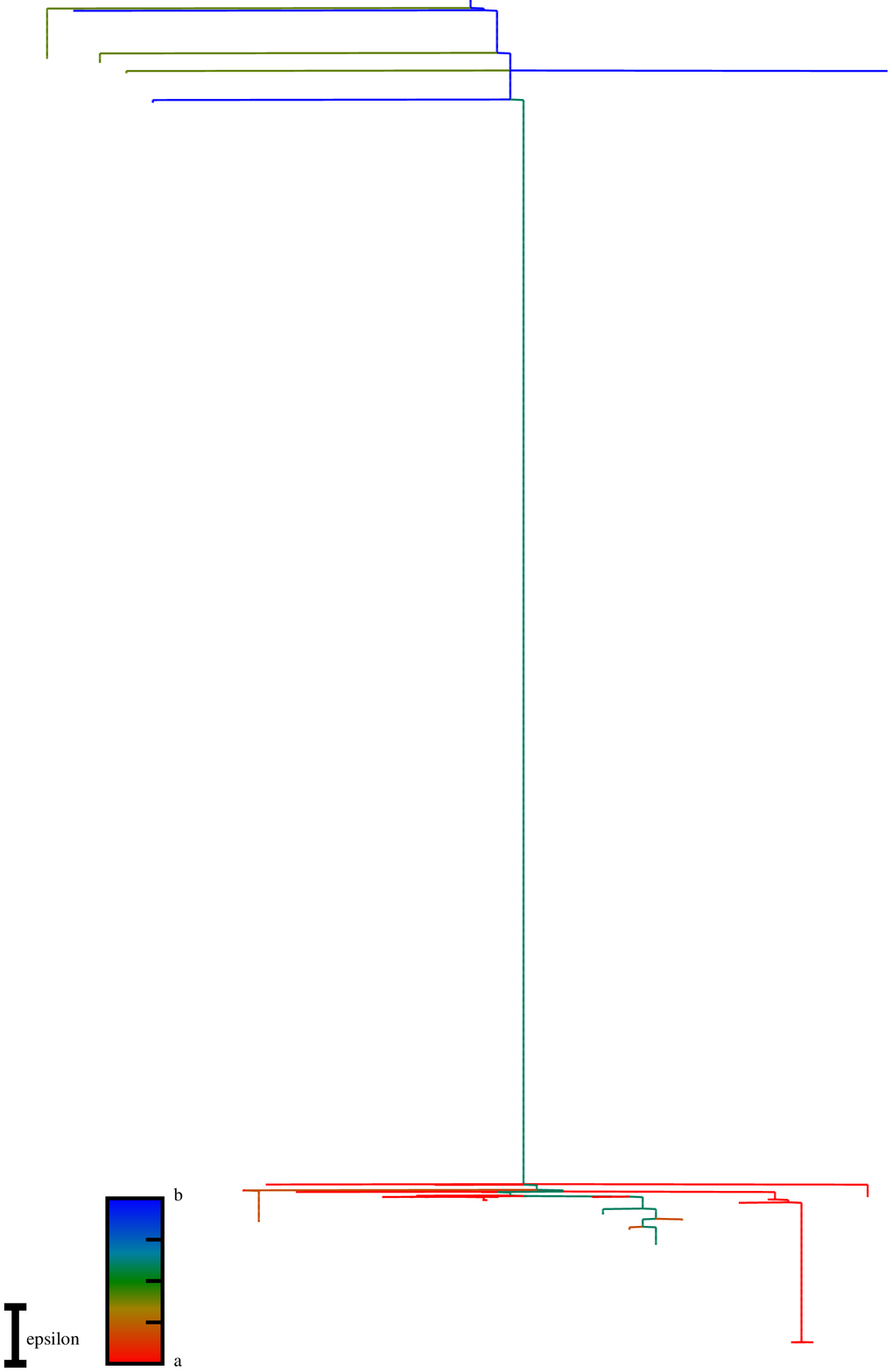}
        \caption{Outlier 3}
    \end{subfigure}%
    \begin{subfigure}[t]{0.28\textwidth}
        \centering
        \psfrag{epsilon}{\small{}}
        \psfrag{a}{\scriptsize{0}}
        \psfrag{b}{\scriptsize{10}}
        \includegraphics[width=1.00\textwidth]{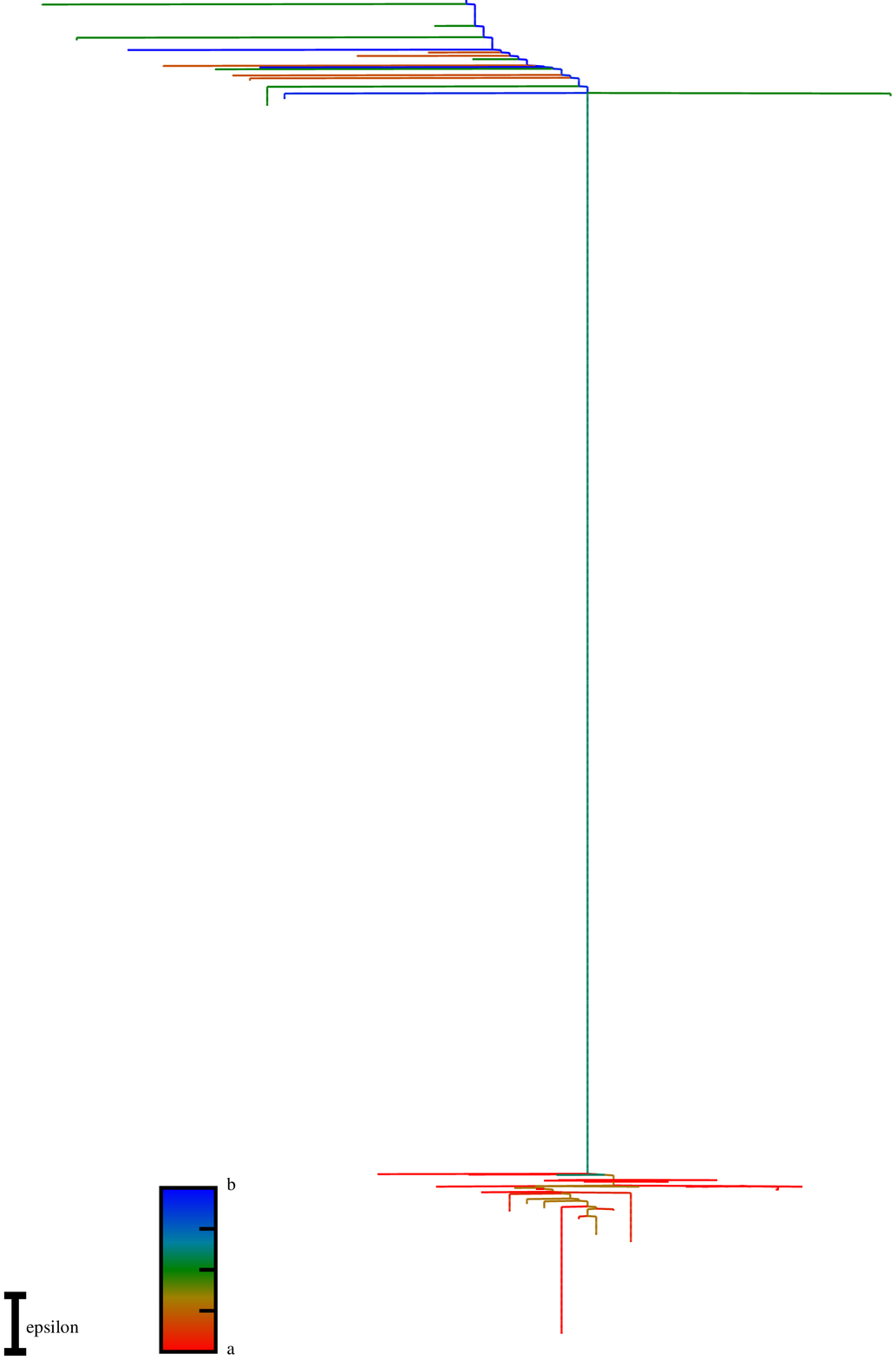}
        \caption{Outlier 4}
    \end{subfigure} \\
    \caption{$K$-means landscapes for Fisher's \textit{Iris} dataset with $K=3$
and an increasing number of outliers. The scale bar represents the same cost
function range in all plots. Minima are coloured according to the scheme, given
in Fig.~\ref{OutliersScheme}, for distinguishing different types of $K$-means
solutions.}
    \label{OutlierStructureIris3DGs}
\end{figure}

\begin{figure}
    \centering
    \begin{subfigure}[t]{0.3\textwidth}
        \centering
        \psfrag{epsilon}{\small{10}}
        \includegraphics[width=1.00\textwidth]{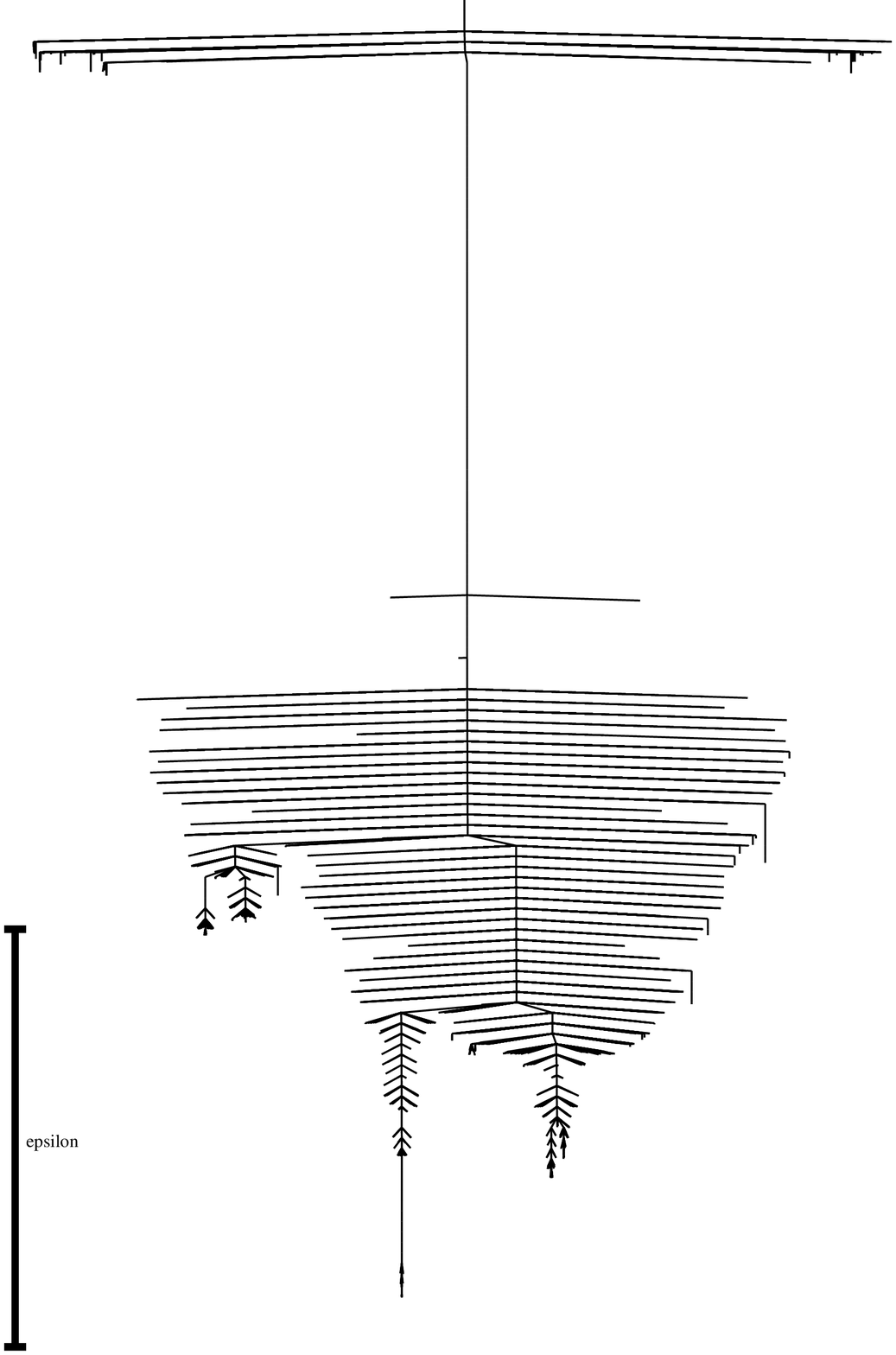}
        \caption{Original}
    \end{subfigure}%
    \begin{subfigure}[t]{0.3\textwidth}
        \centering
        \psfrag{epsilon}{\small{}}
        \psfrag{a}{\scriptsize{0}}
        \psfrag{b}{\scriptsize{1}}
        \includegraphics[width=1.00\textwidth]{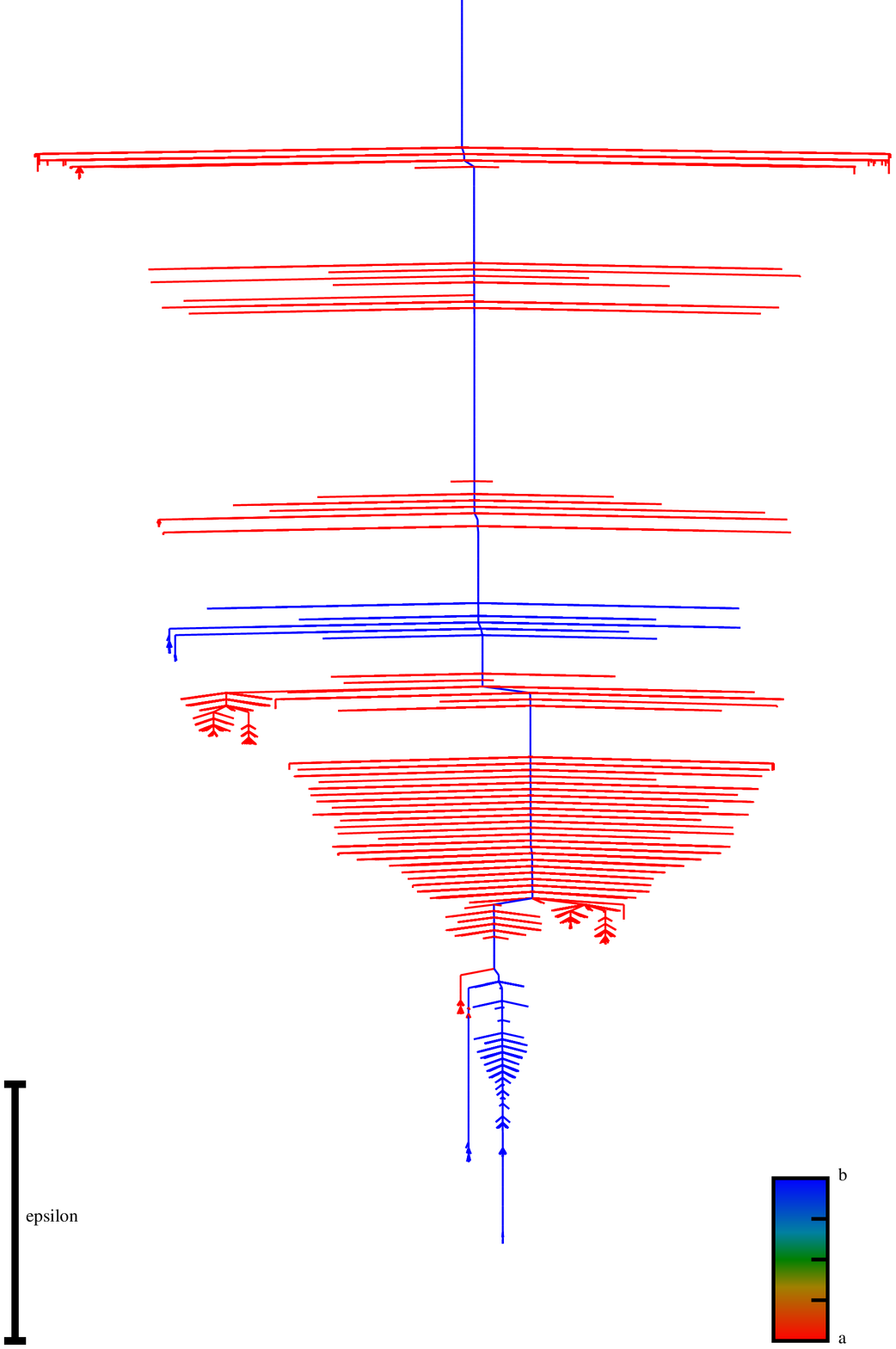}
        \caption{Outlier 1}
    \end{subfigure}%
    \begin{subfigure}[t]{0.3\textwidth}
        \centering
        \psfrag{epsilon}{\small{}}
        \psfrag{a}{\scriptsize{0}}
        \psfrag{b}{\scriptsize{3}}
        \includegraphics[width=1.00\textwidth]{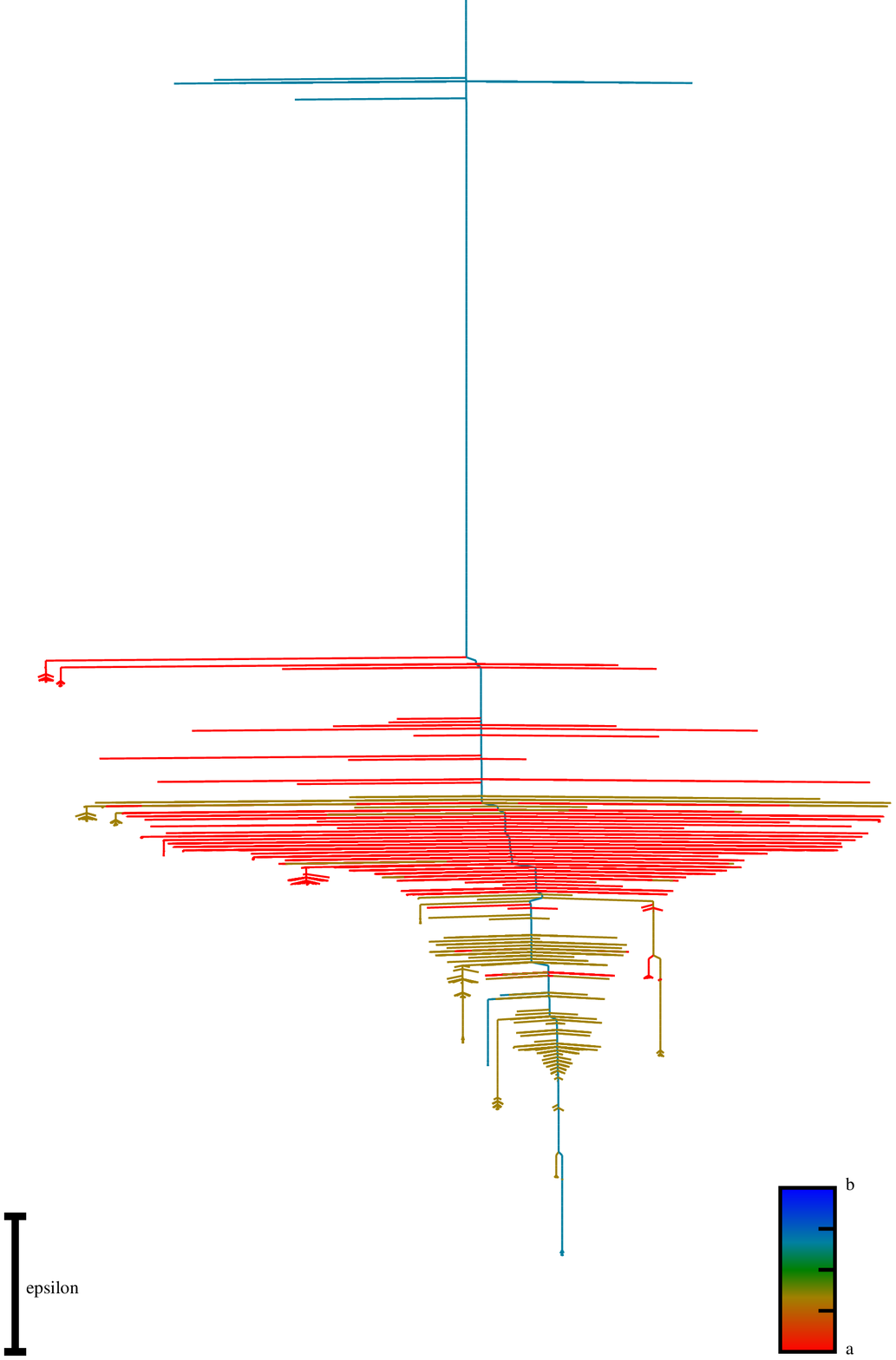}
        \caption{Outlier 2}
    \end{subfigure} \\
    \begin{subfigure}[t]{0.3\textwidth}
        \centering
        \psfrag{epsilon}{\small{}}
        \psfrag{a}{\scriptsize{0}}
        \psfrag{b}{\scriptsize{5}}
        \includegraphics[width=1.00\textwidth]{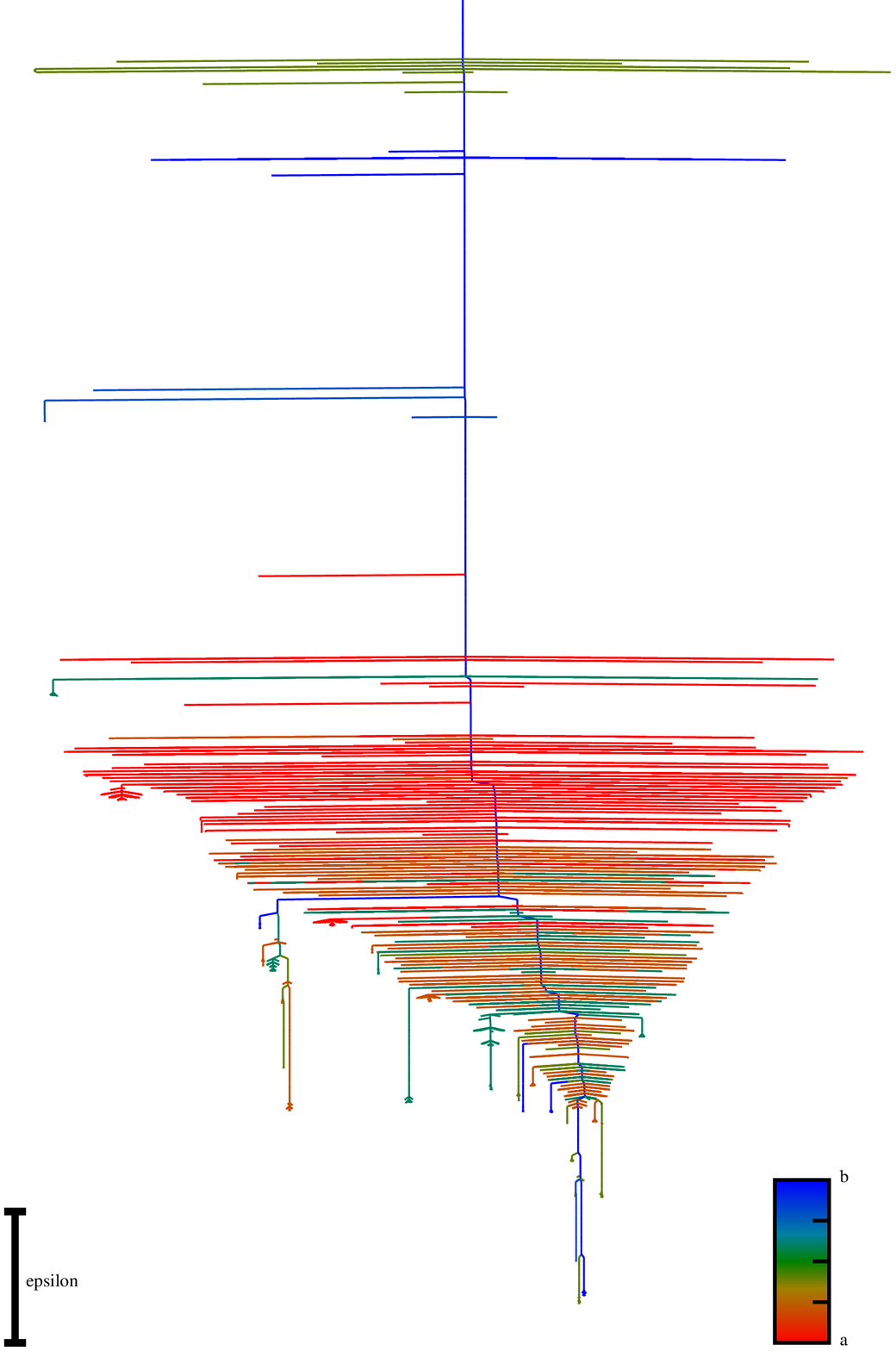}
        \caption{Outlier 3}
    \end{subfigure}%
    \begin{subfigure}[t]{0.3\textwidth}
        \centering
        \psfrag{epsilon}{\small{}}
        \psfrag{a}{\scriptsize{0}}
        \psfrag{b}{\scriptsize{9}}
        \includegraphics[width=1.00\textwidth]{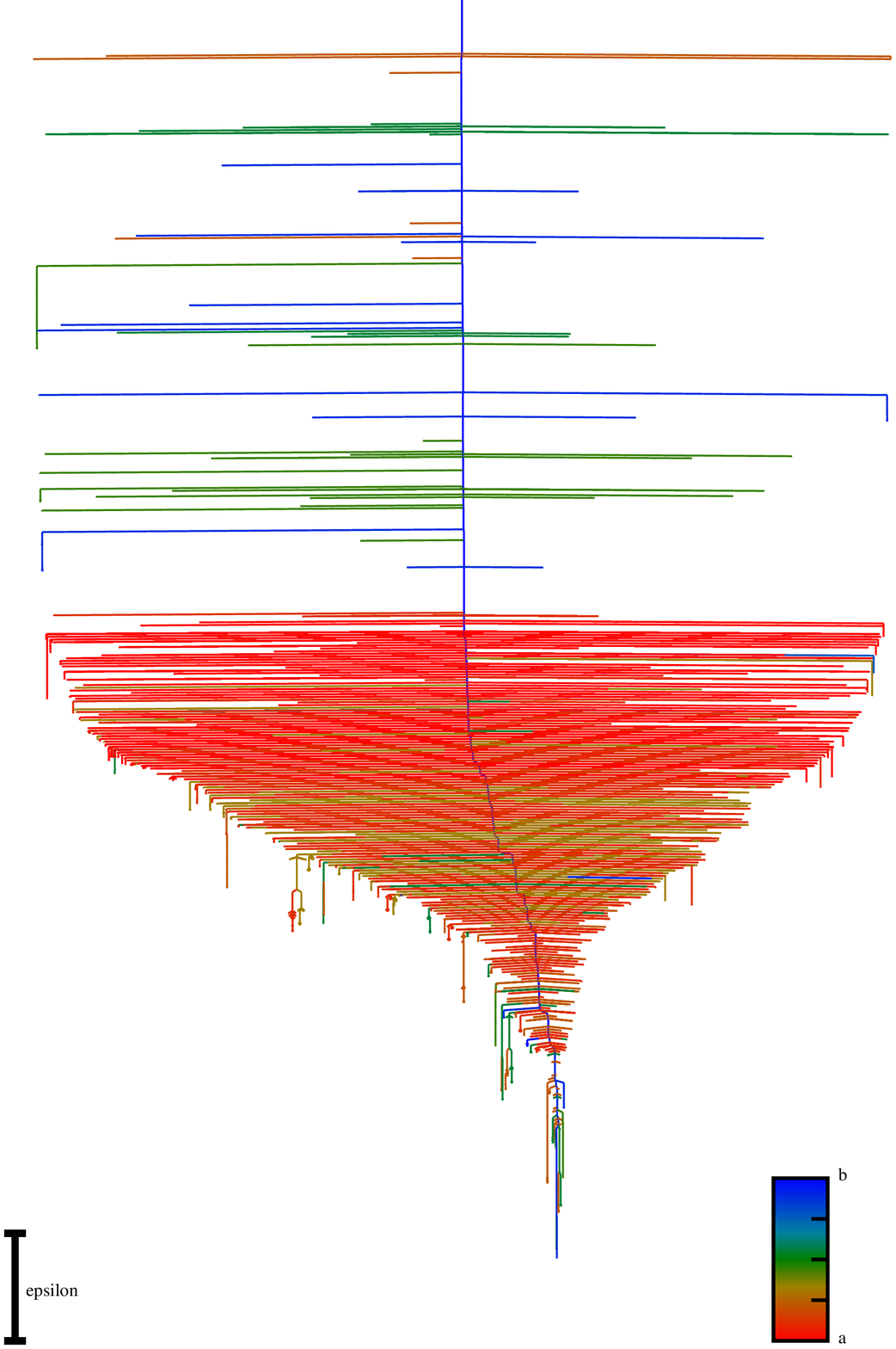}
        \caption{Outlier 4}
    \end{subfigure}
    \caption{Disconnectivity graphs showing the $K$-means solution landscapes
for variants of Fisher's \textit{Iris} dataset with six clusters. The scale bar
represents the same cost function range in each case, and
minima are coloured according to the scheme given for distinguishing different
types of $K$-means solutions.}
    \label{OutlierStructureIris6DGs}
\end{figure}

\begin{figure}
    \centering
    \begin{subfigure}[t]{0.3\textwidth}
        \centering
        \psfrag{epsilon}{\small{80}}
        \includegraphics[width=1.0\textwidth]{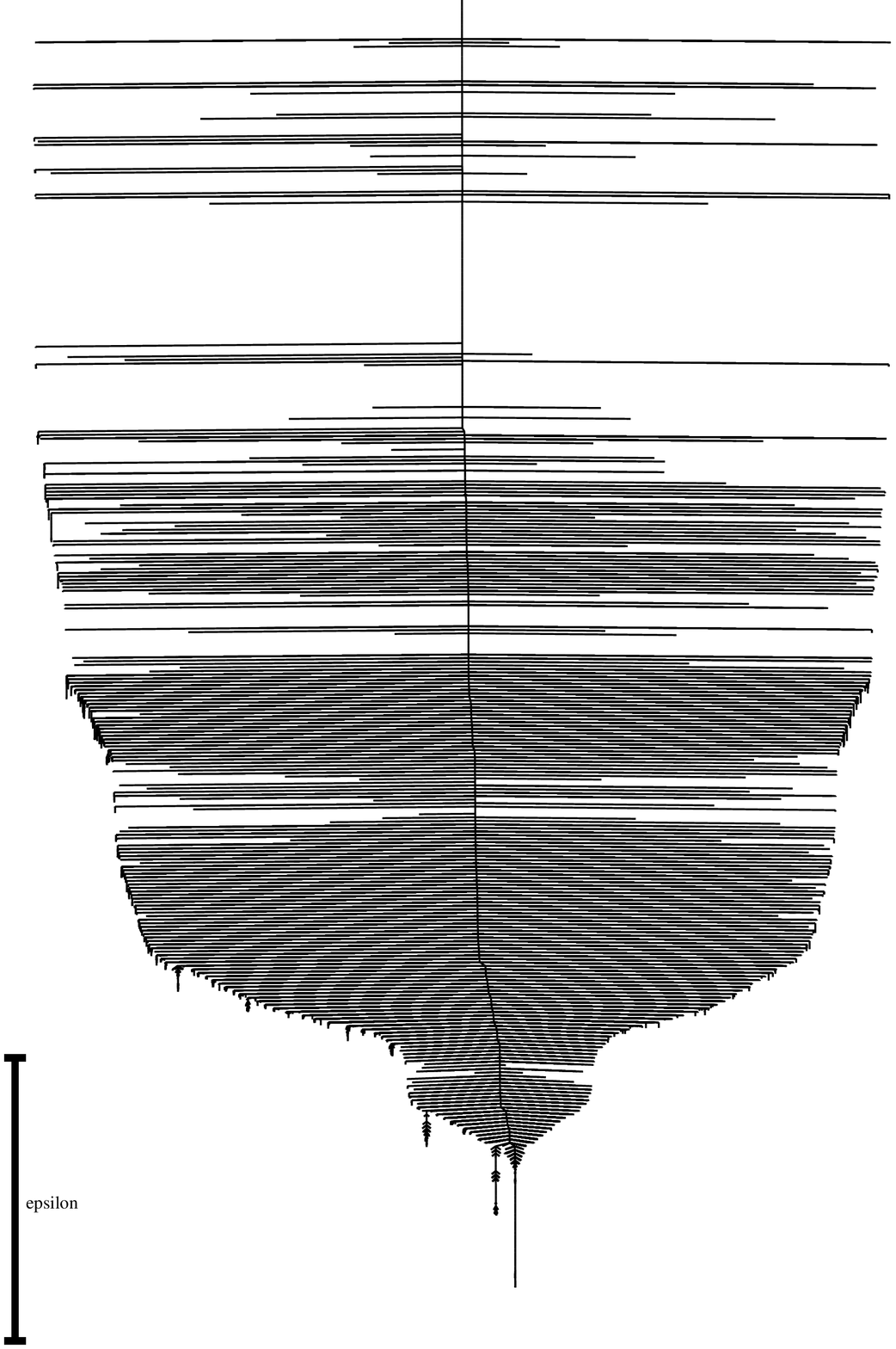}
        \caption{Original}
    \end{subfigure}%
    \begin{subfigure}[t]{0.3\textwidth}
        \centering
        \psfrag{epsilon}{\small{}}
        \psfrag{a}{\scriptsize{0}}
        \psfrag{b}{\scriptsize{1}}
        \includegraphics[width=1.0\textwidth]{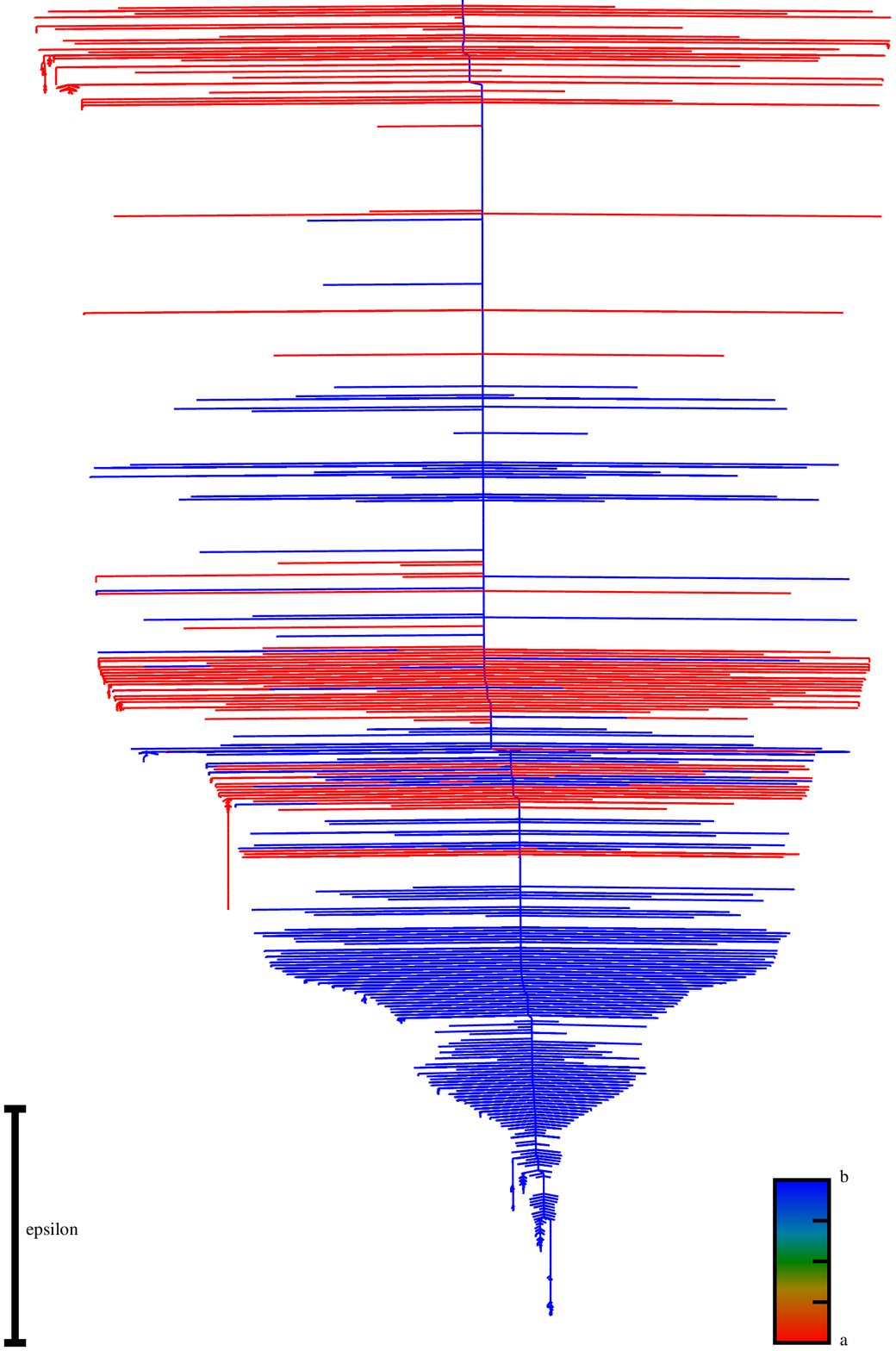}
        \caption{Outlier 1}
    \end{subfigure} \\
    \begin{subfigure}[t]{0.3\textwidth}
        \centering
        \psfrag{epsilon}{\small{}}
        \psfrag{a}{\scriptsize{0}}
        \psfrag{b}{\scriptsize{3}}
        \includegraphics[width=1.0\textwidth]{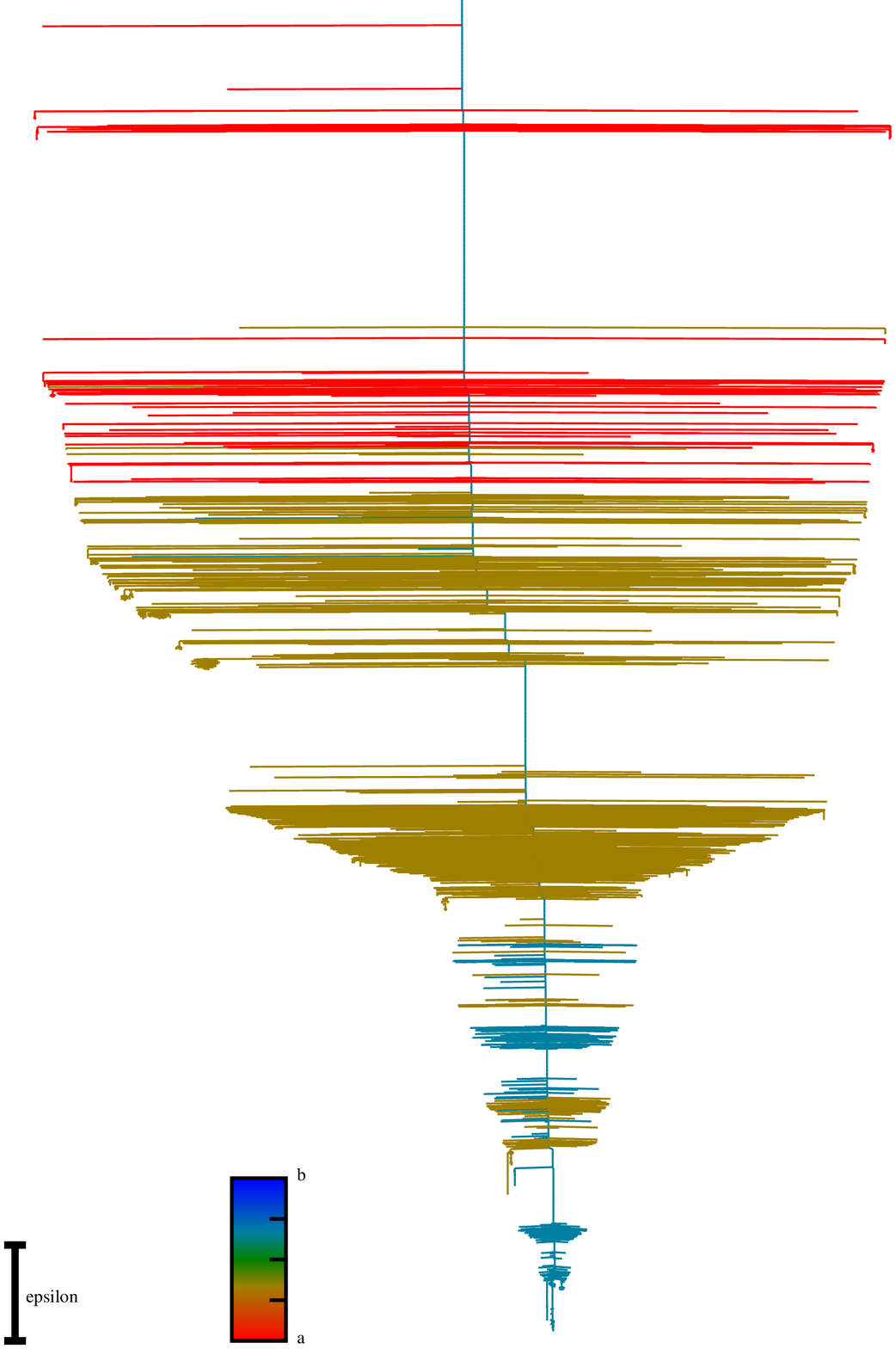}
        \caption{Outlier 2}
    \end{subfigure}%
    \begin{subfigure}[t]{0.3\textwidth}
        \centering
        \psfrag{epsilon}{\small{}}
        \psfrag{a}{\scriptsize{0}}
        \psfrag{b}{\scriptsize{5}}
        \includegraphics[width=1.0\textwidth]{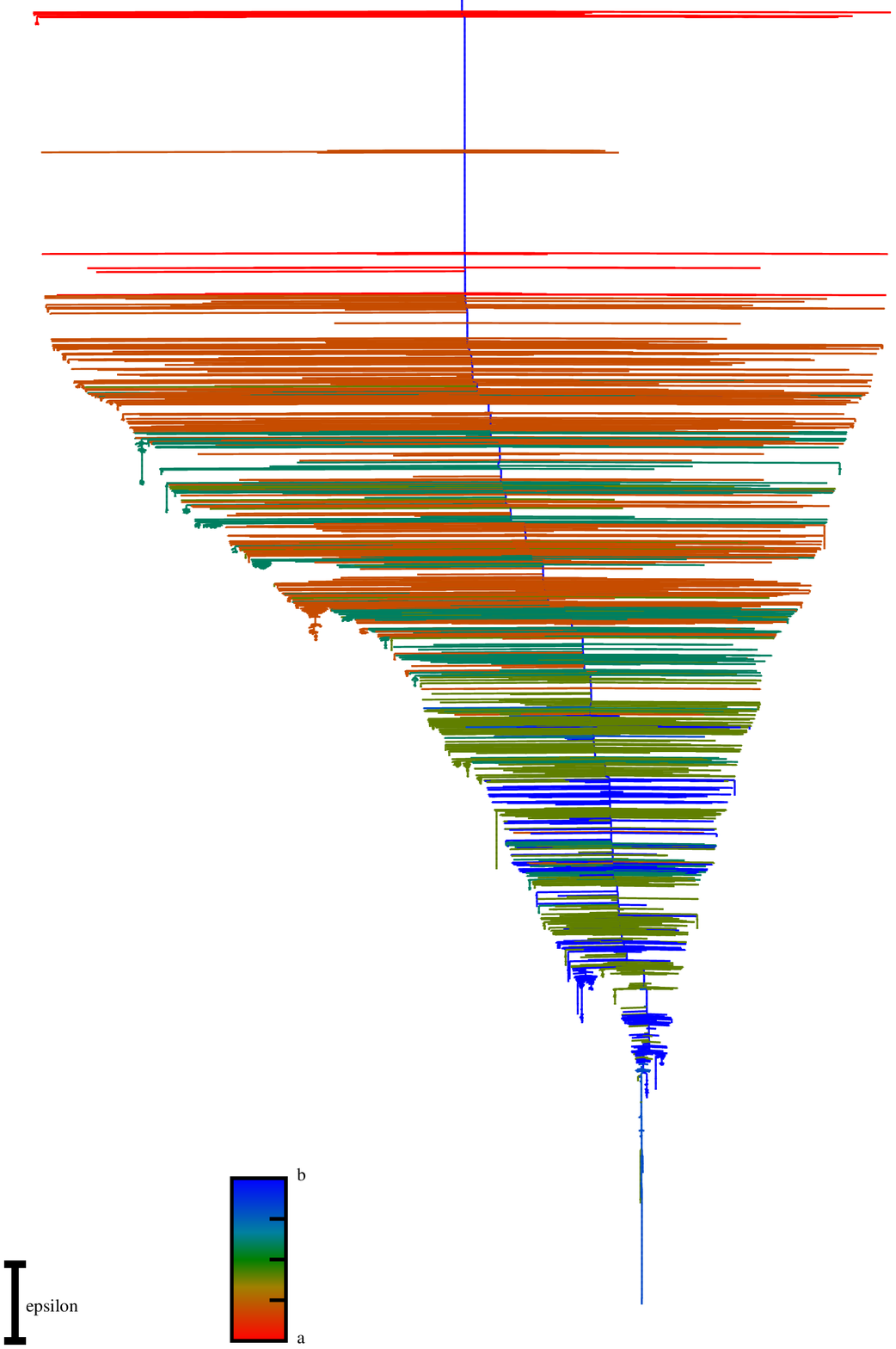}
        \caption{Outlier 3}
    \end{subfigure}%
    \caption{$K$-means solution landscapes for the glass identification
dataset, and variations with additional outliers. Minima are coloured according
to the structure types specified in Fig.~\ref{OutliersScheme}. The scale bar
corresponds to the same cost function range in each case.}
    \label{OutlierStructureGlassDGs}
\end{figure}

We present $K$-means landscapes, visualised using disconnectivity graphs, for
Fisher's \textit{Iris} dataset ($K=3$ and $K=6$) and the glass identification
dataset ($K=6$) prepared with an increasing number of outliers in
Figs.~\ref{OutlierStructureIris3DGs}, \ref{OutlierStructureIris6DGs} and
\ref{OutlierStructureGlassDGs}, respectively. The minima are coloured according
to structure type as specified in the previous section. Despite the large
increase in the cost function range across $K$-means landscapes with additional
outliers, there is little change in the barrier heights between transition
states and their connected minima, which generally remain very small
throughout. Furthermore, the barriers to the global minimum, which can involve
pathways containing many minima and transition states, also remain small for
the vast majority of minima. This structure produces a
strong funnelling in the cost function surface. Single-funnel
landscapes\cite{socciow98} are characterised by efficient relaxation to a
well-defined global minimum, as for biomolecules that have evolved to perform a
single function.

Despite the largely funnelled organisation of minima, there remain regions of
solution space separated from the global minimum by significant barriers in all
the $K$-means landscapes. These regions of the surface are referred to as
kinetic traps, because they slow relaxation to the global minimum in physical
systems, and impede global optimisation algorithms.\cite{Bryngelson1995,
Onuchic1997} The kinetic trap regions tend to retain a strong local funnelling
to their lowest minimum, and constitute regions of solution space that are
distinct from the global minimum.

The number of kinetic traps is generally comparable in $K$-means landscapes
corresponding to datasets with more outliers, despite the large increase in the
number of different structure types available to the clustering. However, as
the number of outliers increases there is a smaller proportion of minima
associated with these kinetic traps. Therefore, with the small barriers
retained for many minima, and fewer minima in kinetic traps, the landscapes
appear increasingly funnelled for a greater number of outliers.

\subsubsection{Frustration}

One important quantification of landscape funnelling is the frustration, which
we diagnose using the Shannon entropy,\cite{Shannon1948} which has
previously been for both molecular energy
landscapes\cite{deSouza2017} and $K$-means landscapes.\cite{Wu2023} Increasing
frustration values are associated with a reduction in landscape funnelling due
to more kinetic traps, larger barriers, and more alternative low-valued
solutions. Greater frustration highlights an increased challenge for global
optimisation and an increase in the number of different regions of solution
space that need to be considered during clustering.

\begin{figure}
    \centering
    \begin{subfigure}[t]{0.5\textwidth}
        \centering
        \includegraphics[width=1.00\textwidth]{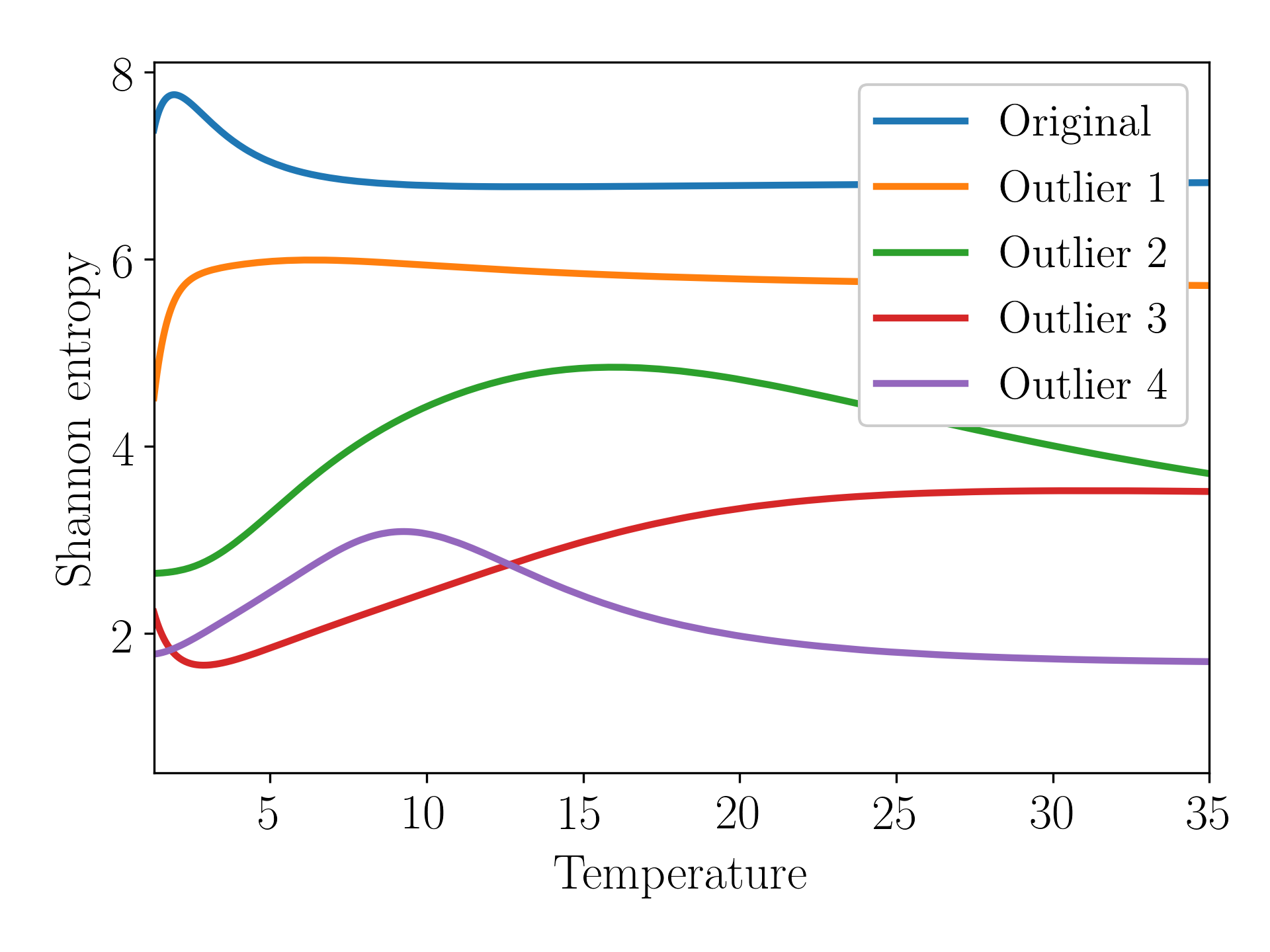}
        \caption{\textit{Iris} ($K=6$)}
    \end{subfigure}%
    \begin{subfigure}[t]{0.5\textwidth}
        \centering
        \includegraphics[width=1.00\textwidth]{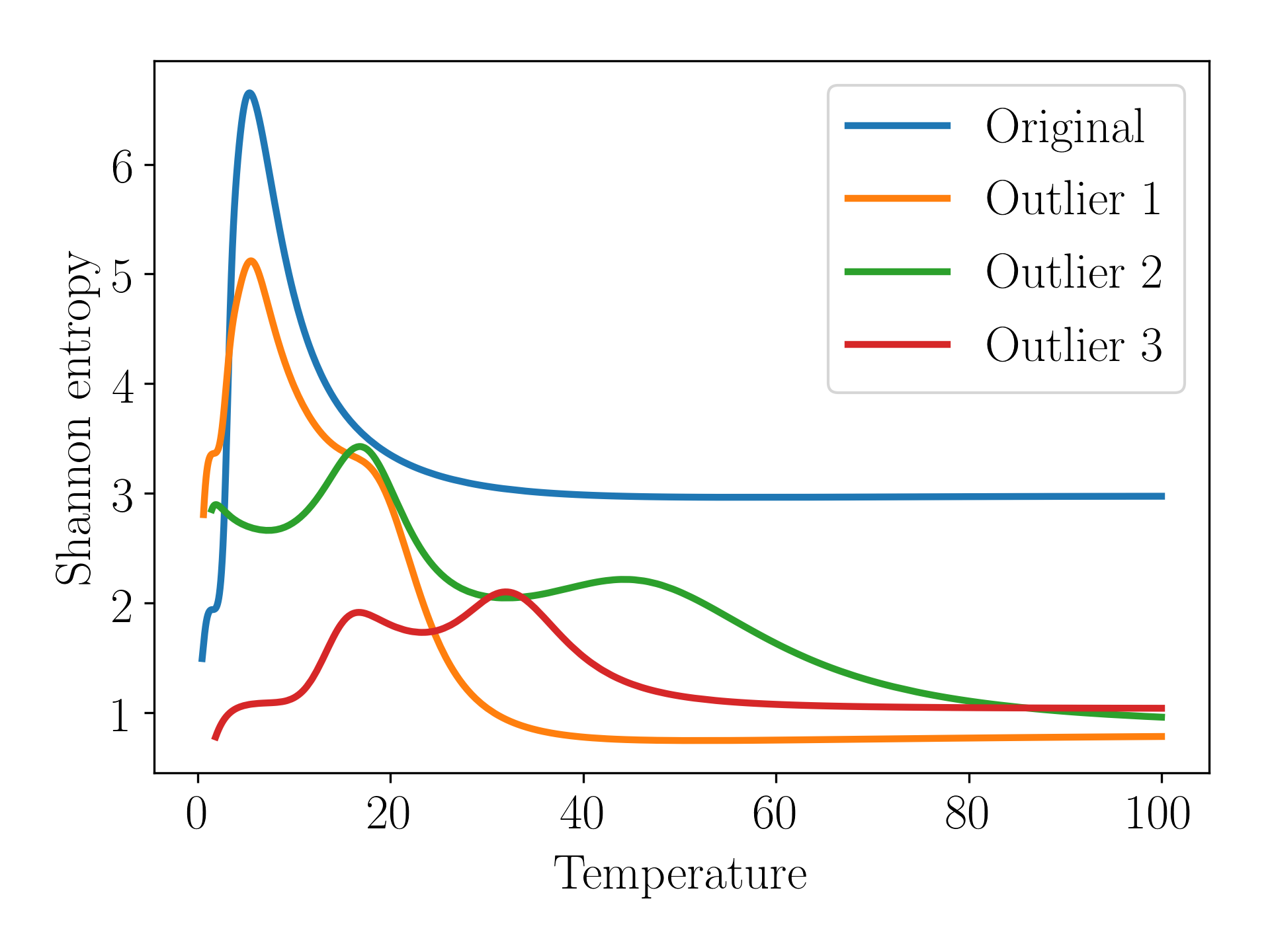}
        \caption{Glass}
    \end{subfigure}%
    \caption{The Shannon entropy, used as a measure of landscape frustration,
calculated for each solution landscape from the \textit{Iris} and glass
datasets. The entropy quantifies the complexity of the cost function surface 
as a function of the fictitious
temperature, where $k_{\mathrm{B}}T$ has the same units as the cost function. Local minima
with higher cost functions are progressively occupied as the temperature increases.}
    \label{OutliersFrustration}
\end{figure}

Shannon entropy profiles are shown for the \textit{Iris} and glass landscapes
in Fig.~\ref{OutliersFrustration}. The frustration metric is not presented for
the \textit{Iris} dataset with three clusters, due to the limited number of
minima. For both landscapes we see a smooth decrease in frustration as the
number of outliers increases. Therefore, the stronger funnelling seen in the
disconnectivity graphs is further reflected by the frustration metric, which
demonstrates that the addition of outliers drives funnelling of the landscapes.
Furthermore, for the glass landscapes we observe multiple peaks in the
frustration profiles. These features correspond to strong fluctuations in
either the degeneracy of minima, or the size of their respective basins of
attraction, for minima at progressively higher cost function. Such fluctuations
hint at the appearance of distinct types of clustering becoming available,
which would otherwise be difficult to discern without analysing all the data
point assignments.

\subsubsection{Kinetic analysis}

Disconnectivity graphs provide a representation of the overall barriers, but the
application of kinetic analysis produces a more detailed view of
the solution space organisation. Rates and pathways calculated between
specified minima include contributions from both the intermediate barriers and
minima that describe the space between solutions. We separate out two types of
pathways to the global minimum, those starting from minima within the main
funnel and those from kinetic traps.

\begin{figure}
    \centering
    \begin{subfigure}[t]{0.45\textwidth}
        \centering
        \includegraphics[width=1.00\textwidth]{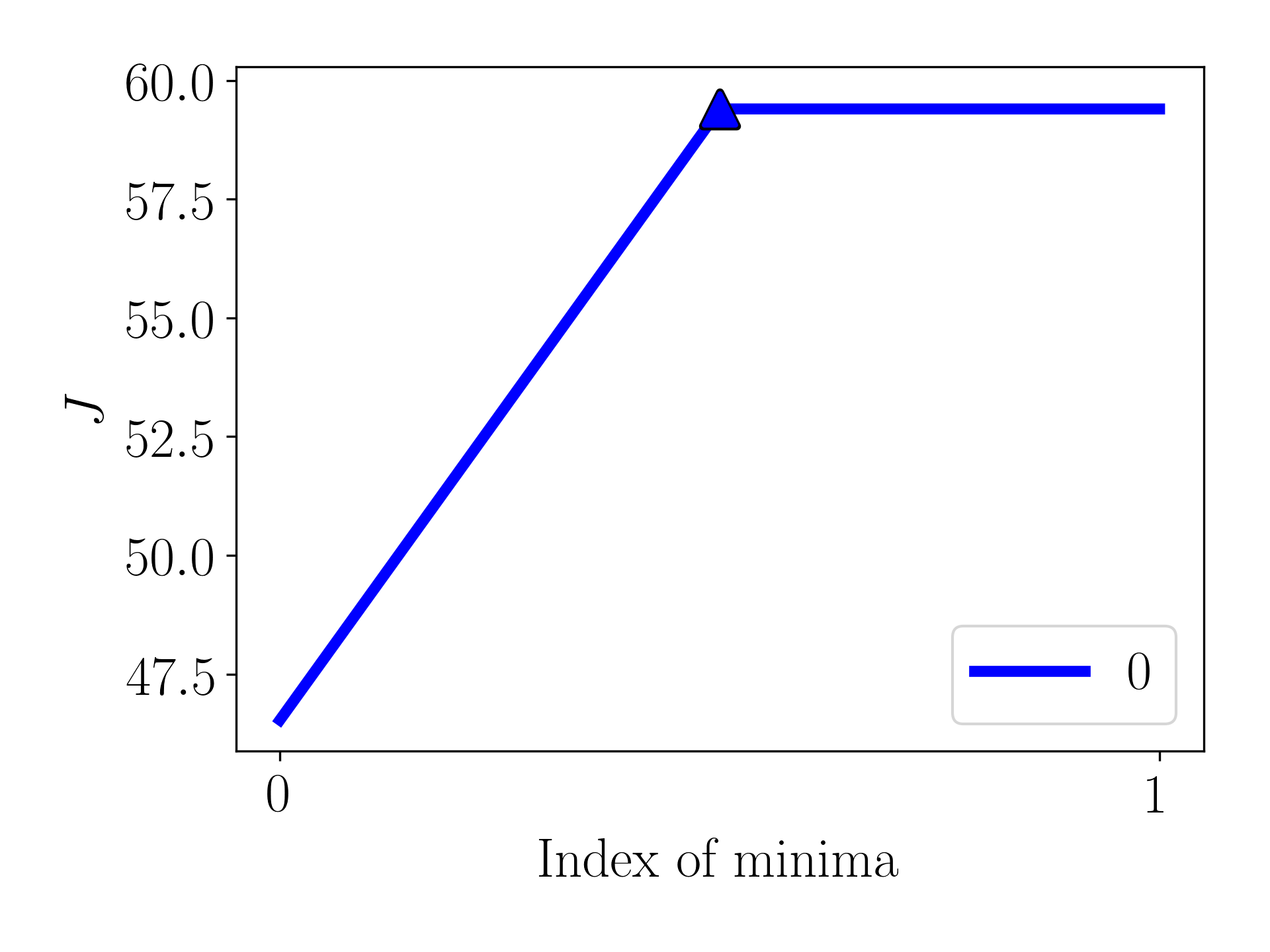}
        \caption{Outlier 1}
    \end{subfigure}%
    \begin{subfigure}[t]{0.45\textwidth}
        \centering
        \includegraphics[width=1.00\textwidth]{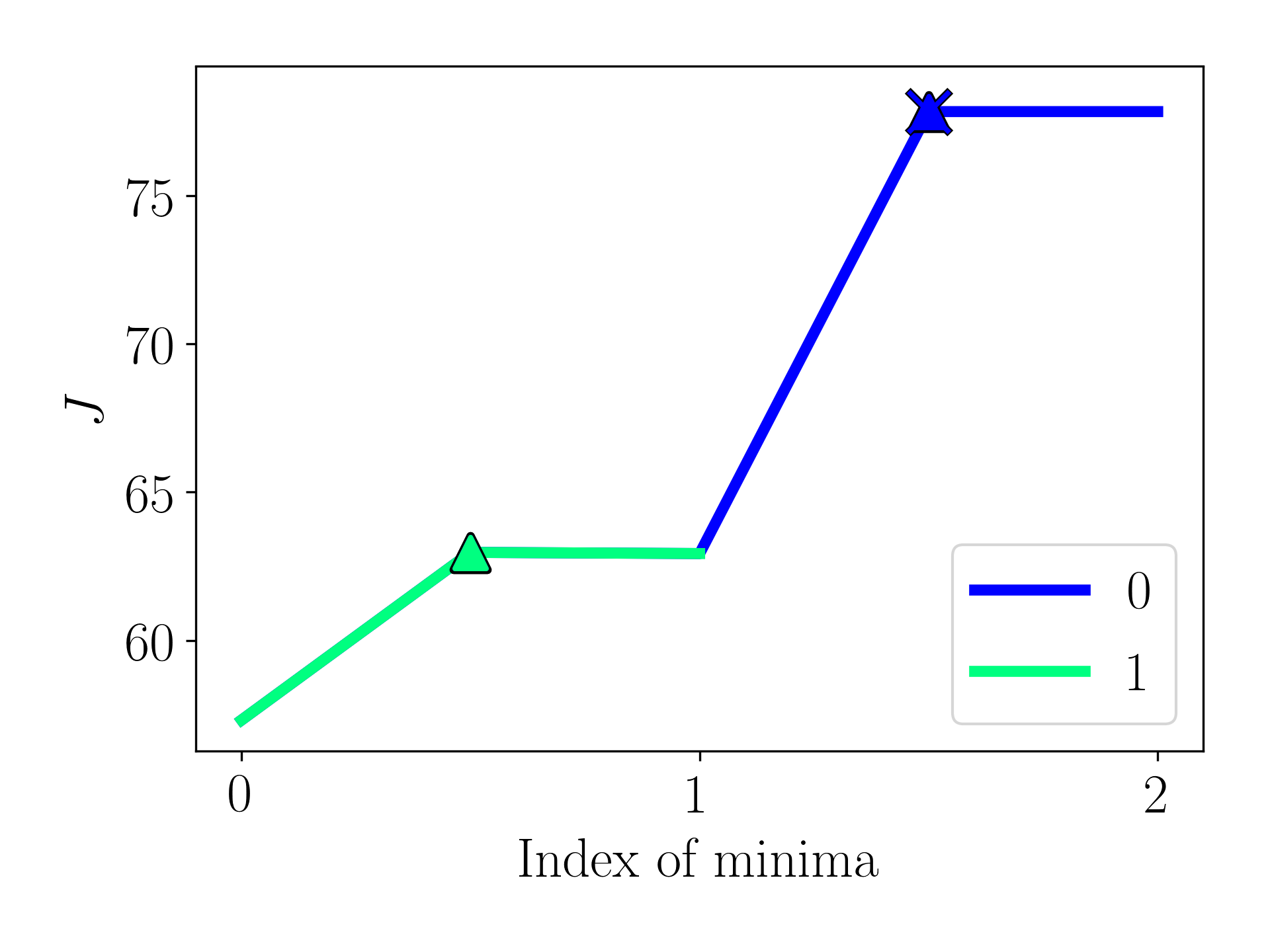}
        \caption{Outlier 2}
    \end{subfigure}\\
    \begin{subfigure}[t]{0.45\textwidth}
        \centering
        \includegraphics[width=1.00\textwidth]{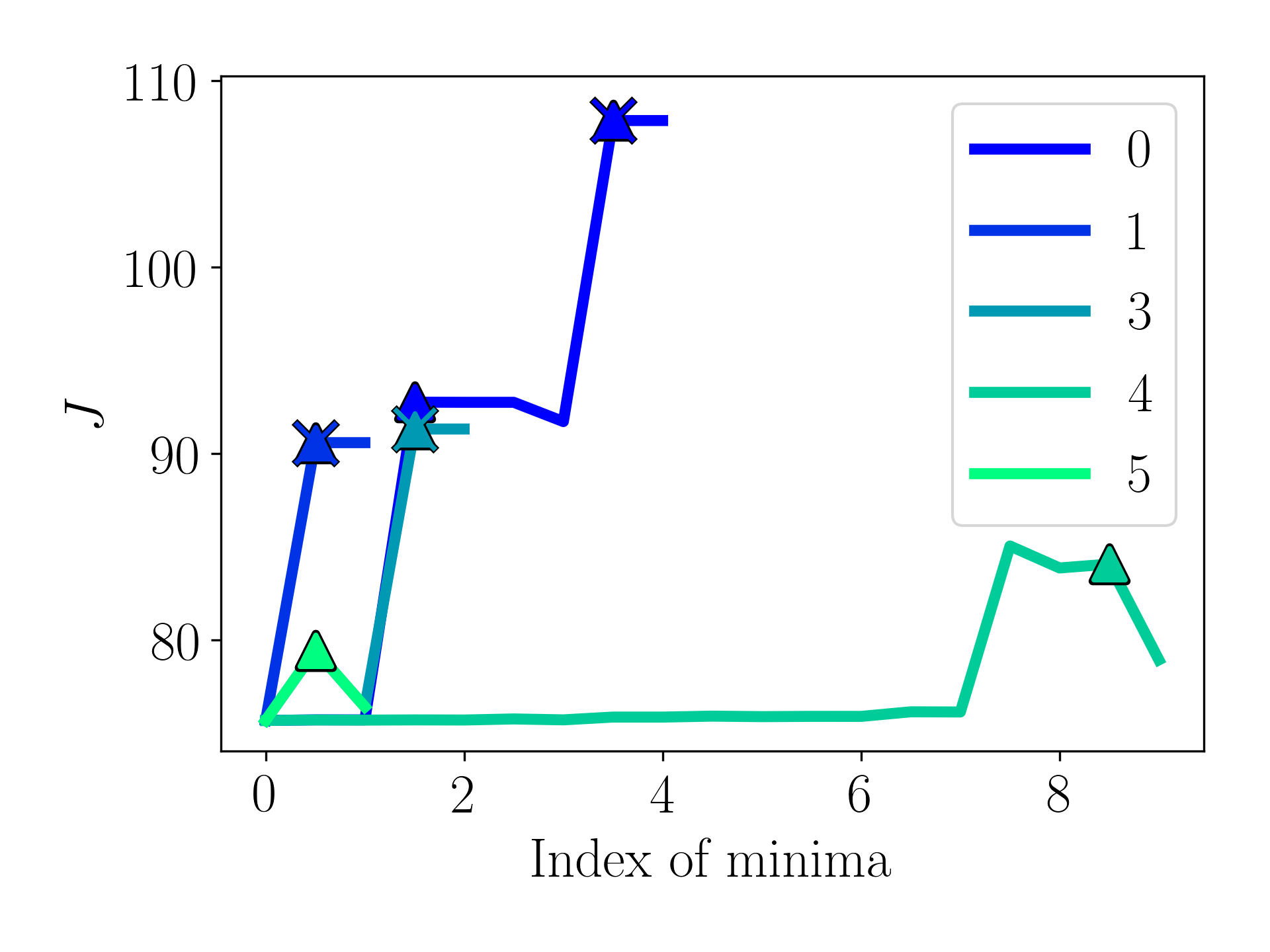}
        \caption{Outlier 3}
    \end{subfigure}%
    \begin{subfigure}[t]{0.45\textwidth}
        \centering
        \includegraphics[width=1.00\textwidth]{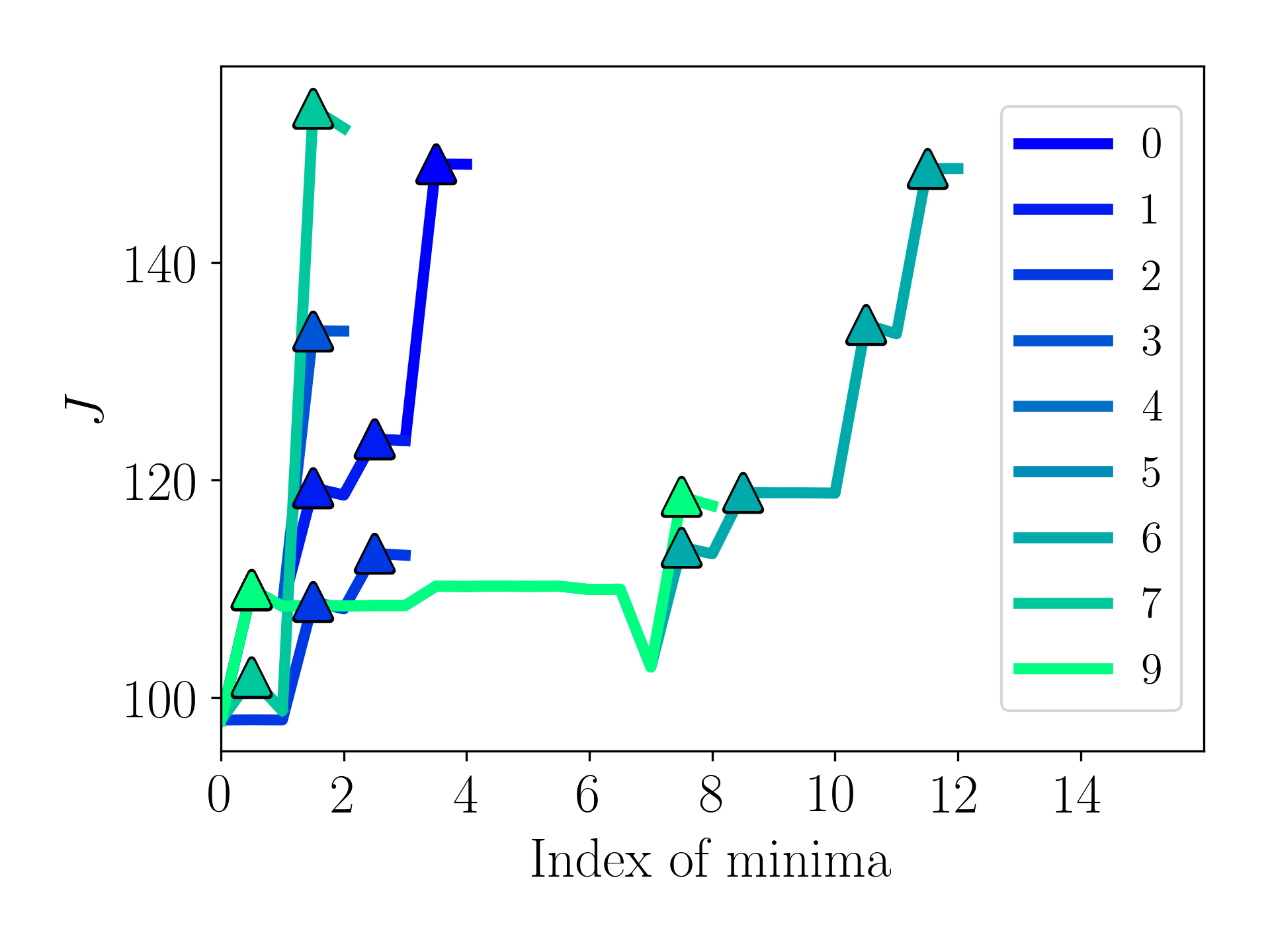}
        \caption{Outlier 4}
    \end{subfigure}\\
    \caption{The fastest paths from minima with a given structure type to the
global minimum for the \textit{Iris} dataset with $K=6$, and its variations
with outliers. All stationary points on the pathway are shown, and changes in
structure type and partition are denoted by triangles and crosses on transition
states, respectively.}
    \label{OutliersIrisPathways}
\end{figure}

For the majority of minima not within kinetic traps, the small barriers that
need to be overcome to reach the global minimum permit different structure
types to lie on a funnelled landscape. However, these small downhill barriers
do not imply that the pathways between two different structures and the global
minimum are similar. We compute the pathways from all different structure types
embedded in the funnel to the global minimum to probe the landscape
between the optimal solution and sub-optimal solutions. The cost function
profiles of the fastest pathways between each set of minima are visualised in
Fig.~\ref{OutliersIrisPathways} for the \textit{Iris} datasets, and
Fig.~S1 for the glass datasets.

The pathways reveal repeated features, such as short plateau regions separated
by sharp decreases in cost function. These features are retained as outliers
are added, and show that the solution space remains composed of many
shallow funnelled regions within a broad funnel that encompasses almost all
the space. The shallow funnels correspond to a set of minima that belong
to the same underlying partitioning or structure type and differ only in the
assignment of a small number of data points, leading to small changes in cost
function when displacing the cluster positions between them. The overall funnel
contains all these regions corresponding to different partitioning and
structure types. The sharp changes in cost function arise due to regions
between partitionings and structure types that do not support any solutions, leading to long minimisation pathways that result in a significant
reduction in $J$.

Outliers permit longer pathways within a funnel,
because there are additional alternative structure types. Similar structure types retain short
pathways, even with four outliers, but a greater proportion of transitions
within the funnel pass through more intermediate shallow funnels to attain the
global minimum. The difference in these short and long pathways are highlighted
in Figs.~S2 and S3
for the \textit{Iris} dataset with four outliers.

\begin{figure}
    \centering
    \begin{subfigure}[t]{0.33\textwidth}
        \centering
        \includegraphics[width=1.00\textwidth]{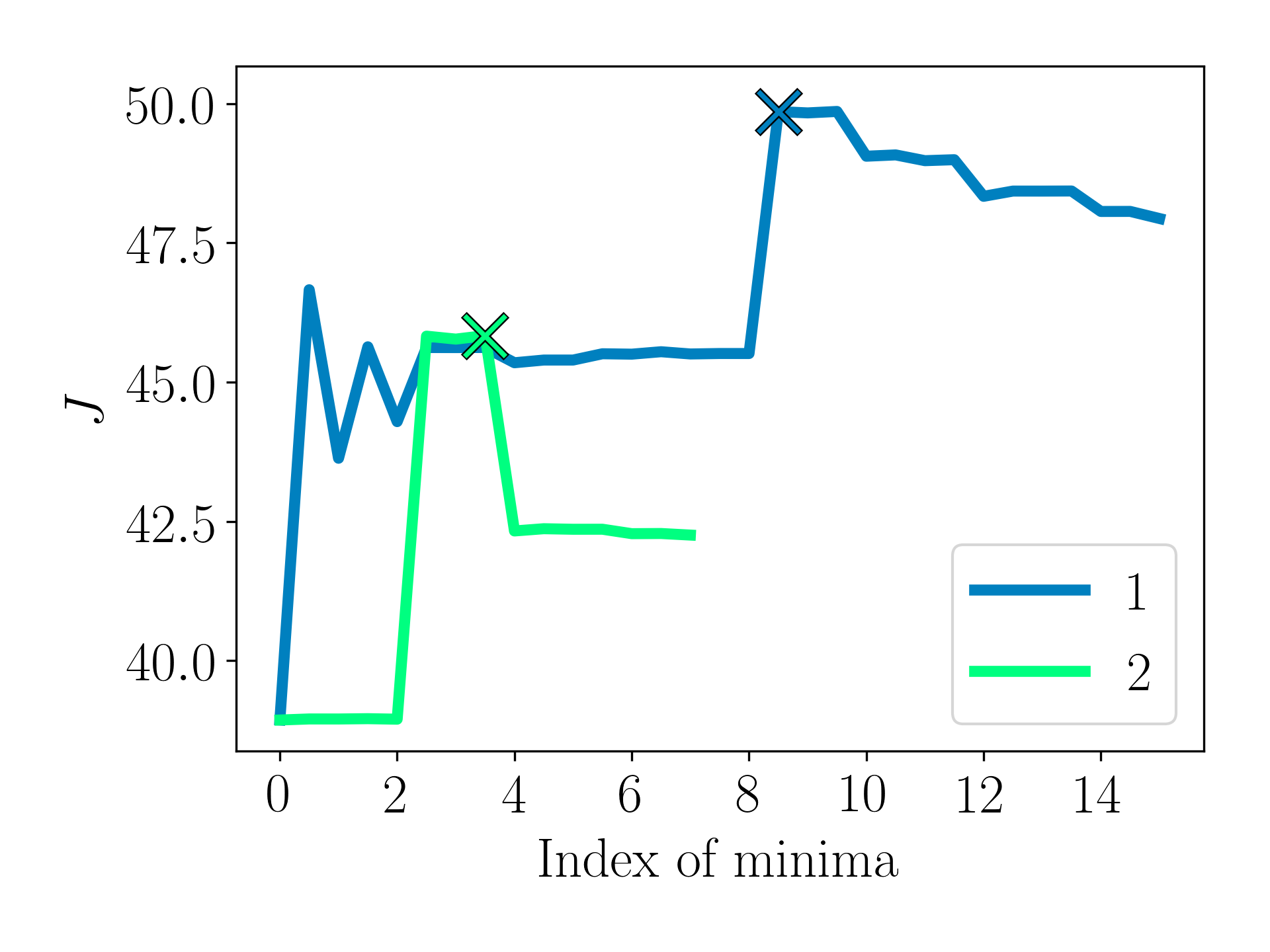}
        \caption{Original}
    \end{subfigure}%
    \begin{subfigure}[t]{0.33\textwidth}
        \centering
        \includegraphics[width=1.00\textwidth]{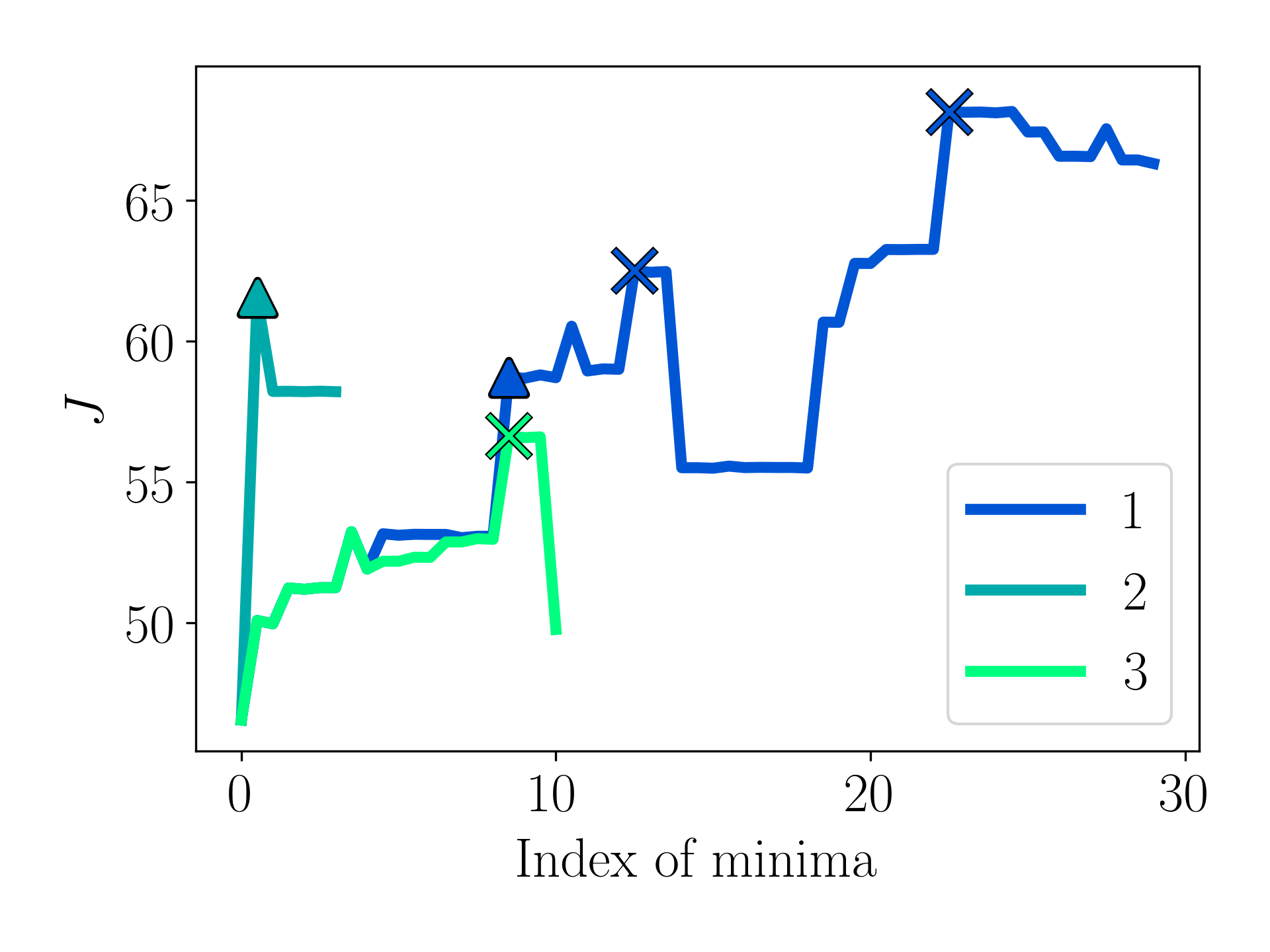}
        \caption{Outlier 1}
    \end{subfigure}%
    \begin{subfigure}[t]{0.33\textwidth}
        \centering
        \includegraphics[width=1.00\textwidth]{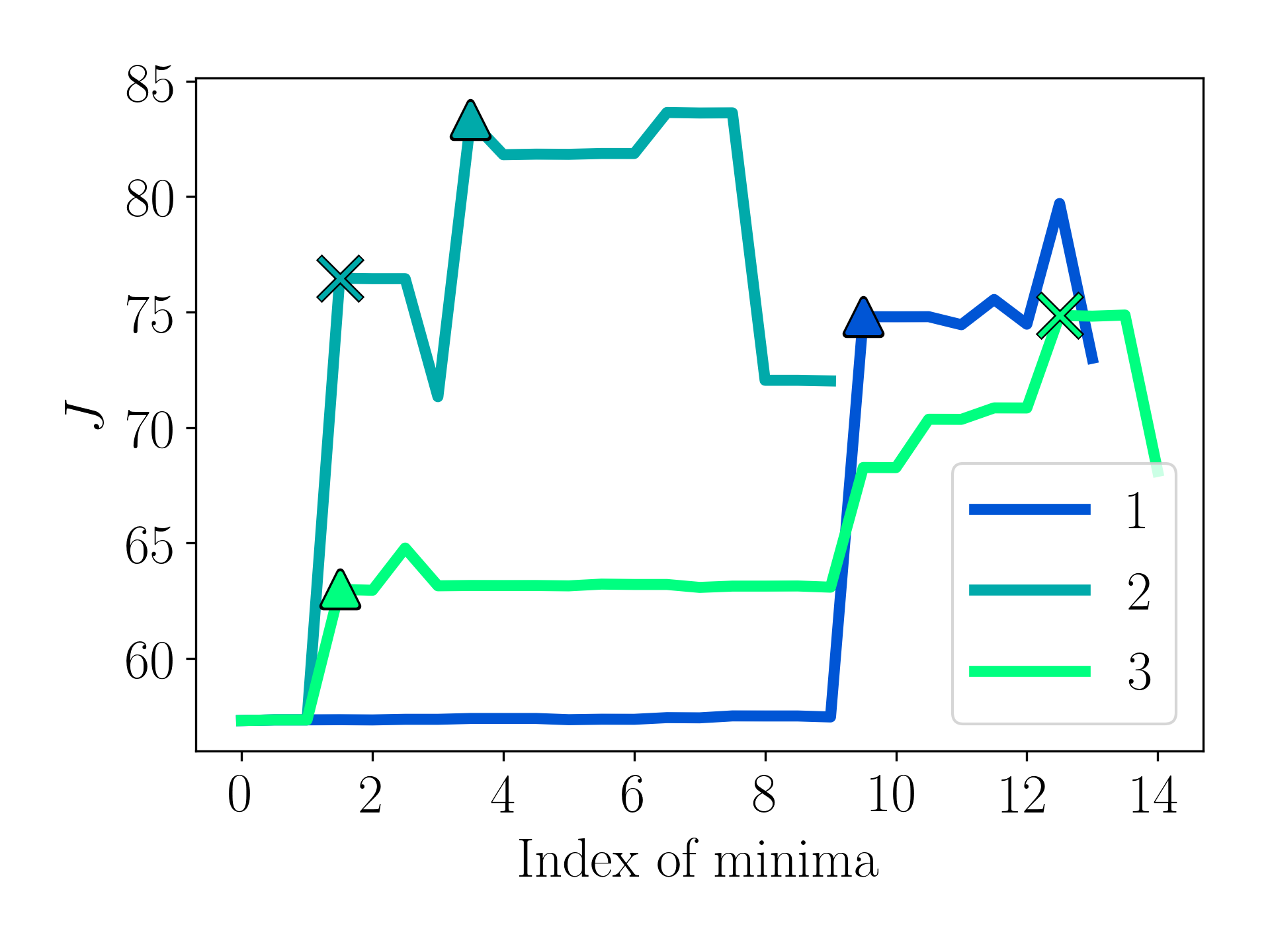}
        \caption{Outlier 2}
    \end{subfigure}\\
    \begin{subfigure}[t]{0.33\textwidth}
        \centering
        \includegraphics[width=1.00\textwidth]{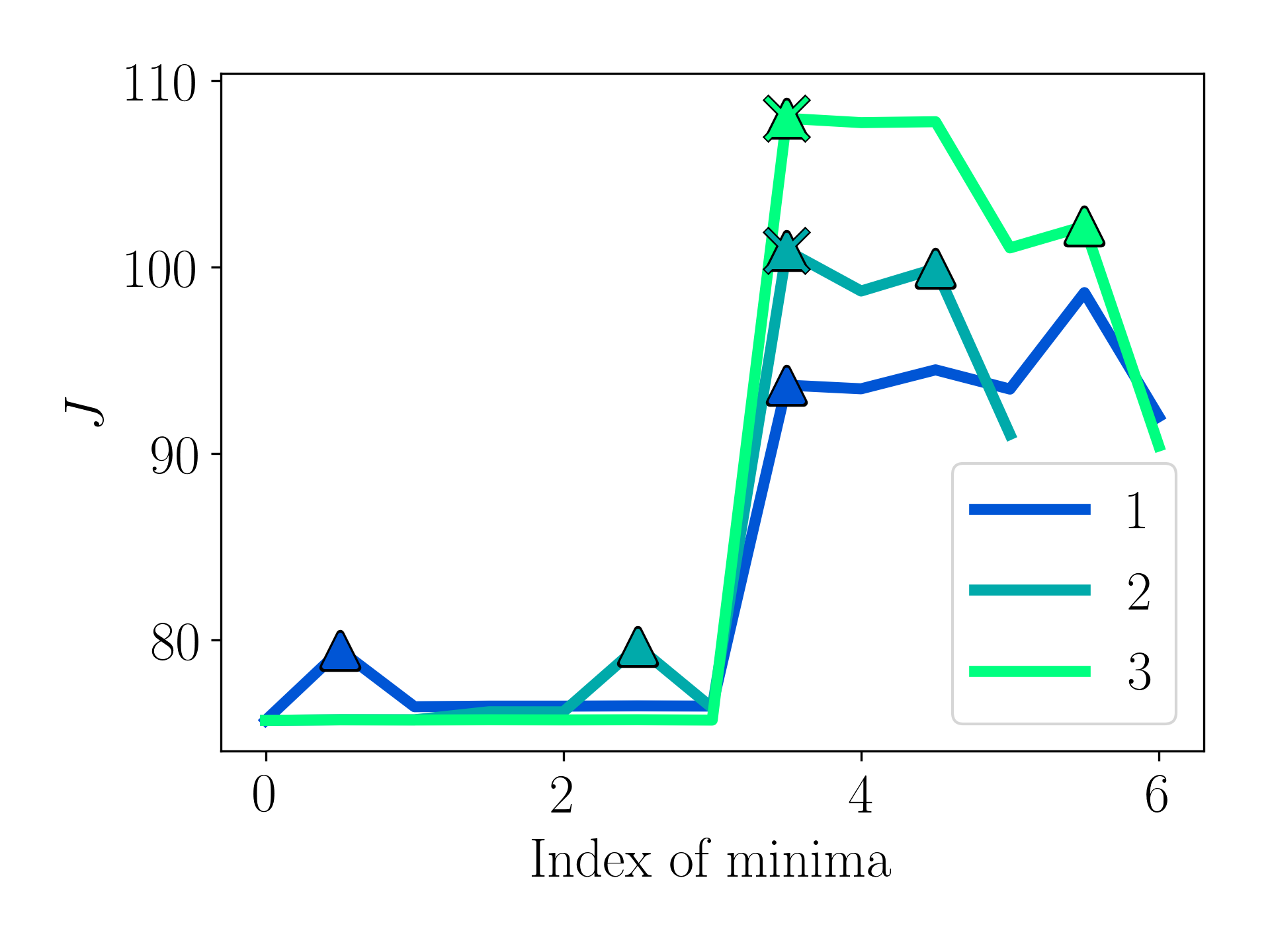}
        \caption{Outlier 3}
    \end{subfigure}%
    \begin{subfigure}[t]{0.33\textwidth}
        \centering
        \includegraphics[width=1.00\textwidth]{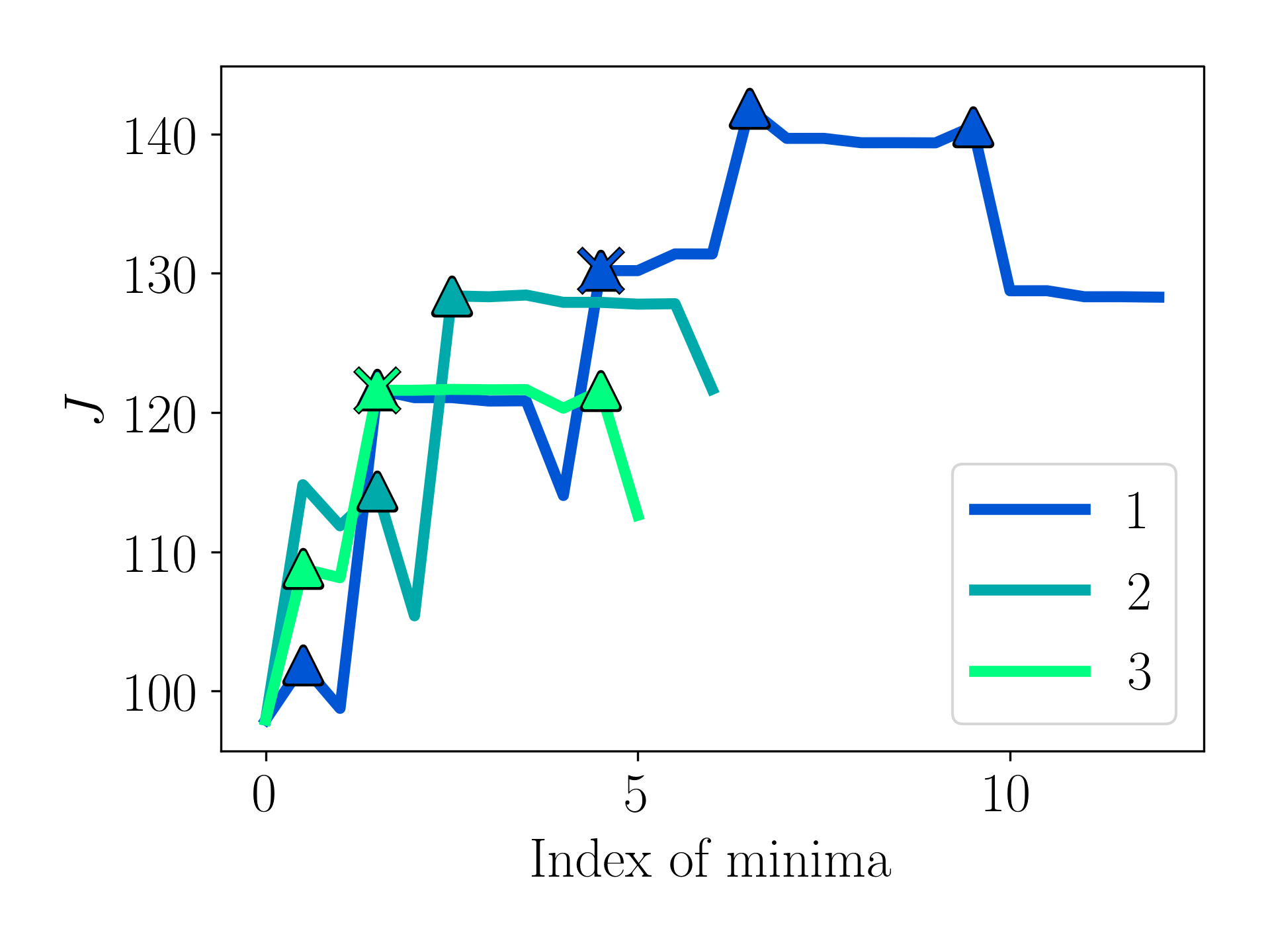}
        \caption{Outlier 4}
    \end{subfigure}\\
    \caption{The fastest paths between minima corresponding to the kinetic
traps given in Fig.~S5 and the global minimum for the
\textit{Iris} dataset with $K=6$, and variants with additional outliers. Paths
are represented by minima, at integer values along the horizontal axis, and the
transition states between them. These minima and transition states are
connected by straight lines to guide the eye. Changes in structure type are
denoted by triangles, and changes in partition by crosses, placed on the
transition states that connect the two differing minima.}
    \label{OutliersIrisKTPathways}
\end{figure}

For minima in kinetic traps there are much more significant barriers to the
global minimum along the pathways, as shown in
Fig.~\ref{OutliersIrisKTPathways} for the \textit{Iris} landscape, and
Fig.~S4 for the glass dataset. These large barriers
are associated with changes in partition and structure type.
Intermediate regions of the pathways still show shallow regions that
indicate rearrangement within a partition or structure type to allow for the
larger structural transitions. The glass pathways are much shorter than those
for the \textit{Iris} dataset, and have fewer large barriers, which is likely
due to the selected kinetic traps having the same structure type as the global
minimum for all but the Outlier 3 landscape. The pathways on other landscapes
still contain large barriers, and these are likely due to changes in partition,
which we cannot diagnose straightforwardly.

\subsubsection{Clustering solution properties}

\begin{figure}
    \centering
    \begin{subfigure}[t]{0.47\textwidth}
        \centering
        \includegraphics[width=1.00\textwidth]{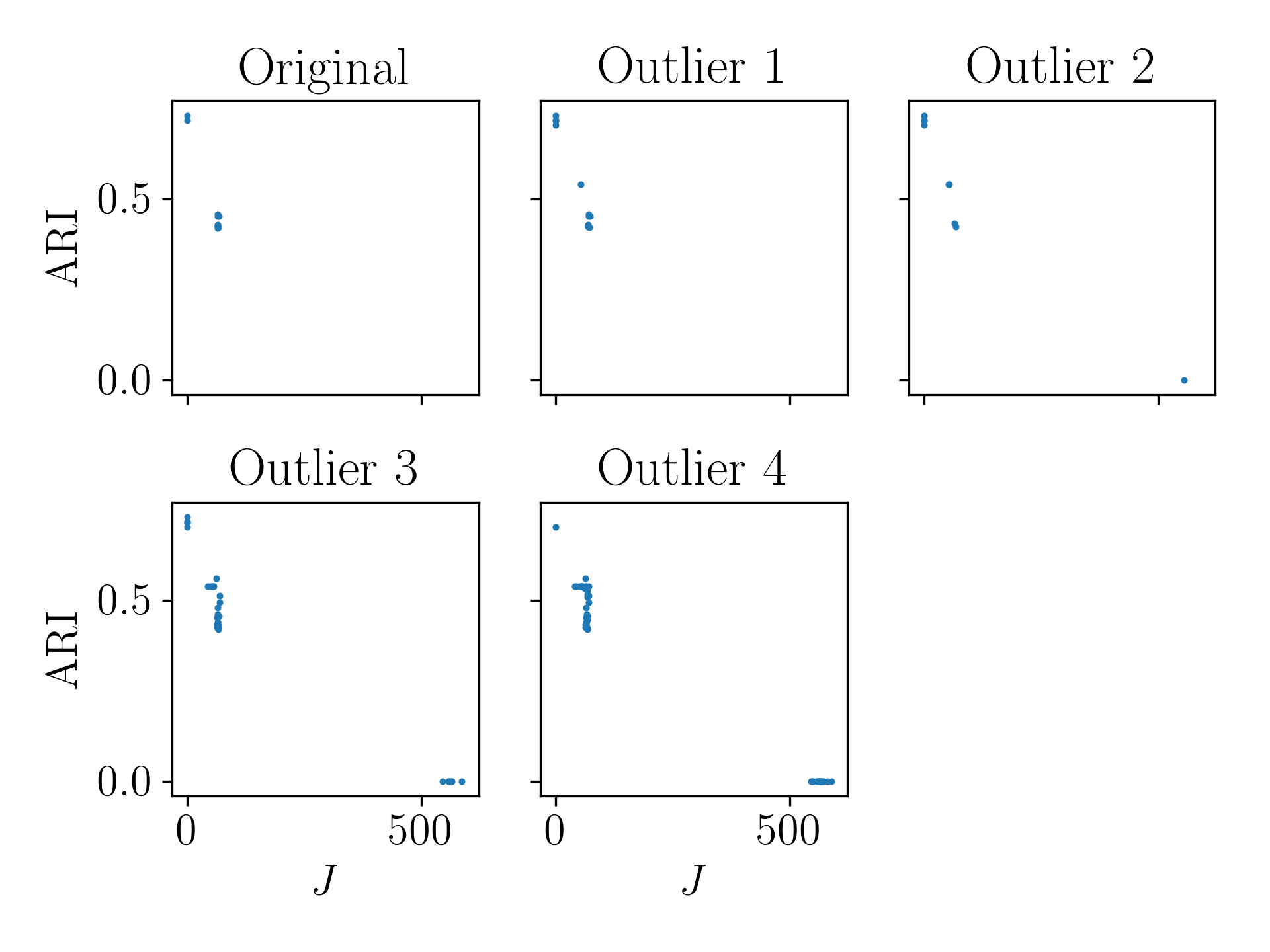}
        \caption{\textit{Iris}}
    \end{subfigure}%
    \begin{subfigure}[t]{0.47\textwidth}
        \centering
        \includegraphics[width=1.00\textwidth]{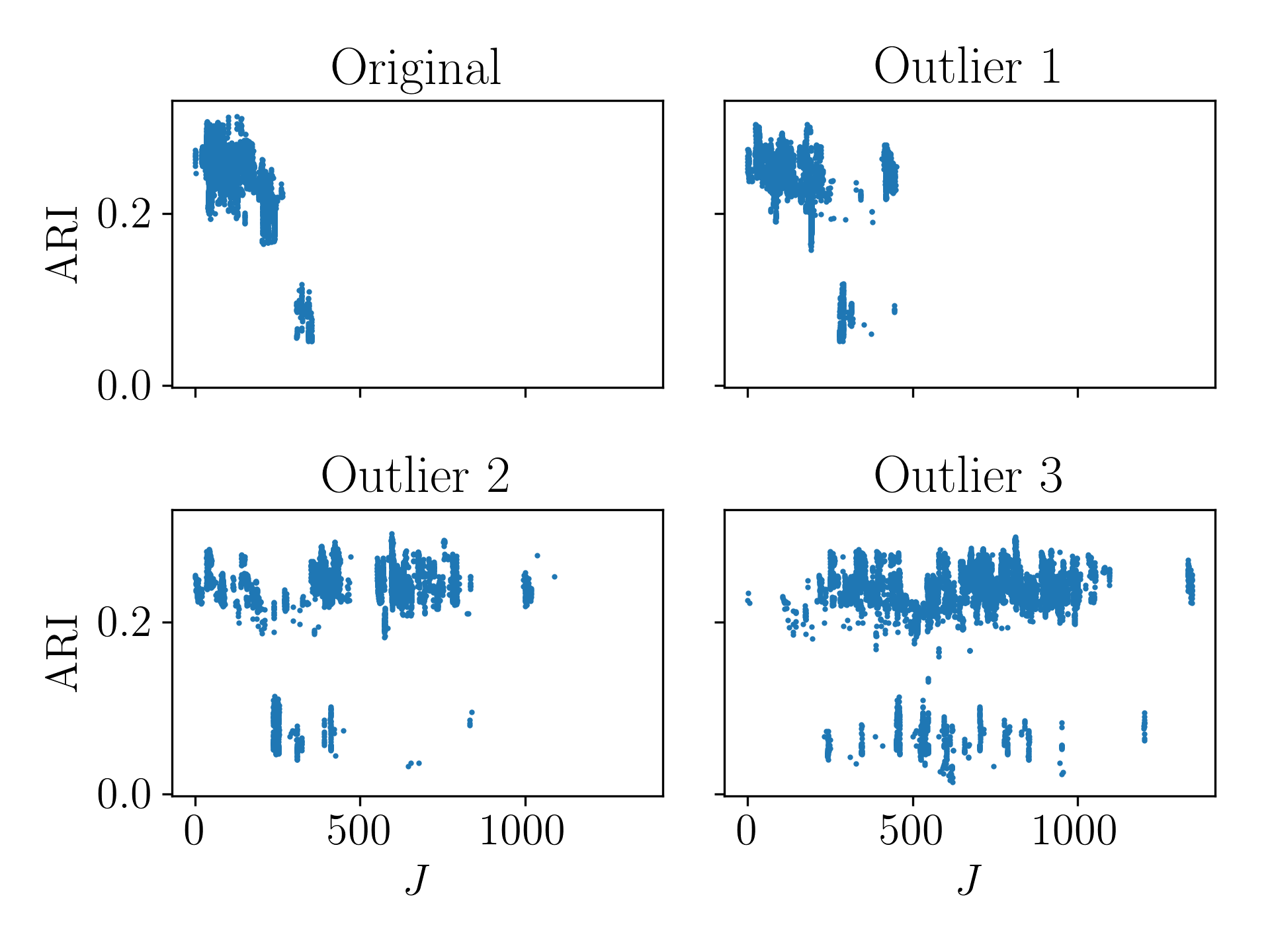}
        \caption{Glass}
    \end{subfigure}\\
    \caption{The distribution of clustering solution accuracy for the glass and
\textit{Iris} dataset ($K=3$). The accuracy was quantified using the adjusted
Rand index (ARI) between each clustering and the ground truth labels.}
    \label{OutliersDiversityDistributions}
\end{figure}

For a complete picture of the solution space it is important to determine the
properties of clustering solutions, such as accuracy. If minima lower in cost
function do not have higher accuracy then global optimisation may not
produce the best clusterings. The cost function favours solutions that have low
within-cluster variance, which may not necessarily correlate with high
accuracy. Accuracy is here computed as the ARI between the ground truth cluster
labels and the cluster labels of a given minimum. The accuracy distribution of
the minima is given in Fig.~\ref{OutliersDiversityDistributions}, and the
disconnectivity graphs coloured by accuracy are given in
Figs.~S7 and S8.

For the \textit{Iris} dataset with $K=3$ the cost function correlates strongly
with clustering accuracy, due to the small amount of overlap. The highest
achievable accuracy is 0.730, as the $K$-means algorithm is unable to separate
the \textit{Versicolor} and \textit{Virginica} clusters. There is little
reduction in the best accuracy on addition of increasing outliers, because the
global minimum accommodates the outliers by appending them to existing
clusters (Fig.~S9), which results in little change in
the cluster boundaries from the original dataset. However, for landscapes of
two outliers or more we see low-valued solutions that correspond to one
outlier cluster; the \textit{Iris} dataset can be well represented with
two clusters due to the overlap of \textit{Versicolor} and \textit{Virginica}
data points.

The accuracy distribution of the glass identification dataset exhibits
different behaviour. This dataset contains significant overlap and, although
the accuracy still generally decreases with increasing cost function for the
original glass dataset, the $K$-means cost function is less appropriate for
clustering the dataset. The relationship between the cost function value and
accuracy becomes weaker with a growing number of outliers, and at three
outliers the accuracy has no relation to the cost function. Both the highest
attainable accuracy and the accuracy of the global minimum decrease with
additional outliers. The most accurate clustering moves to a higher cost
function value, relative to the global minimum, and the most accurate
clustering is not the global minimum in any landscape. With additional outliers
the most accurate solutions appear repeatedly at different values of the cost
function.

For the \textit{Iris} dataset with $K=6$ we cannot evaluate the accuracy, but
we can rationalise the significant kinetic traps in the original \textit{Iris}
landscape by equating them to different partitionings of the six $K$-means
clusters amongst the three underlying clusters.
Partitionings remain useful to understand kinetic traps, but as the number of
outliers increases the structure type exerts more control over cost function
topography. Partitionings become less clear, as many clusters now incorporate
outliers, and a growing number of kinetic traps can be rationalised by a
difference in structure type, and not partition. Disconnectivity graphs
coloured by partitioning are given for \textit{Iris} datasets with both cluster
numbers in Fig.~S10, and the different structure
types of the global minima are shown in Fig.~S11.

\subsection{Rate-based clustering comparison}

We propose that the rate computed between clustering solutions can be used as a
measure of their distinctness. All current external metrics for comparing
clusterings use only the initial and final configurations, but the rates depend on the
path between them, including information about the number of
intermediate minima and the barriers for cluster displacements. Hence this
metric encodes the cumulative change in cluster position and assignment,
rather than just the difference between endpoints. It is
related to the difficulty in navigating the landscape between two clusterings, providing valuable insight into global optimisation
performance for $K$-means. Moreover, rates implicitly weight the importance of
data points by the change they induce in the cost function, which allows a
robust measure of cluster distinctness, even in the presence of dataset
outliers.

\begin{figure}
    \centering
    \includegraphics[width=0.5\textwidth]{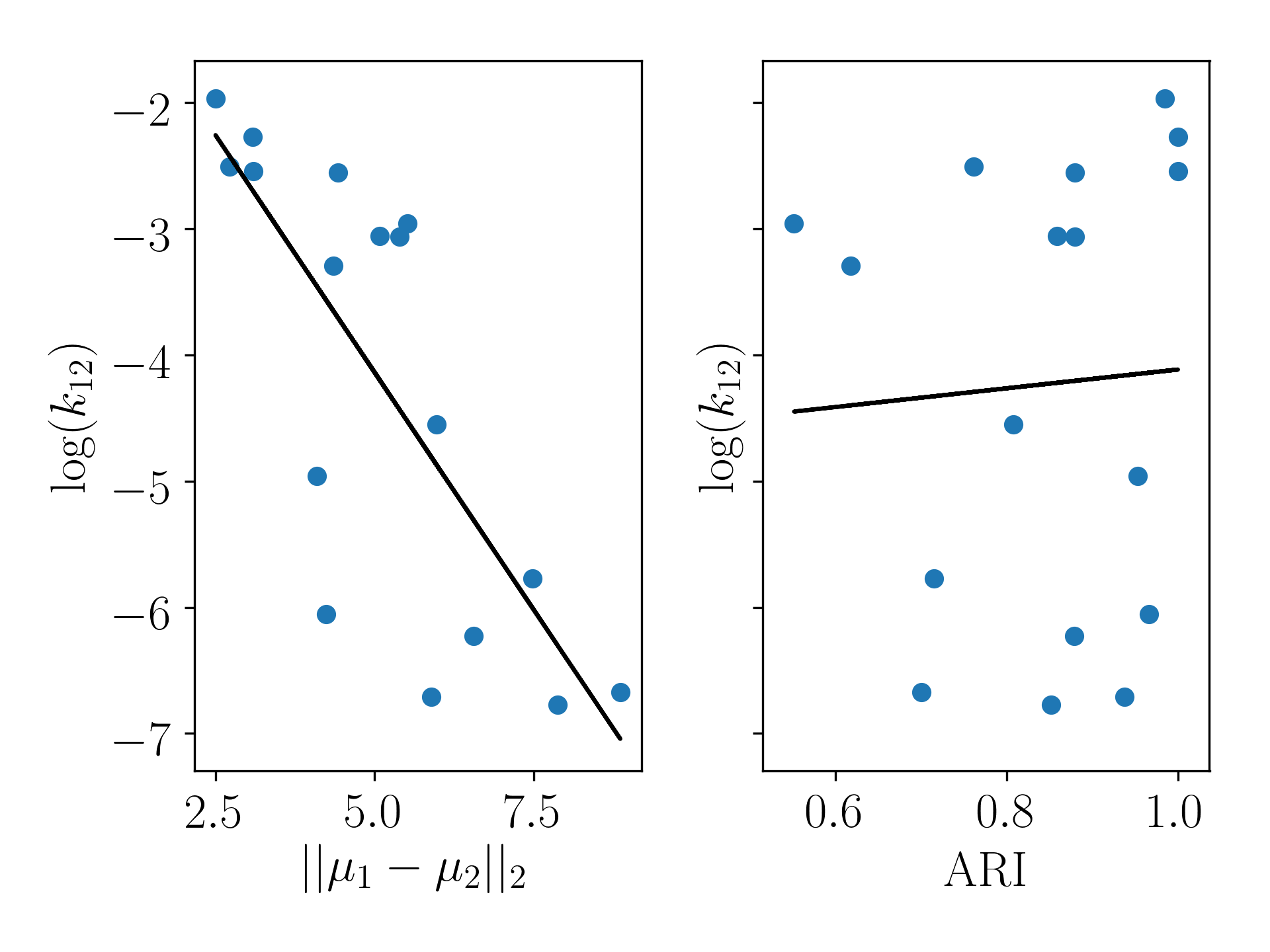}
    \caption{Correlation between rates, $k_{12}$, and two other external
metrics for cluster comparison: the distance between cluster centres (left) and
the ARI (right). The pairs of clusterings are different structure types of the
\textit{Iris} datasets containing outliers. The line of best fit is the solid
black line.}
    \label{ComparativeClustering}
\end{figure}

We examine the correlation between rates, cluster position (distance), and
cluster assignment (ARI) measure in Fig.~\ref{ComparativeClustering}. The
clustering similarity measures were computed between different structure types
in all outlier landscapes for the \textit{Iris} dataset with six clusters. The
structure types differ in the assignment of outliers, which can be challenging
for cluster assignment-based metrics because a few data point label changes produce
a large effect. Consequently, we observe a weak positive correlation with
the ARI. The distance captures the large distorting effect of the outliers on
cluster positions, and we observe a strong correlation with the rate, which therefore
also captures the importance of individual data points for cluster
distinctness. However, the three metrics probe different features of the
clustering distinctness, and there are many examples where the pathway-based
measure does not simply relate to the distance. We illustrate the comparative
measures alongside the clusterings, in three of the four dimensions, for the
same set of minima in Fig.~S12.

We compute the rates for various transitions that will affect global optimisation, such
as between different structure types and for kinetic traps to the global minimum.
These rates reflect the distinctness of
various clusterings from the global minimum, and hence report on the difficulty of
global optimisation from each starting point. The rates between different
structures embedded in the funnel can vary significantly, even for minima with
comparable cost functions, as shown in Tables~\ref{OutliersIrisRates} and
SIV. The rates largely correlate with the number of
different structure types (or more rarely partitions) that must be visited on
the way to reaching the global minimum. The rates remain similar across all
outlier numbers when the structure types involve a small number of changes, as
the similarity of such structures is retained. However, the rates can be much
slower in landscapes with more outliers, due to the greater difference in
arrangements of distant structure types. The rearrangement of multiple outliers
involves more transition states and intermediate minima, and the
rates therefore decrease. The reduction in slowest rate with outliers is much more
pronounced for the glass dataset. Furthermore, the rates remain remarkably
similar between all structure types, without the large preference for
transitions between structures of the same type seen in the \textit{Iris}
dataset.

\begin{table}[]
\caption{Transition rates (reduced units) between the set of minima of a given structure type
to the global minimum, calculated at an effective temperature of $T=1.0$ (in 
reduced units consistent with the cost function). Rates
are calculated for the \textit{Iris} dataset with $K=6$, and those datasets
with additional outliers. The dot indicates the structure type of the global
minimum.}
\label{OutliersIrisRates}
\centering
\begin{tabularx}{\textwidth}{Y|Y|Y|Y|Y}
Type      &   Outlier 1 & Outlier 2 & Outlier 3 & Outlier 4 \\ \hline
0 & $3.11 \times 10^{-3}$ & $8.82 \times 10^{-4}$ & $2.82 \times 10^{-5}$ & $1.68 \times 10^{-7}$ \\
1 & $\cdot$ & $1.08 \times 10^{-2}$ & $2.78 \times 10^{-3}$ & $1.94 \times 10^{-7}$ \\
2 &  & $\cdot$ & $\cdot$ & $8.81 \times 10^{-7}$ \\
3 &  &  & $8.68 \times 10^{-4}$ & $5.90 \times 10^{-7}$ \\
4 &  &  & $5.11 \times 10^{-4}$ & $5.34 \times 10^{-3}$ \\
5 &  &  & $2.85 \times 10^{-3}$ & $1.10 \times 10^{-5}$ \\
6 &  &  &  & $2.12 \times 10^{-7}$ \\
7 &  &  &  & $1.11 \times 10^{-3}$ \\
8 &  &  &  & $\cdot$ \\
9 &  &  &  & $1.70 \times 10^{-6}$ \\
\end{tabularx}
\end{table}

The rates between kinetic traps and the global minimum are given in
Table~\ref{OutliersEscapeRatesIris} and Table~SV
for the \textit{Iris} and glass datasets, respectively. These rates generally
decrease as the number of outliers increases. The escape rates from different
kinetic traps within the same landscape exhibit large fluctuations as a
result of the specific features. 
However, there remains an overall
reduction in escape rates and, consequently, escaping kinetic traps becomes
more difficult as the number of outliers increases.

\begin{table}
\caption{Escape rates (reduced units) from kinetic traps on the solution landscape for the
\textit{Iris} dataset with $K=6$, and variations with additional outliers. The
rates are calculated from the set of minima at the bottom of the kinetic trap
to the global minimum, at an effective temperature of $T=1.0$ (reduced units). The selected
kinetic traps are given in Fig.~S2. The geometric mean
rate for each landscape is given in the final row.}
\label{OutliersEscapeRatesIris}
\centering
\begin{tabularx}{\textwidth}{Y|Y|Y|Y|Y|Y}
Kinetic trap & Original     &   Outlier 1 & Outlier 2 & Outlier 3 & Outlier 4 \\ \hline
1 & $3.65 \times 10^{-4}$ & $2.25 \times 10^{-5}$ & $2.64 \times 10^{-4}$ & $3.23 \times 10^{-4}$ & $7.25 \times 10^{-7}$ \\
2 & $1.74 \times 10^{-5}$ & $1.02 \times 10^{-3}$ & $6.55 \times 10^{-8}$ & $3.26 \times 10^{-6}$ & $1.12 \times 10^{-6}$ \\
3 & - & $2.66 \times 10^{-5}$ & $2.20 \times 10^{-5}$ & $2.26 \times 10^{-9}$ & $8.57 \times 10^{-7}$ \\ \hline
Mean & $7.97 \times 10^{-5}$ & $8.48 \times 10^{-5}$ & $7.25 \times 10^{-6}$ & $1.34 \times 10^{-6}$ & $8.86 \times 10^{-7}$ \\
\end{tabularx}
\end{table}

\section{Conclusions}

Successful $K$-means clustering depends on two factors: efficient location of the global
minimum (or low-valued minima), and the appropriateness of the cost function,
so that low-valued minima are the most accurate. The efficiency of global
optimisation is encoded in the organisation of the solution space. We
find that these landscapes have a funnelled structure for the datasets we consider with
any number of outliers. The degree of funnelling, as quantified using a
frustration metric, increases with the number of outliers.

Kinetic analysis reveals that the overall funnel seen in each landscape is
composed of many different shallow regions that correspond to different types
of structure. These shallow regions of space involve small barriers between
different structure types, and they are separated by cluster solution space that
does not contain any minima. Therefore, there can be large changes in cost
function between these regions associated with long pathways. The number of
such structure types increases with the number of outliers, which can result in
much longer pathways for navigation of the funnel.

We have introduced kinetic analysis of the $K$-means landscapes 
as a novel clustering similarity measure. The rate extends external
clustering metrics by incorporating information about pathways 
between clustering solutions, rather than just the endpoints. Such a
measure can quantify the cumulative change in cluster position and assignment,
which may be more relevant for rationalising global optimisation performance. We
compute rates within the \textit{Iris} and glass datasets to analyse the
similarity between different regions of the solution space. We observe that
high similarity is conserved between adjacent structure types across all
outlier landscapes, but the lowest similarity decreases as the number of
outliers increases.

A funnelled surface is usually favourable for global optimisation in physical
systems. However, the rates illustrate that it becomes more challenging to both escape kinetic traps, and relax within the main funnel, to locate the global minimum with increasing outliers. There is a sharp increase in the number of ways in which outliers can be accommodated into clusterings as the number of outliers increases, and this increase leads to longer pathways between different regions of space. Therefore, although movement between adjacent structure types remains fast, it becomes more challenging to move about the main funnel between a greater number of structure types. For successful global optimisation runs from distinct starting configurations may be helpful, and an efficient moveset
would require both local perturbations, which optimise cluster boundaries, and
larger changes between different structure types by exchanging outliers.
However, with an appropriate moveset the single-funnel solution space should
allow the global minimum to be located quite efficiently.

The correlation between cost function and accuracy is degraded by the increased
number of outliers. The outliers perturb the cluster centre positions, leading
to reduced accuracy of the low-valued solutions. Additionally, the increase in
the number of structure types produces greater variation in the accuracy throughout
the cost function range. Therefore, for datasets containing multiple outliers,
obtaining the global minimum does not necessarily give the most accurate
clustering solution that $K$-means can support. Without a clear way to
distinguish accurate solutions in cost function it is important to use an
internal metric, such as the silhouette coefficient,\cite{Rousseeuw1987} to
independently rank them. However, except for the case of
three outliers in the highly-overlapping glass dataset, we observe that the
low-valued region does contain minima of comparable accuracy to the most
accurate clustering. Hence locating low-valued minima is useful, but it
becomes essential to generate an ensemble of clustering solutions, which can
capture the many possible arrangements of the various
outliers.\cite{Alqurashi2019}

\section{Supplementary material}

See the supplementary material for additional methodological details (S-I and
S-III), a description of the datasets used in this work (S-II), and additional
figures referenced in this work (S-IV).

\section{Acknowledgements}

This work was supported by an EPSRC knowledge transfer fellowship.
DJW gratefully acknowledges an
International Chair at the Interdisciplinary Institute for Artificial
Intelligence at 3iA Cote d'Azur, supported by the French government [grant
number ANR-19-P3IA-0002],
which has provided interactions that furthered the present research project.

\bibliographystyle{unsrt}

\bibliography{Outliers_References}

\end{document}



\title[]{Supplementary material -- Evolution of $K$-means solution landscapes with the addition of dataset outliers and a robust clustering comparison measure for their analysis}

\author{L. Dicks}
\affiliation{ 
Yusuf Hamied Department of Chemistry, Lensfield Road, Cambridge CB2 1EW, United Kingdom
}
\affiliation{
IBM Research Europe, Hartree Centre, Sci-Tech Daresbury, United Kingdom
}
\author{D. J. Wales}%
 \email{dw34@cam.ac.uk}
\affiliation{ 
Yusuf Hamied Department of Chemistry, Lensfield Road, Cambridge CB2 1EW, United Kingdom
}

\date{\today}

\maketitle

\section{Constructing disconnectivity graphs} \label{DGconstruction}

All clustering solutions (local minima) are represented by vertical lines originating at their
corresponding cost function value on the vertical axis. For any given cost function threshold, the
minima can be separated into groups connected by barriers below the
threshold. The overall barrier between any two minima is given by the cost function at
the highest transition state on a path between them, which many be one, or
more, transition states in length. Starting from a cost function value above
the highest-valued minimum we reduce the threshold by regular intervals and
separate the minima into disjoint sets, known as superbasins.\cite{Becker1997}
The vertical lines corresponding to minima join when the cost function
threshold is greater than the barrier between them. The horizontal axis is
chosen to arrange minima to provide a clear representation of the topography.

\section{\label{sec:level1} Data preparation}

For the \textit{Iris} dataset, the outlier coordinates lie outside the range of the original data in at least half the features. In the glass datasets, the outliers lie outside the range of the original data in every dimension. 

\begin{table}[!htb]
\centering
\caption{The position of outliers added to the \textit{Iris} dataset. The values of the features are all given in centimetres. The mean and range are given for the original dataset.}
\label{IrisOverlapData}
\begin{tabularx}{\textwidth}{Y|Y|Y|Y|Y}
Outlier & Sepal length & Sepal width & Petal length & Petal width \\ \hline
1 & 9.6 & 6.0 & 5.0 & 4.0 \\
2 & 2.0 & 0.1 & 0.1 & 2.5 \\
3 & 6.0 & 7.0 & 0.1 & 4.0 \\
4 & 1.0 & 4.0 & 8.0 & 0.5 \\ \hline
Mean & 5.84 & 3.05 & 3.76 & 1.20 \\
Range & 4.3--7.9 & 2.0--4.4 & 1.0--6.9 & 0.1--2.5
\end{tabularx}
\end{table}

\begin{table}[!htb]
\centering
\caption{The coordinates of the outliers added to the glass dataset. Features are provided in the same order as the data taken from the UCI machine learning data repository. The mean and range are given for the original dataset.}
\label{GlassOverlapData}
\begin{tabularx}{\textwidth}{Y|Y|Y|Y|Y|Y|Y|Y|Y|Y}
Dataset & $x_1$ & $x_2$ & $x_3$ & $x_4$ & $x_5$ & $x_6$ & $x_7$ & $x_8$ & $x_9$ \\ \hline
1 & 1.3 & 5.0 & 7.0 & 6.0 & 60.0 & 9.0 & 2.0 & 5.0 & 3.0 \\
2 & 1.7 & 23.0 & 7.0 & 7.0 & 90.0 & 9.0 & 21.0 & 5.0 & 4.0 \\
3 & 1.8 & 4.0 & 8.0 & 7.0 & 85.0 & 9.0 & 1.0 & 5.0 & 4.0 \\ \hline
Mean & 1.52 & 13.41 & 2.68 & 1.44 & 72.65 & 0.50 & 8.96 & 0.18 & 0.06 \\ 
Range & \footnotesize{1.51--1.53} & \footnotesize{10.7--17.4} & \footnotesize{0.00--4.49} & \footnotesize{0.29--3.50} & \footnotesize{69.8--75.4} & \footnotesize{0.00--6.21} & \footnotesize{5.43--16.2} & \footnotesize{0.00--3.15} & \footnotesize{0.00--0.51}
\end{tabularx}
\end{table}

\section{Comparing clusterings}

The Rand index,\cite{Rand1971} compares two partitions using the following ratio
\begin{equation}
RI = \frac{TP + TN}{TP + TN + FP + FN},
\end{equation}
where $TP,TN,FP,FN$ are the number of true positives, true negatives, false
positives and false negatives, respectively, evaluated
over all pairs of data points within a dataset.
A true positive occurs when the pair of data points are placed in the same
cluster in both the reference and test partitions, and a true negative when
they are placed in different clusters in both partitions. False positives and
negatives arise when one clustering assigns the two data points to the same
cluster, but not the other. A false negative is distinguished by the two data
points being assigned to the same cluster in the reference clustering, and for
a false positive the data points are in different clusters in the reference.

The Rand index takes values between 0 and 1, with 0 indicating no similarity
between the clusterings, and 1 indicating identical clusterings. The Rand index
is often close to 1, even with significant differences between the clusterings,
because there are often a large number of similar pairwise labellings that
arise at random. The Rand index does not account for the similarity expected
for random datasets, and the value of the Rand index, calculated for two random
clusterings, is greater than zero. The adjusted Rand index\cite{Hubert1985}
(ARI) accounts for cluster labels being the same by chance and provides a new
baseline, such that expected similarity of all pairwise comparisons within a
random model is approximately 0.

\clearpage

\section{Cost function topography}

\subsection{Kinetic analysis}

\begin{figure}
    \centering
    \begin{subfigure}[t]{0.45\textwidth}
        \centering
        \includegraphics[width=1.00\textwidth]{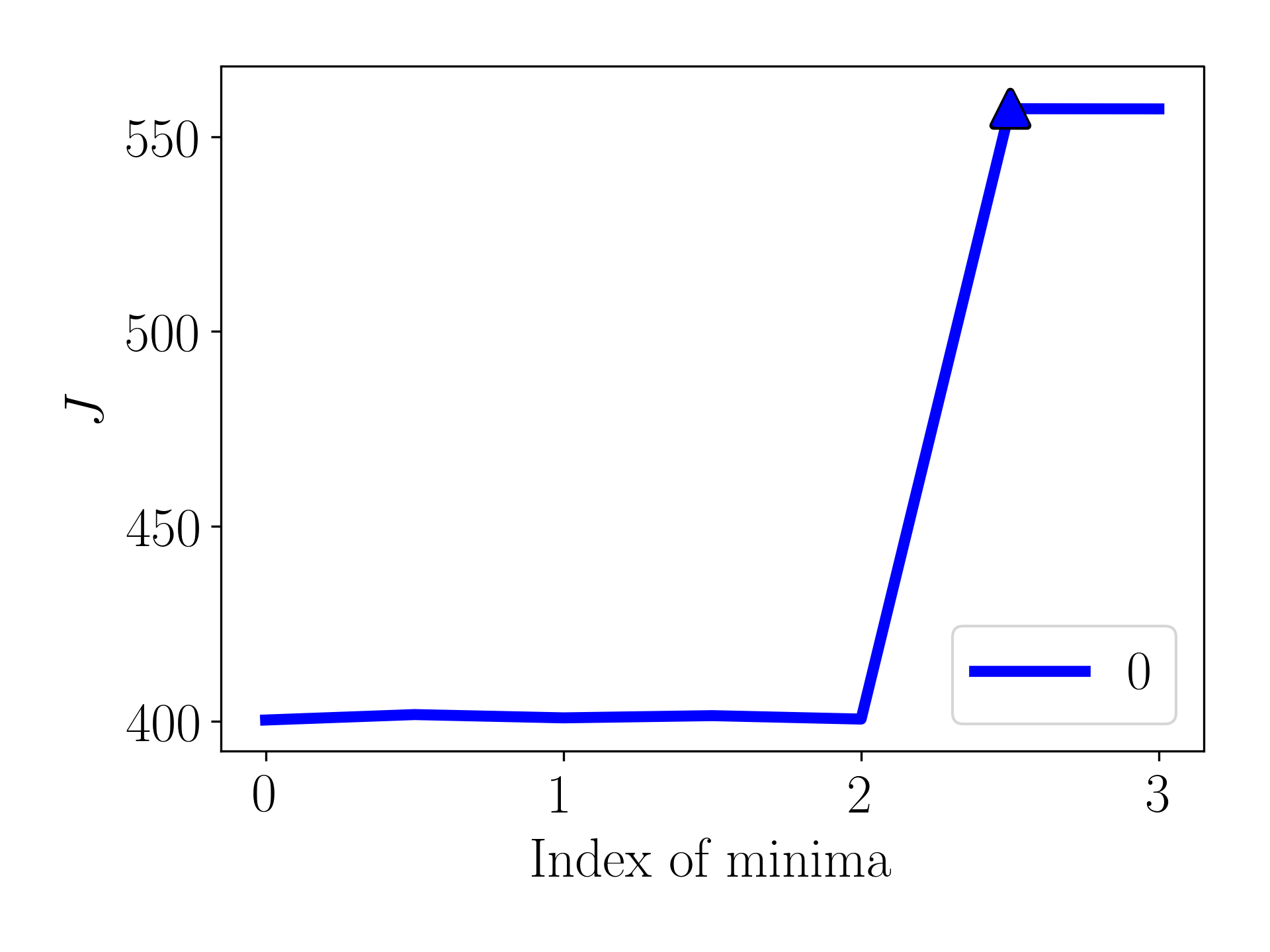}
        \caption{Outlier 1}
    \end{subfigure}%
    \begin{subfigure}[t]{0.45\textwidth}
        \centering
        \includegraphics[width=1.00\textwidth]{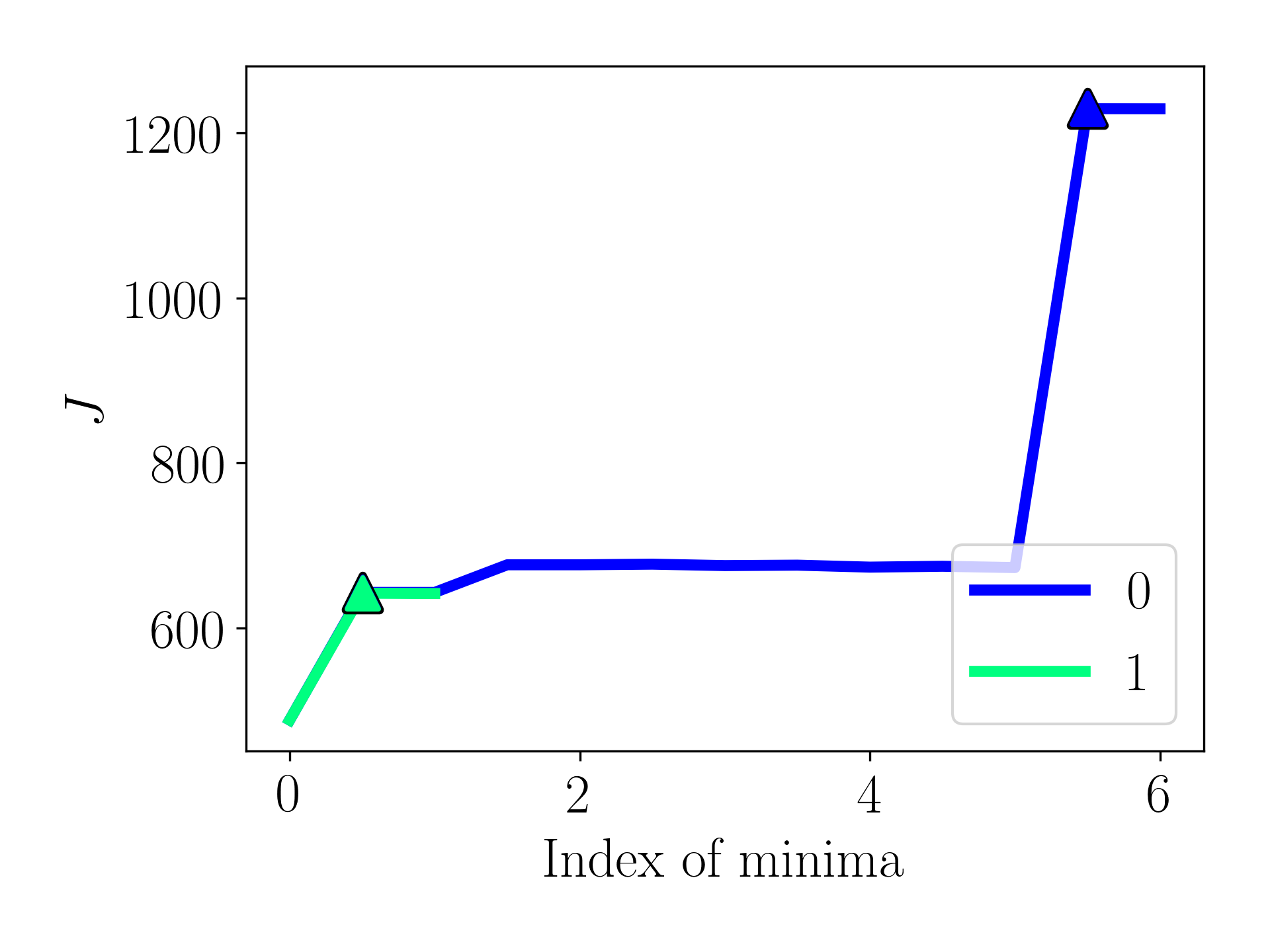}
        \caption{Outlier 2}
    \end{subfigure}\\
    \begin{subfigure}[t]{0.45\textwidth}
        \centering
        \includegraphics[width=1.00\textwidth]{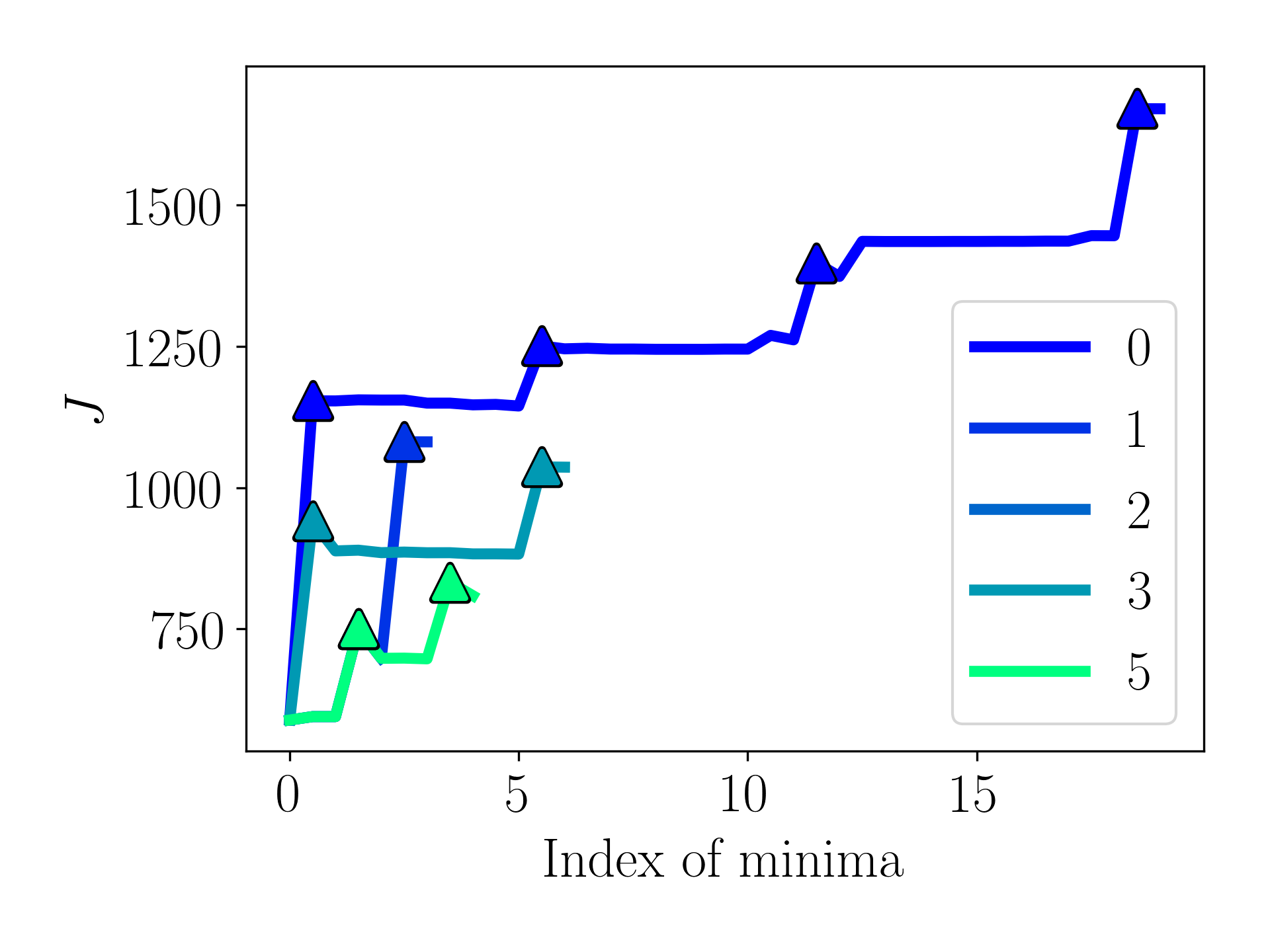}
        \caption{Outlier 3}
    \end{subfigure}%
    \caption{The fastest of all possible paths from a given structure type to
the global minimum for landscapes of the glass dataset, and additional datasets
containing outliers. The paths are shown as sequences of minima and transition
states connected by straight lines. The minima lie at integer values on the
$x$-axis.}
    \label{OutliersGlassPathways}
\end{figure}

\begin{figure}[h!]
    \centering
    \begin{subfigure}[t]{0.33\textwidth}
        \centering
        \includegraphics[width=1.00\textwidth]{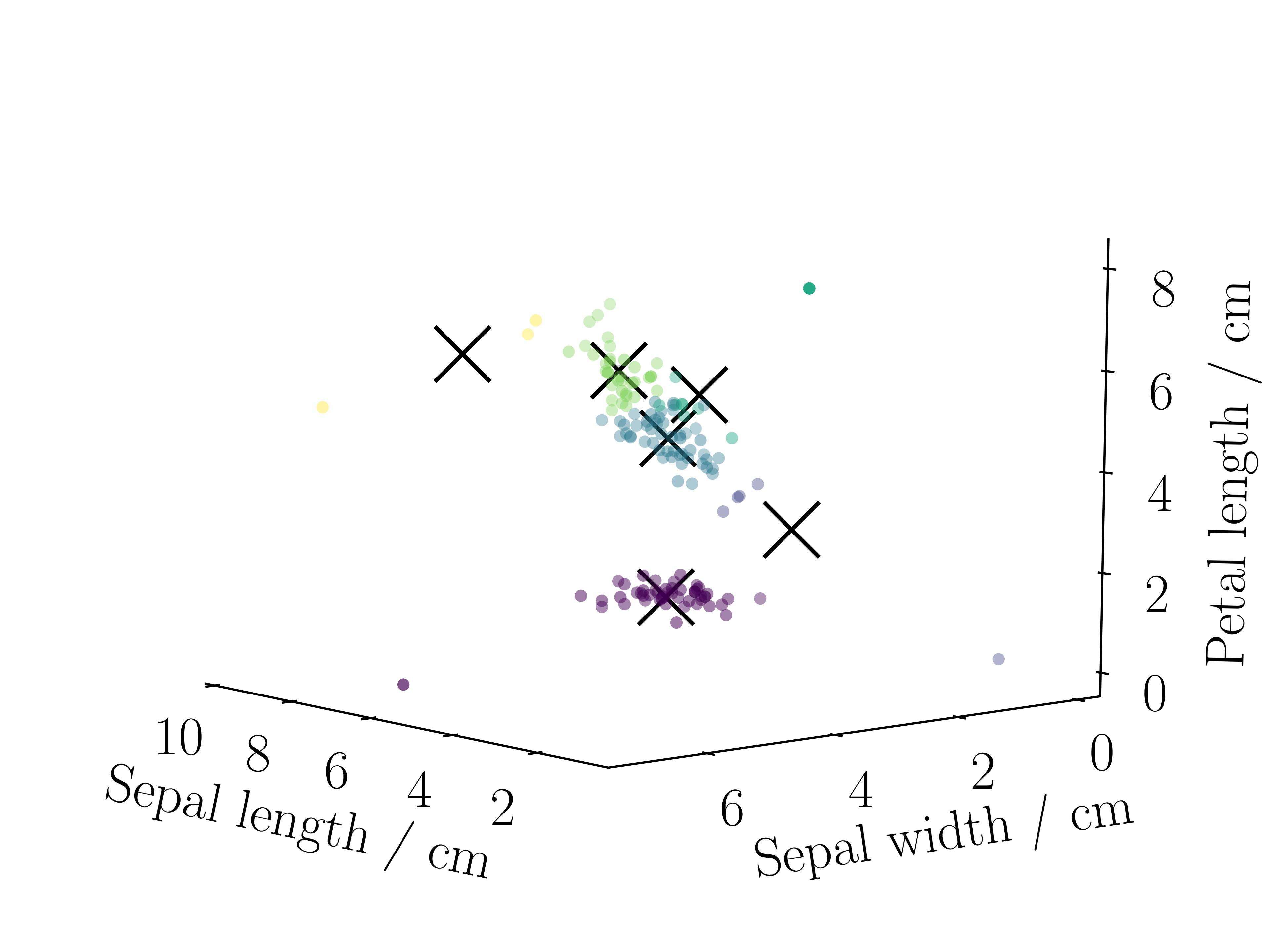}
        \caption{Minimum 1}
    \end{subfigure}%
    \begin{subfigure}[t]{0.33\textwidth}
        \centering
        \includegraphics[width=1.00\textwidth]{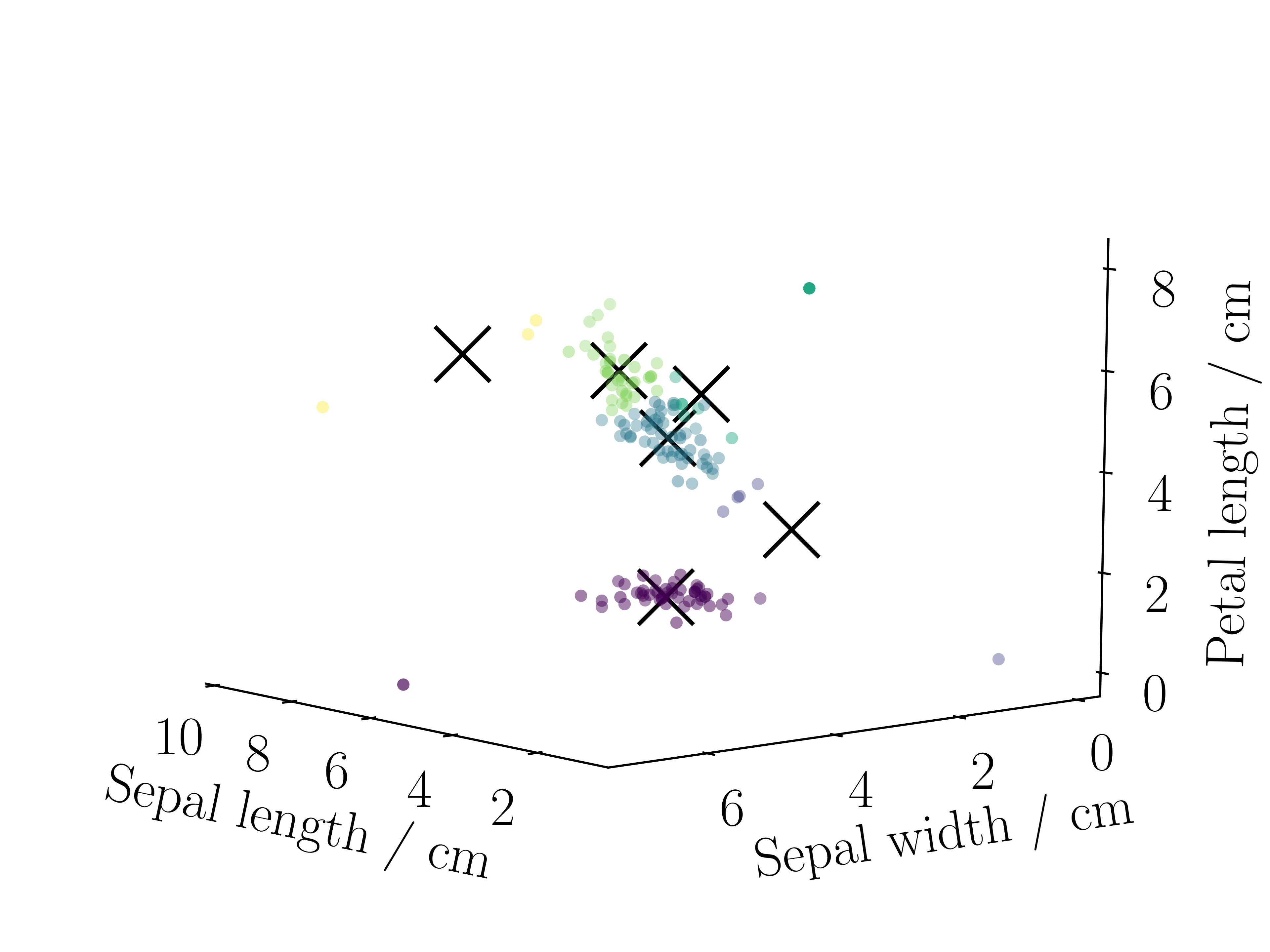}
        \caption{TS 1}
    \end{subfigure}%
    \begin{subfigure}[t]{0.33\textwidth}
        \centering
        \includegraphics[width=1.00\textwidth]{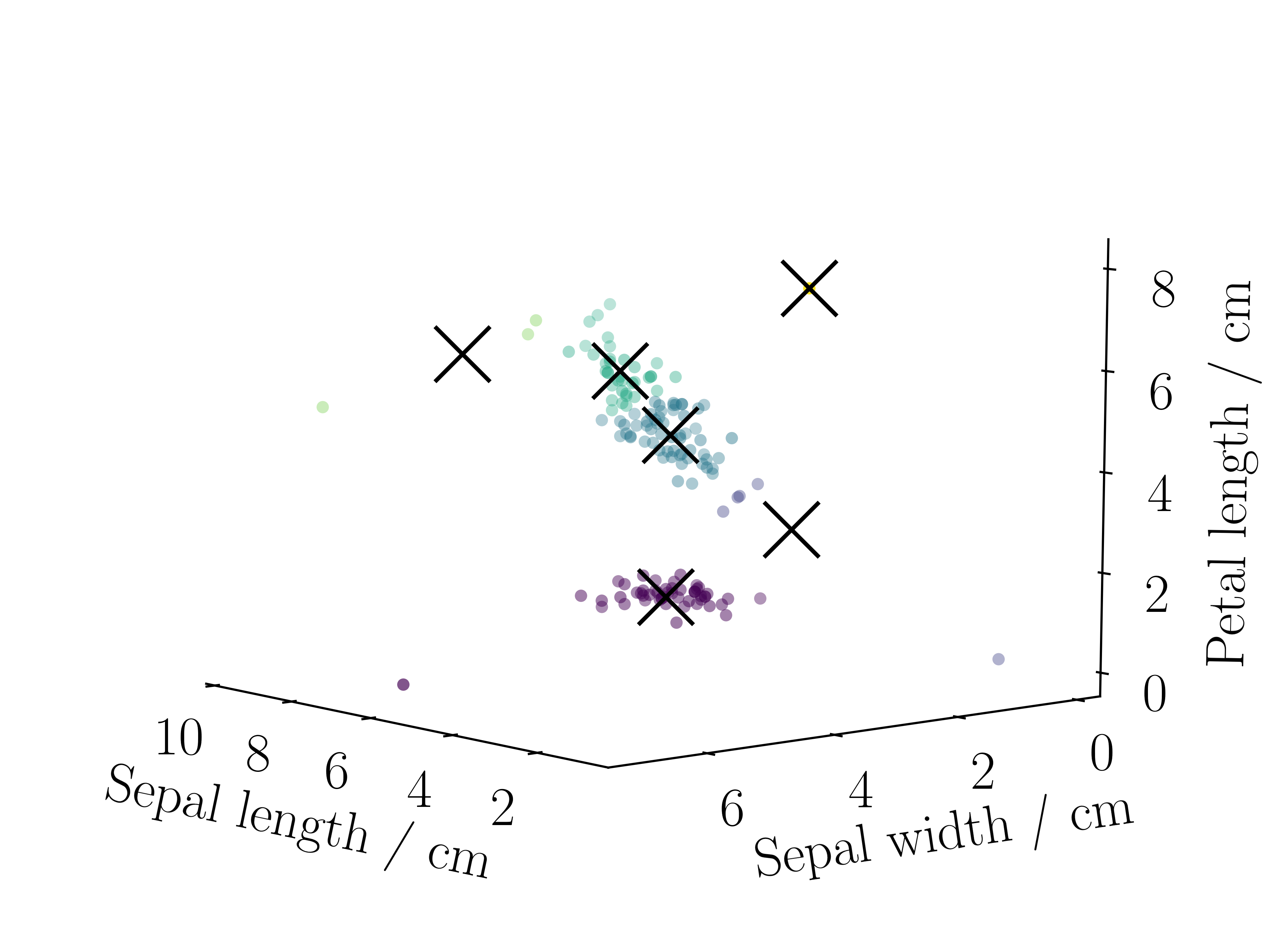}
        \caption{Minimum 2}
    \end{subfigure}\\
    \begin{subfigure}[t]{0.33\textwidth}
        \centering
        \includegraphics[width=1.00\textwidth]{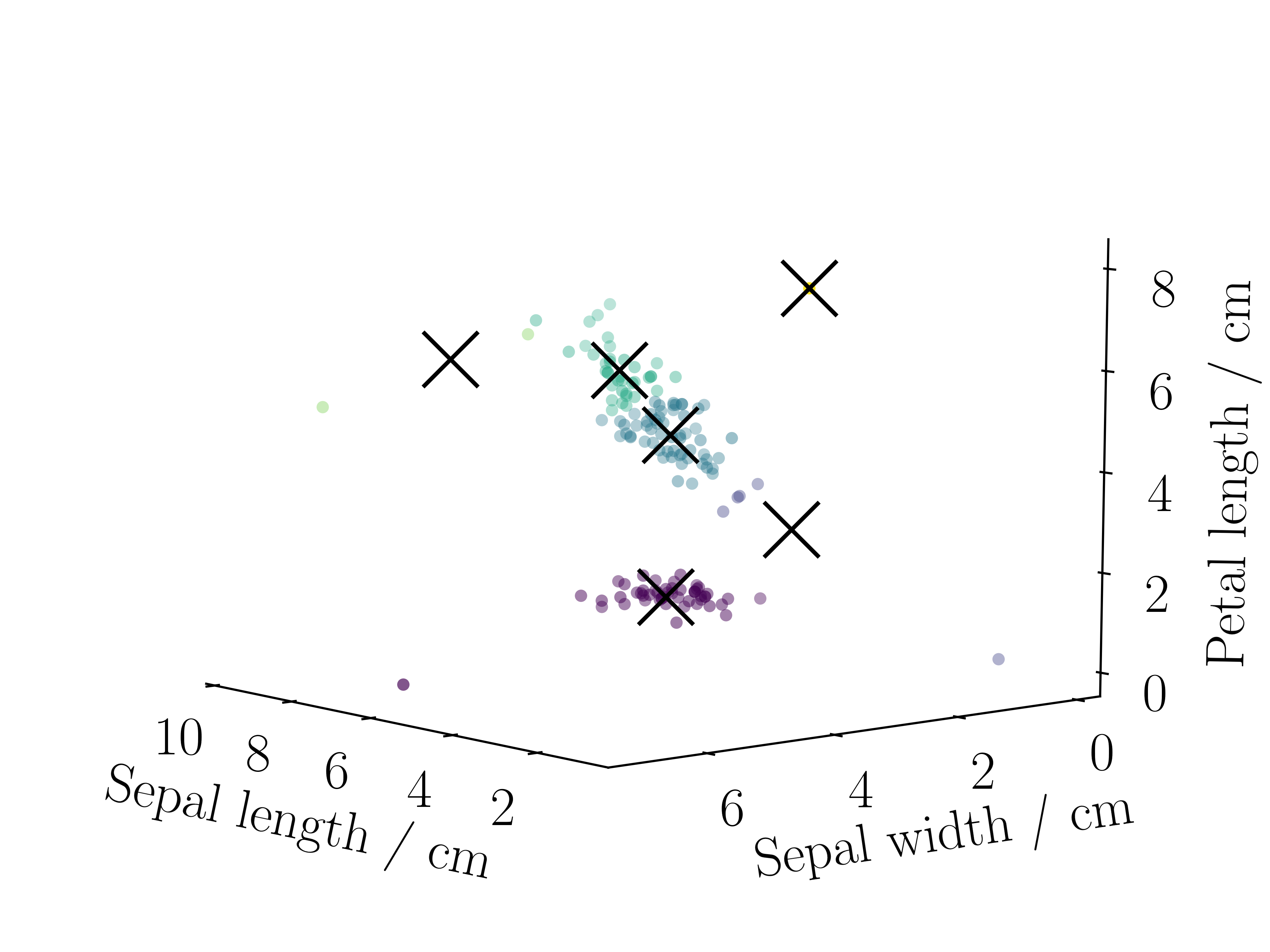}
        \caption{TS 2}
    \end{subfigure}%
    \begin{subfigure}[t]{0.33\textwidth}
        \centering
        \includegraphics[width=1.00\textwidth]{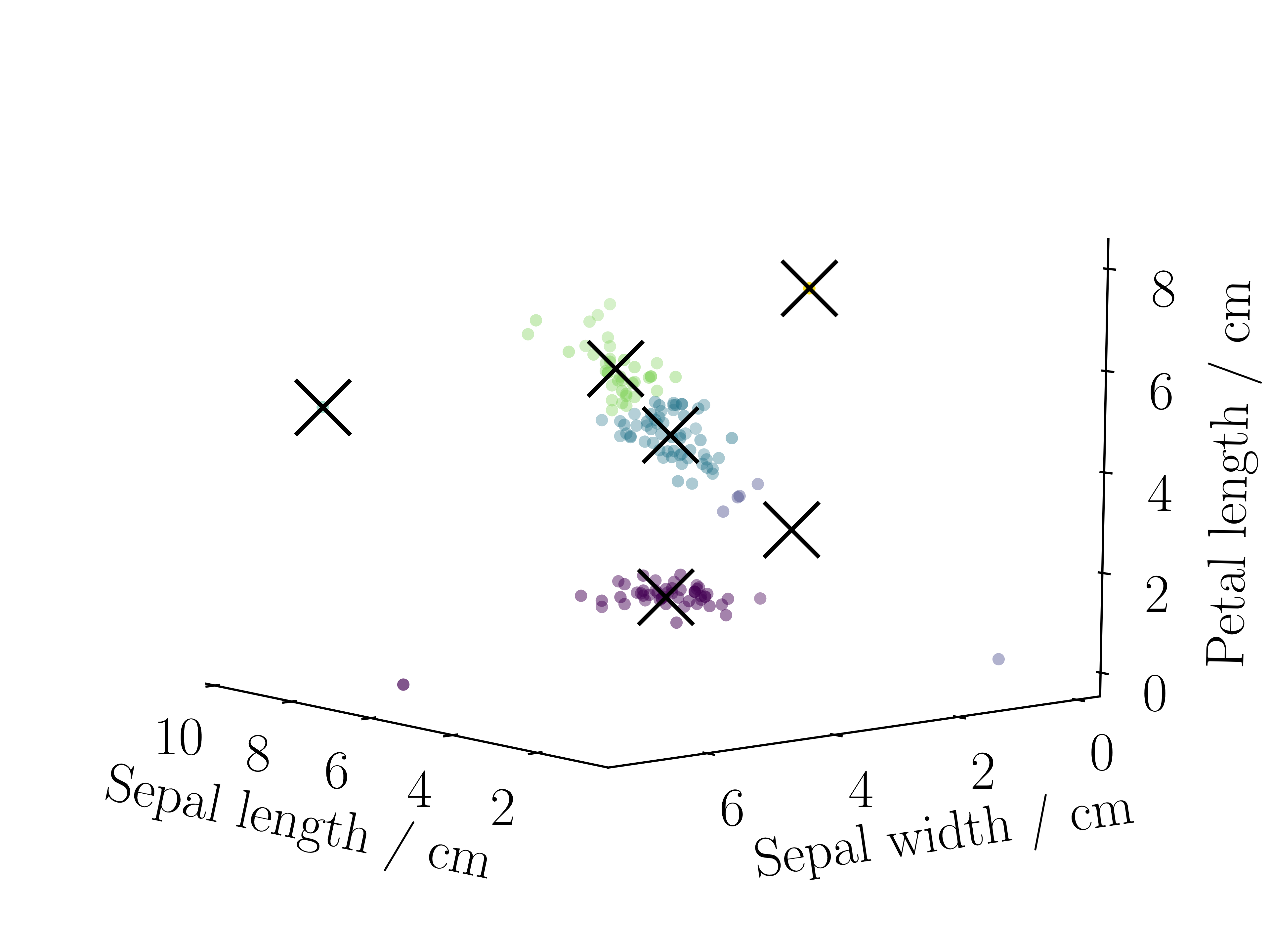}
        \caption{Minimum 3}
    \end{subfigure}%
    \begin{subfigure}[t]{0.33\textwidth}
        \centering
        \includegraphics[width=1.00\textwidth]{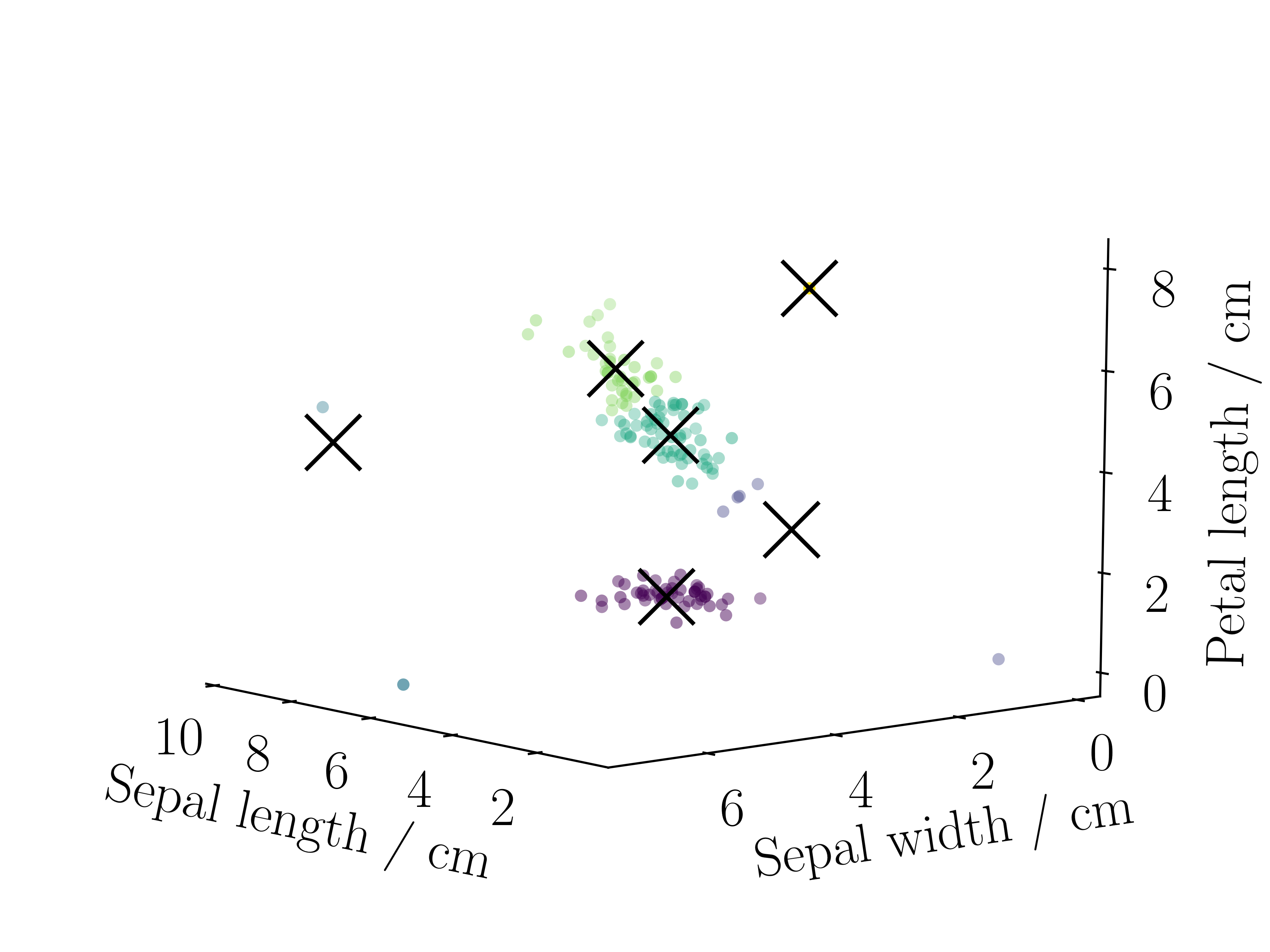}
        \caption{TS 3}
    \end{subfigure}\\
    \begin{subfigure}[t]{0.33\textwidth}
        \centering
        \includegraphics[width=1.00\textwidth]{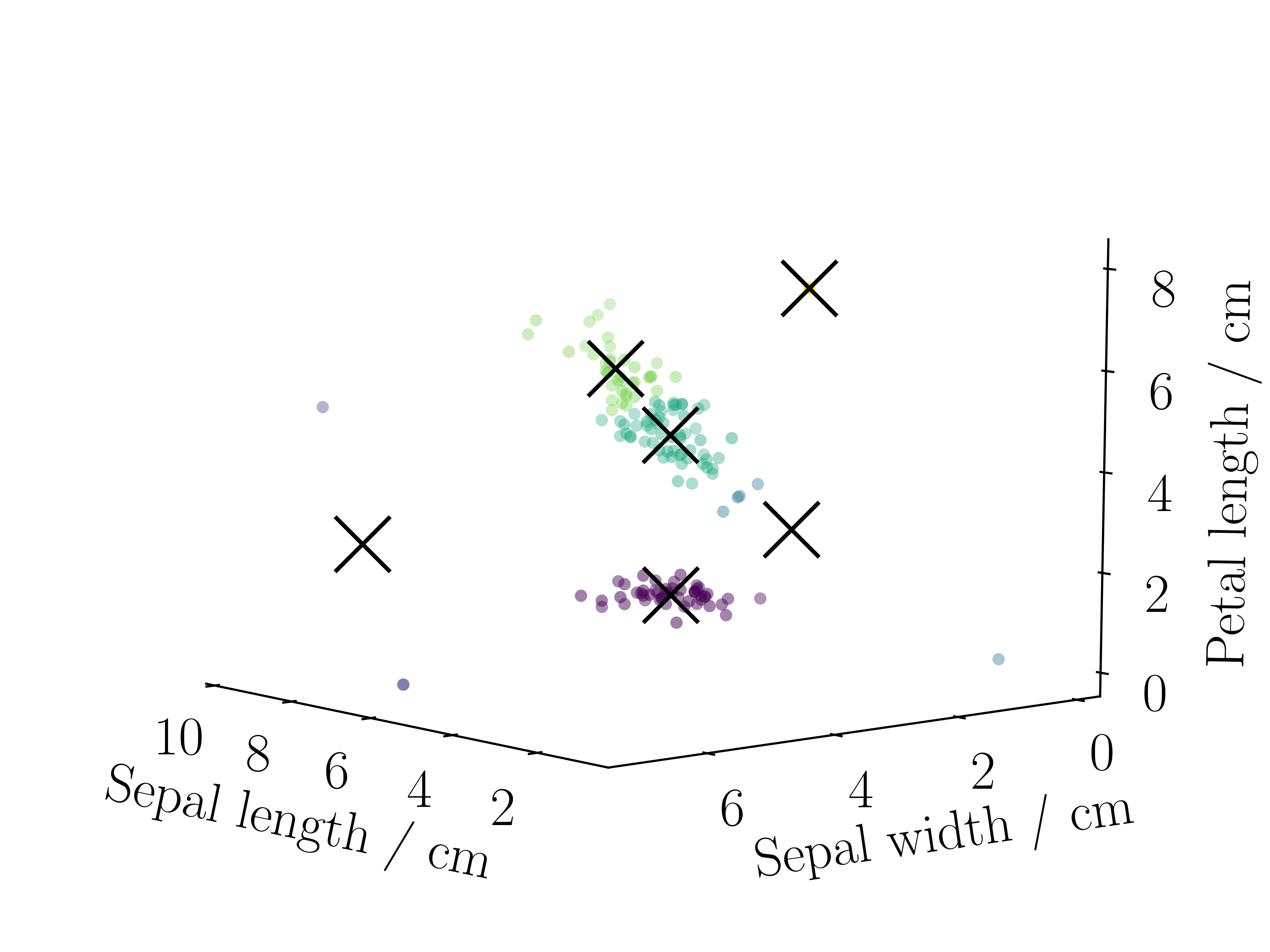}
        \caption{Minimum 4}
    \end{subfigure}%
    \begin{subfigure}[t]{0.33\textwidth}
        \centering
        \includegraphics[width=1.00\textwidth]{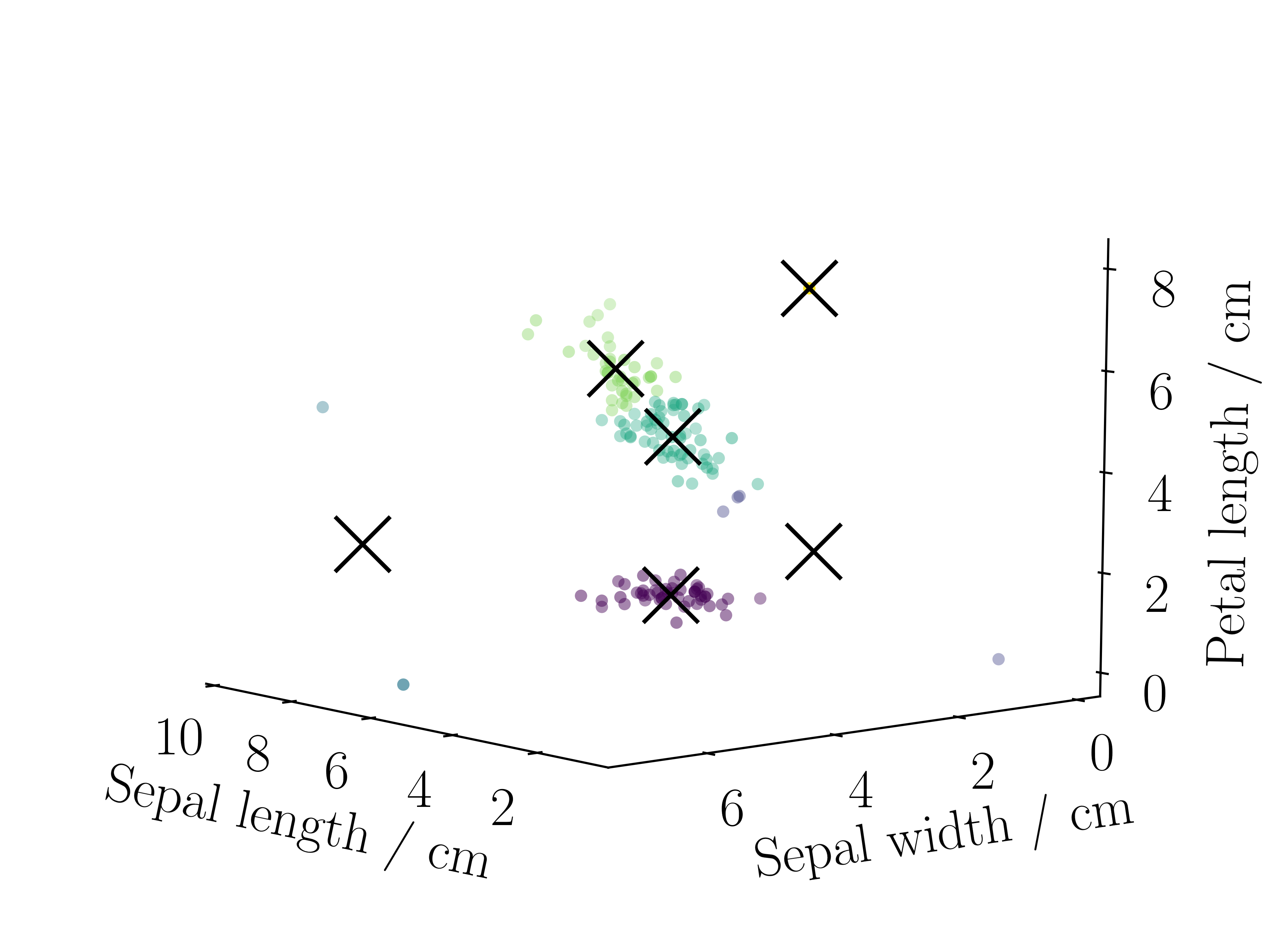}
        \caption{TS 4}
    \end{subfigure}%
    \begin{subfigure}[t]{0.33\textwidth}
        \centering
        \includegraphics[width=1.00\textwidth]{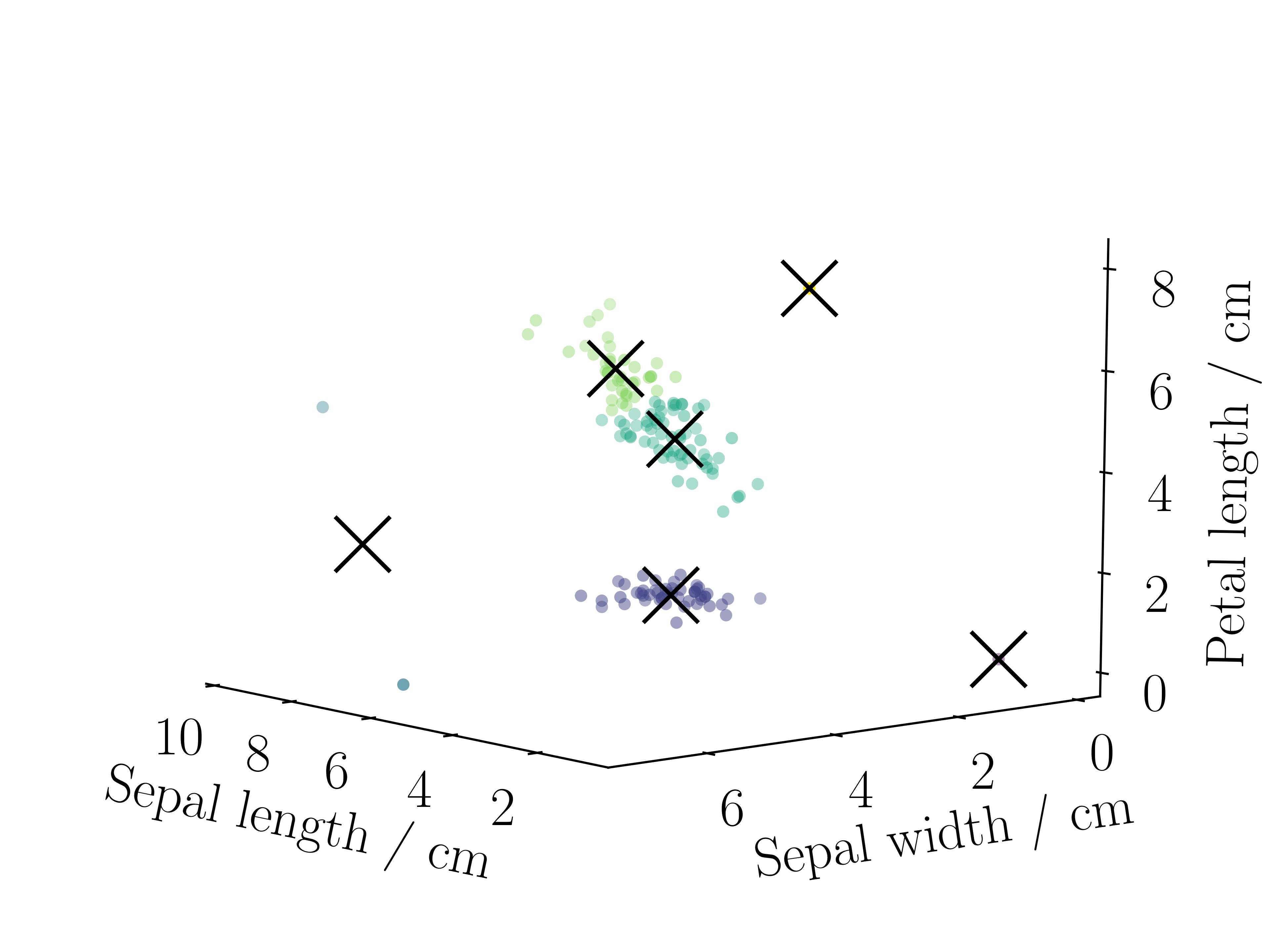}
        \caption{Minimum 5}
    \end{subfigure}%
    \caption{Stationary points on the fastest path between all minima of
structure type 0 and the global minimum of structure type 8 for the
\textit{Iris} dataset with $K=6$ and four outliers. Each stationary point is
visualised using three of the four features. This pathway requires three large
structural changes to progress to the global minimum and, consequently, the
rate is much lower than observed in landscapes with fewer outliers.}
    \label{OutliersIrisOut40Pathway}
\end{figure}

\begin{figure}[h!]
    \centering
    \begin{subfigure}[t]{0.33\textwidth}
        \centering
        \includegraphics[width=1.00\textwidth]{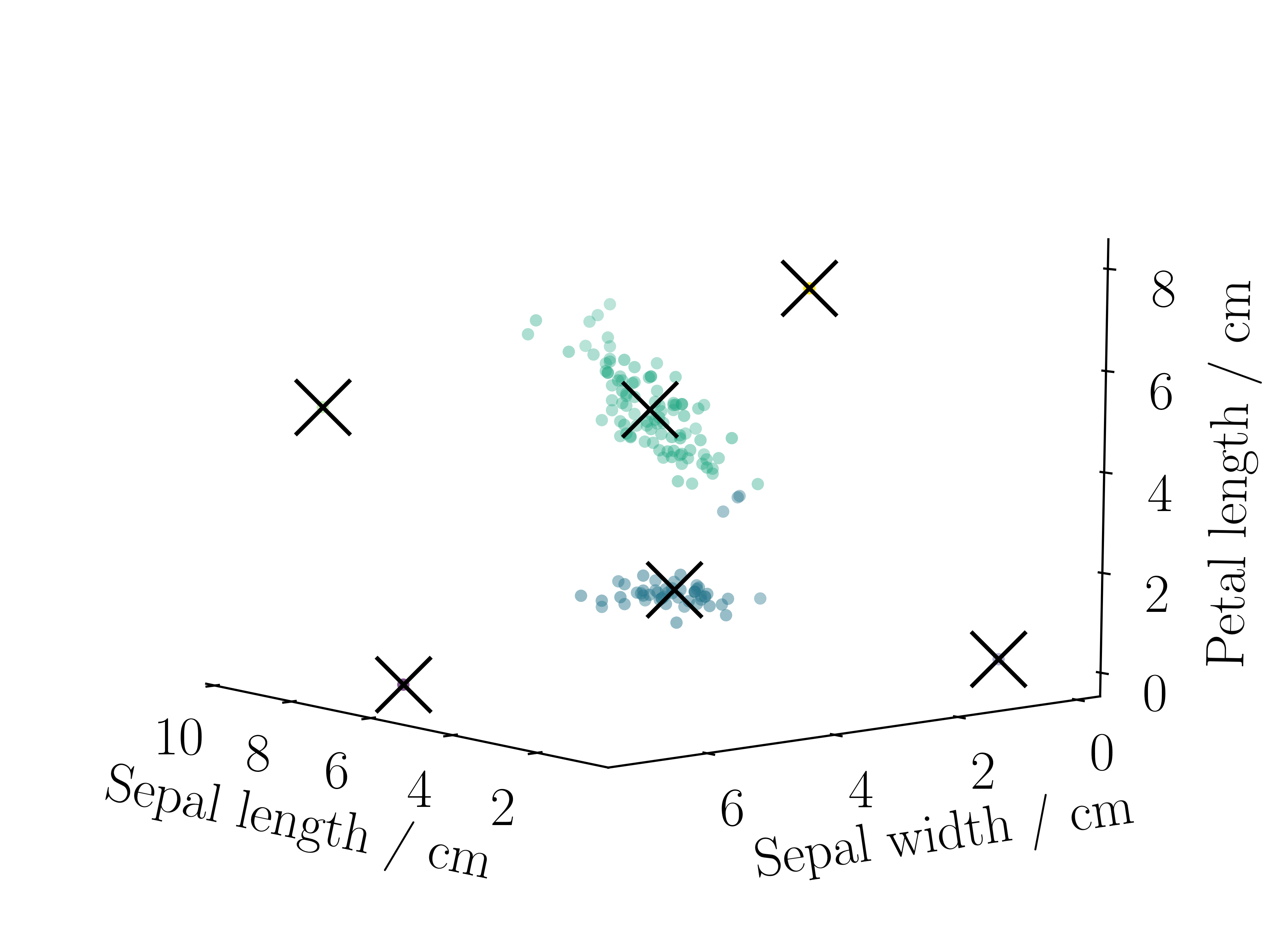}
        \caption{Minimum 1}
    \end{subfigure}%
    \begin{subfigure}[t]{0.33\textwidth}
        \centering
        \includegraphics[width=1.00\textwidth]{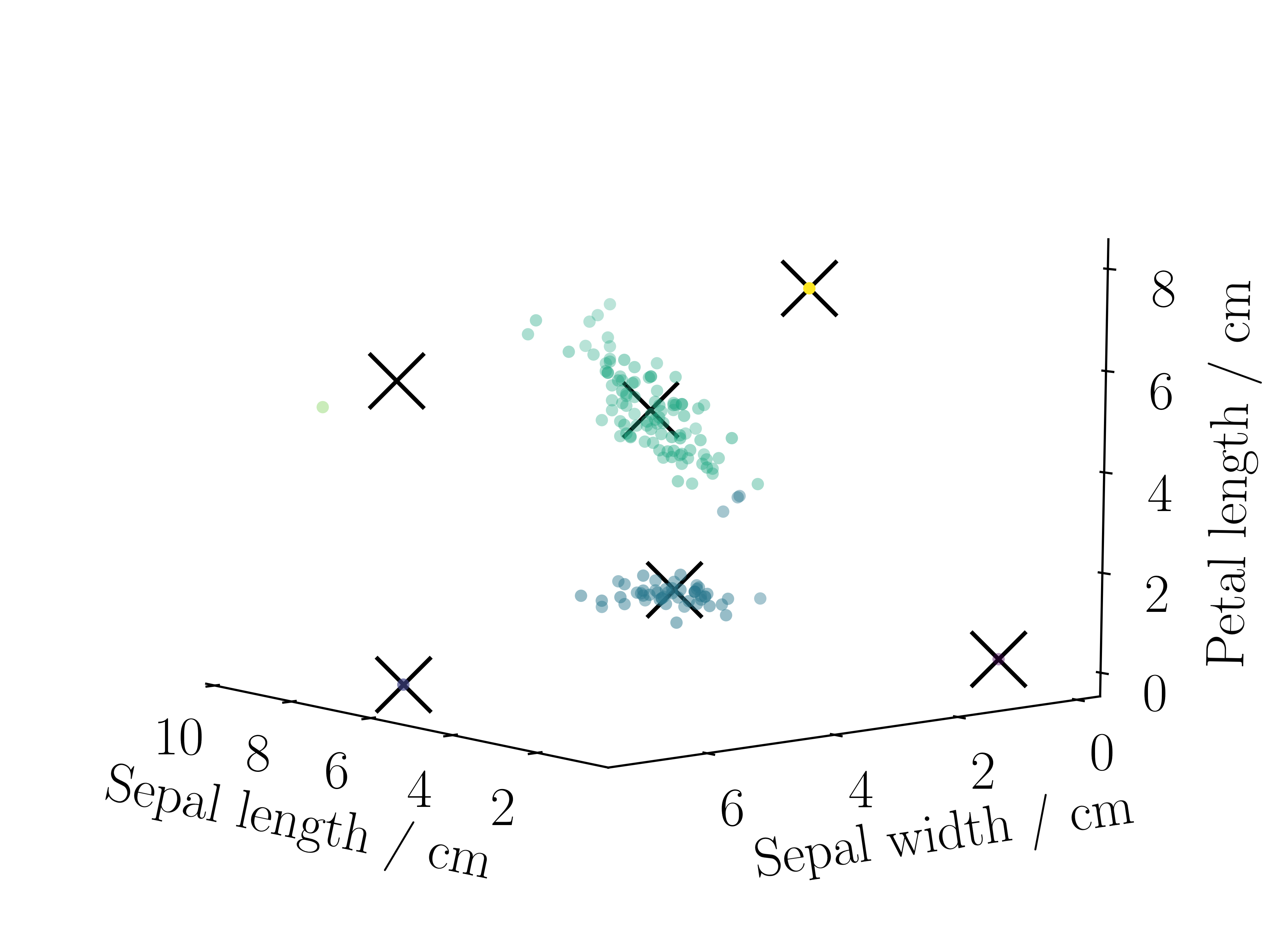}
        \caption{TS 1}
    \end{subfigure}%
    \begin{subfigure}[t]{0.33\textwidth}
        \centering
        \includegraphics[width=1.00\textwidth]{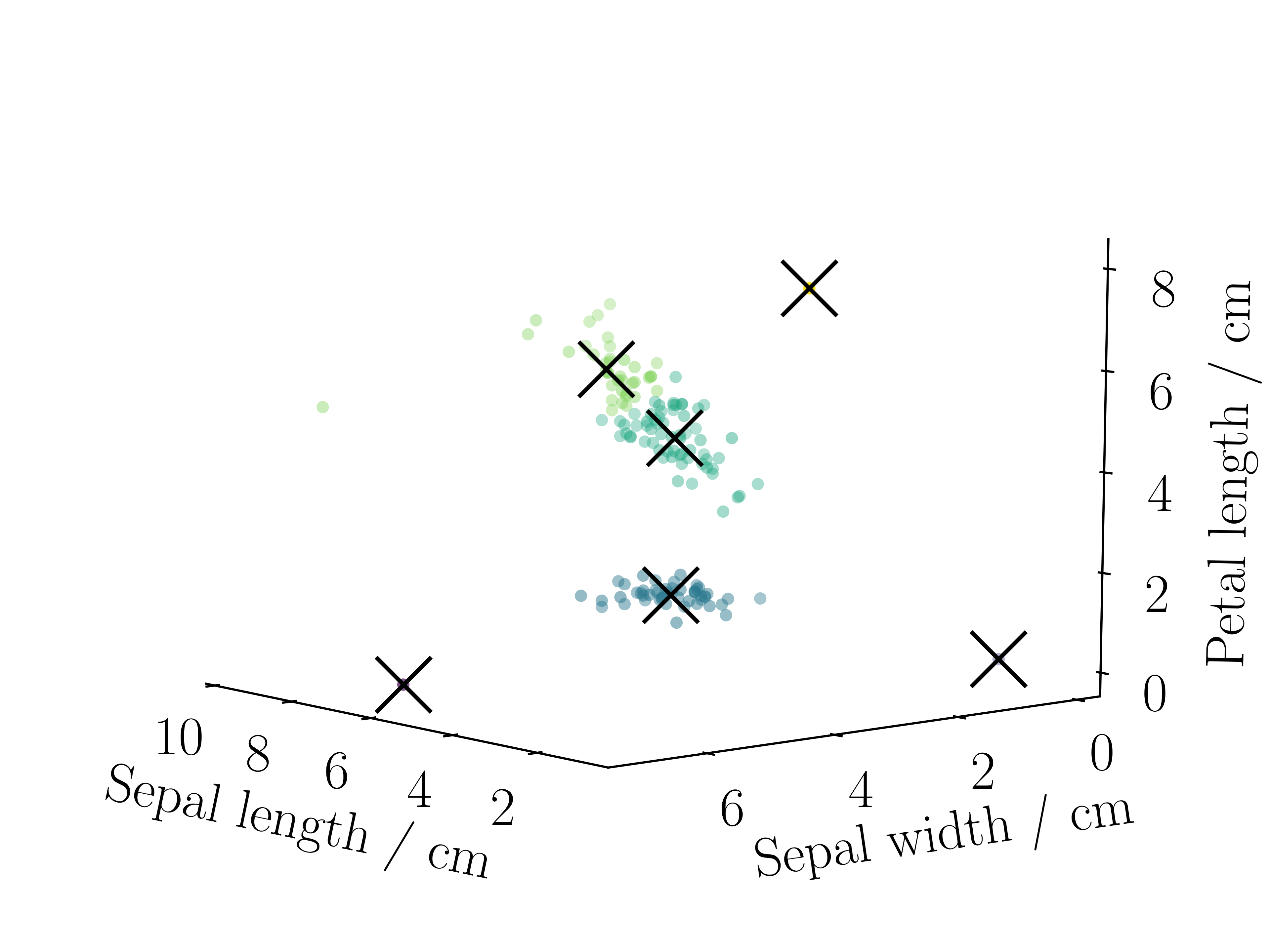}
        \caption{Minimum 2}
    \end{subfigure} \\
    \begin{subfigure}[t]{0.33\textwidth}
        \centering
        \includegraphics[width=1.00\textwidth]{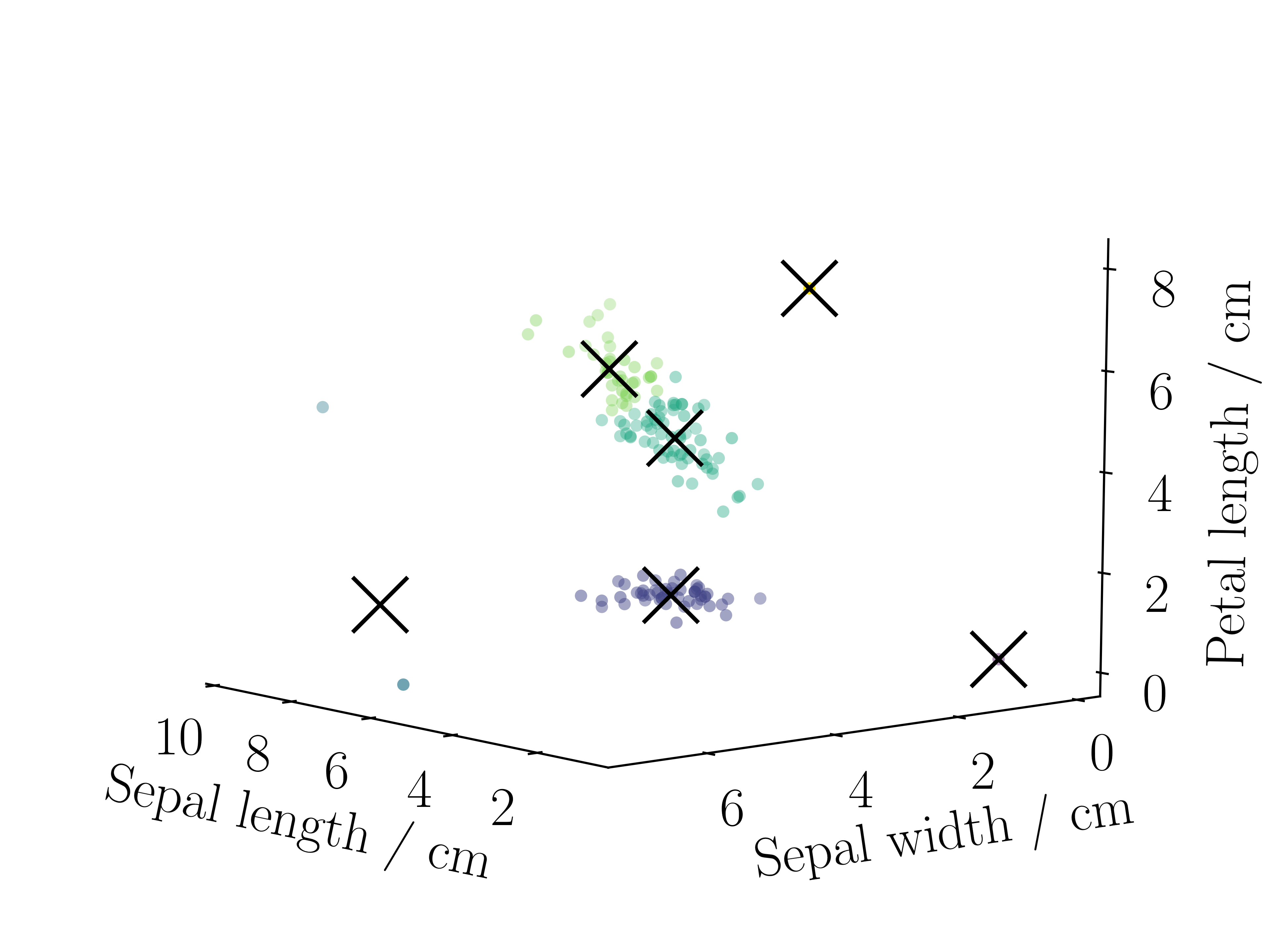}
        \caption{TS 2}
    \end{subfigure}%
    \begin{subfigure}[t]{0.33\textwidth}
        \centering
        \includegraphics[width=1.00\textwidth]{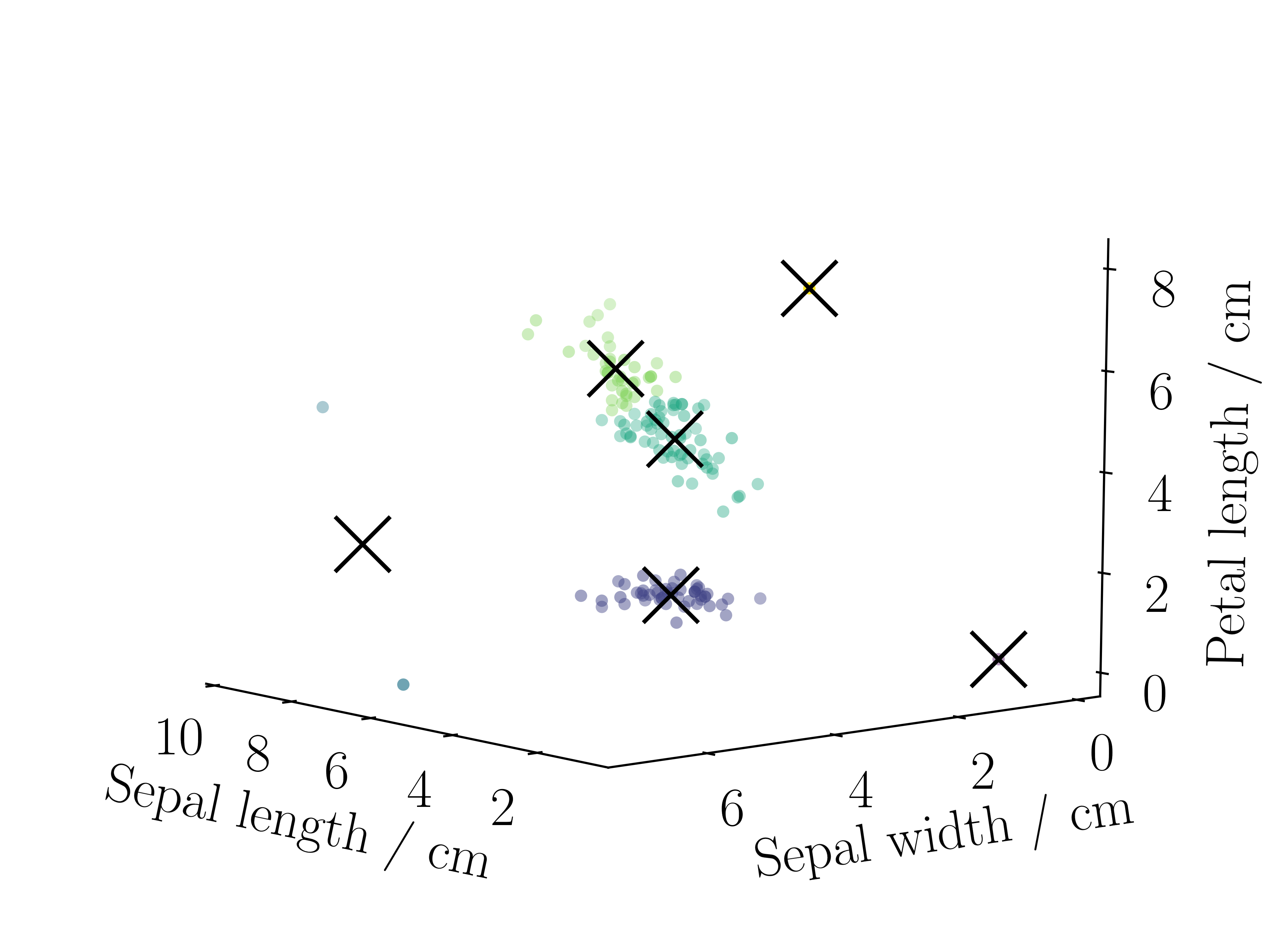}
        \caption{Minimum 3}
    \end{subfigure}%
    \caption{Minima and transition states on the fastest path between the set
of minima of structure type 7 and the global minimum of structure type 8 for
the \textit{Iris} dataset with six clusters and four outliers. The pathway
retains a rate similar paths in the landscape with one outlier because
of the similarity between the start and end minima. The similar structure types
mean that less reorganisation is needed, and the transition can be achieved
with fewer intermediate minima.}
    \label{OutliersIrisOut47Pathway}
\end{figure}

\begin{figure}[!htb]
    \centering
    \begin{subfigure}[t]{0.4\textwidth}
        \centering
        \includegraphics[width=1.00\textwidth]{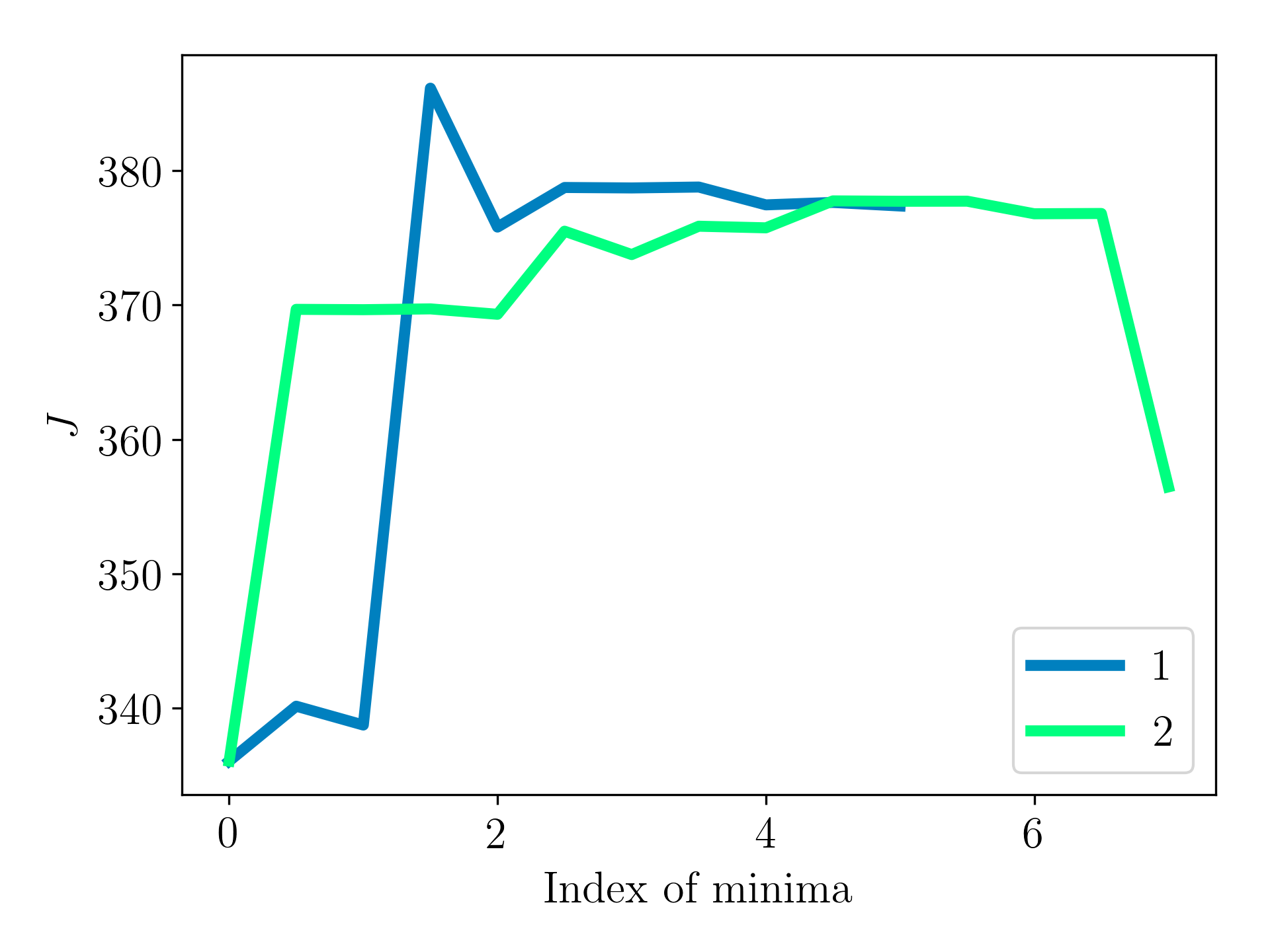}
        \caption{Original}
    \end{subfigure}%
    \begin{subfigure}[t]{0.4\textwidth}
        \centering
        \includegraphics[width=1.00\textwidth]{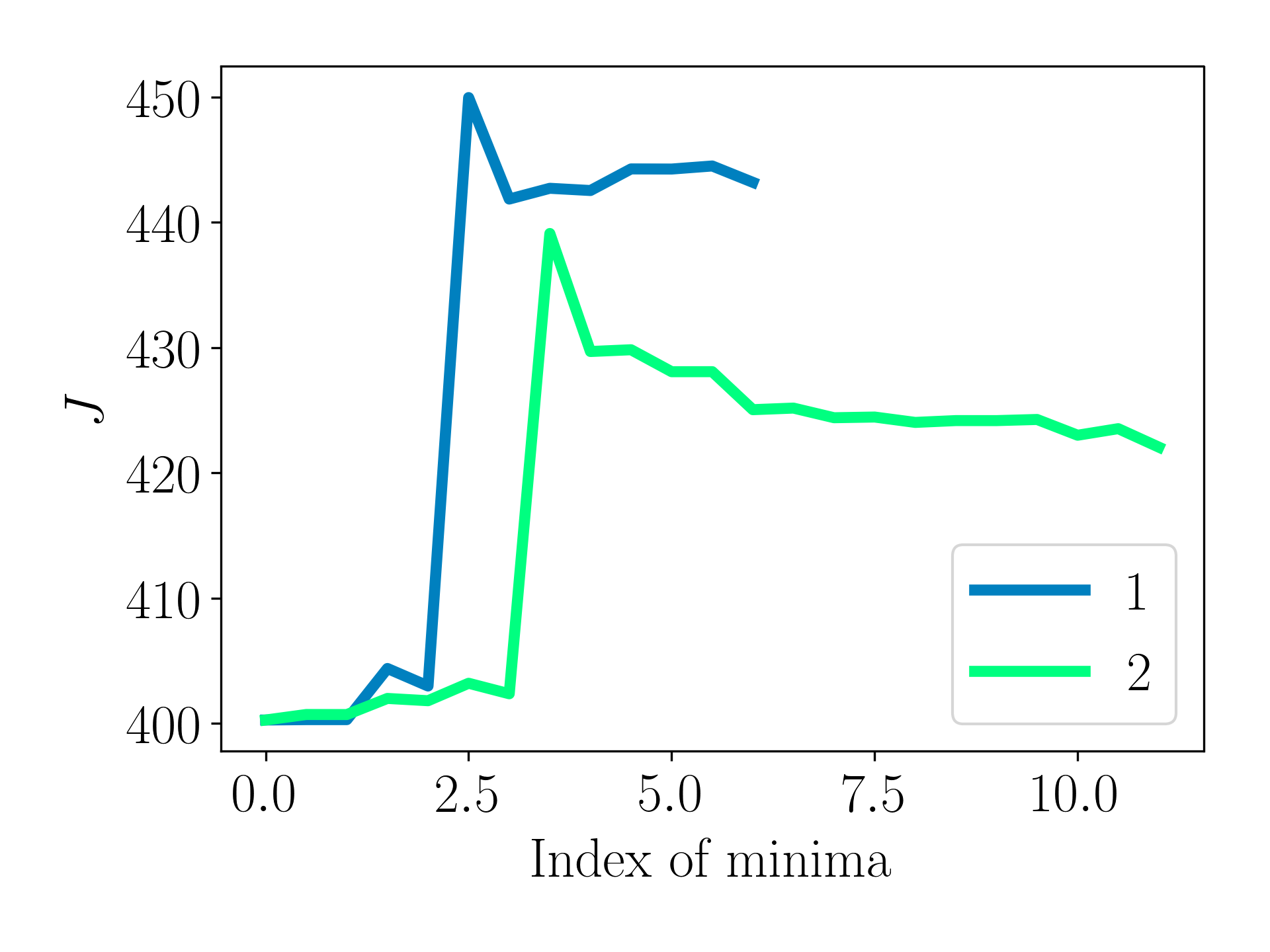}
        \caption{Outlier 1}
    \end{subfigure}\\
    \begin{subfigure}[t]{0.4\textwidth}
        \centering
        \includegraphics[width=1.00\textwidth]{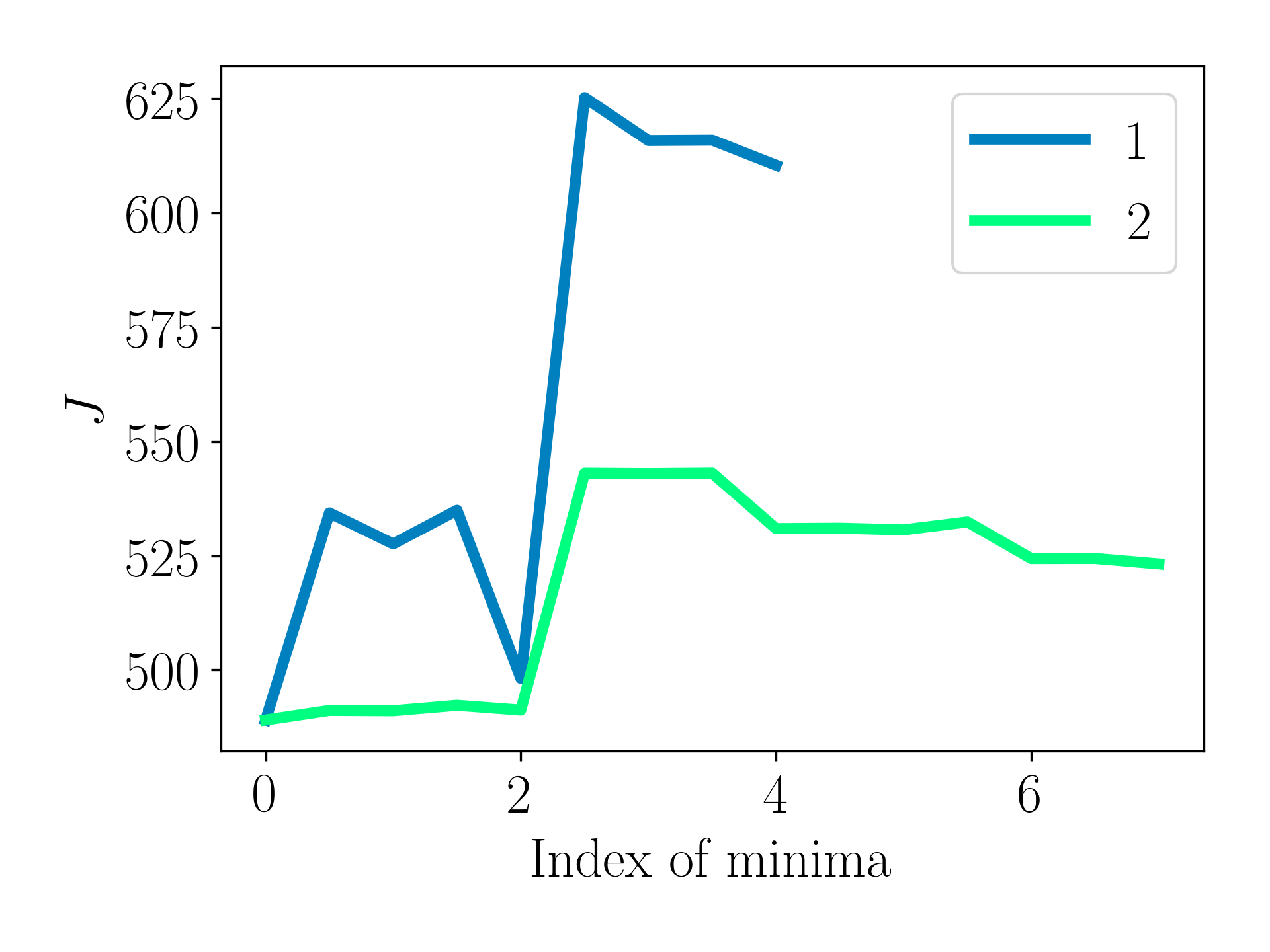}
        \caption{Outlier 2}
    \end{subfigure}%
    \begin{subfigure}[t]{0.4\textwidth}
        \centering
        \includegraphics[width=1.00\textwidth]{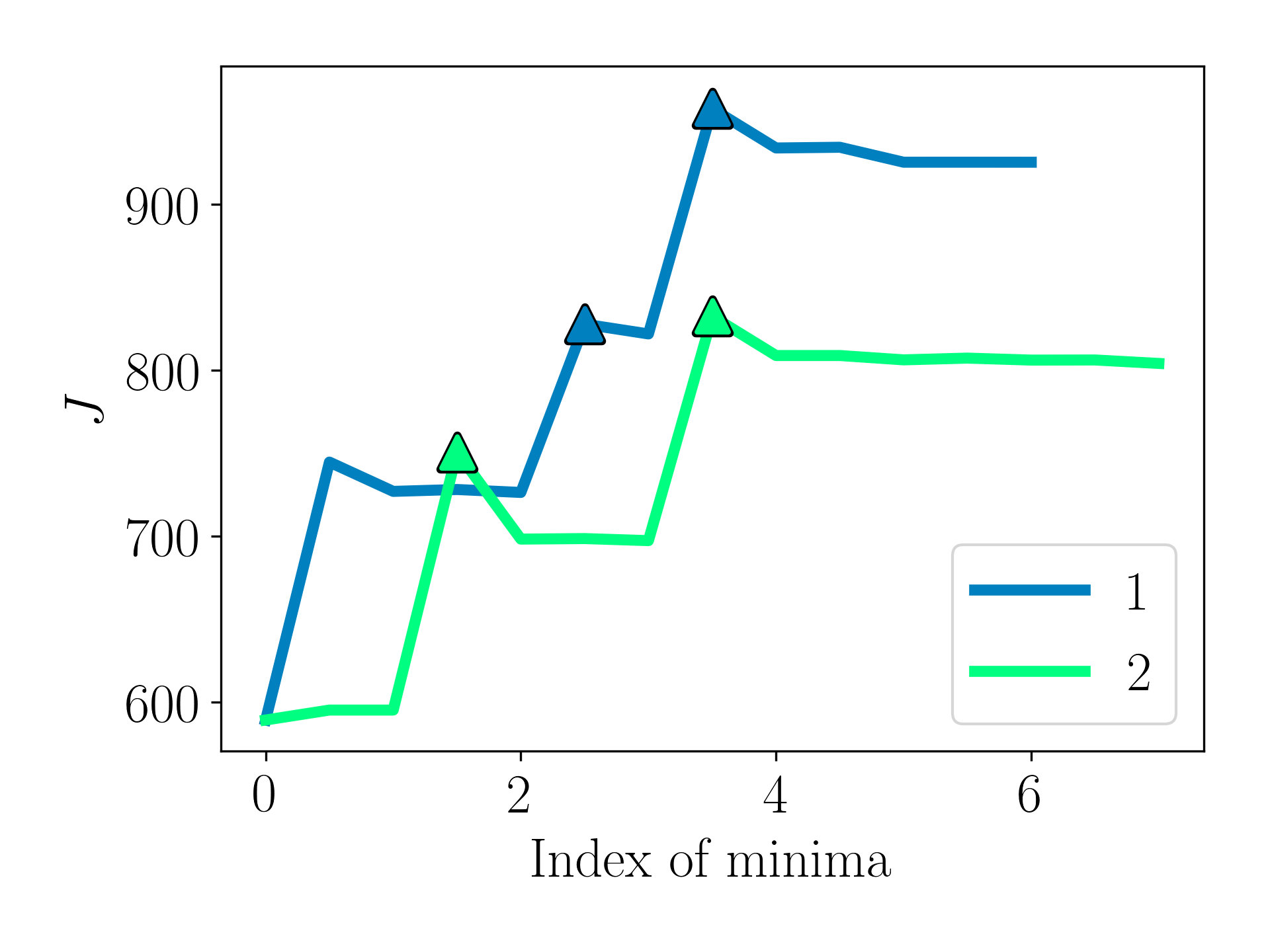}
        \caption{Outlier 3}
    \end{subfigure}\\
    \caption{Fastest paths from selected kinetic traps to the global minimum,
for the $K$-means landscapes of the glass dataset and its variations. Paths are
represented by minima and transition states, connected by straight lines. The
minima lie at integer values on the horizontal axis and transition states give the
cost function barrier between them. Changes in structure type are denoted by
triangles placed on the transition state between the two differing minima.}
    \label{OutliersGlassKTPathways}
\end{figure}

\begin{figure}[!htb]
    \centering
    \begin{subfigure}[t]{0.33\textwidth}
        \centering
        \psfrag{epsilon}{\small{10}}
        \psfrag{950}{\scriptsize{2}}
        \psfrag{1189}{\scriptsize{1}}
        \includegraphics[width=1.0\textwidth]{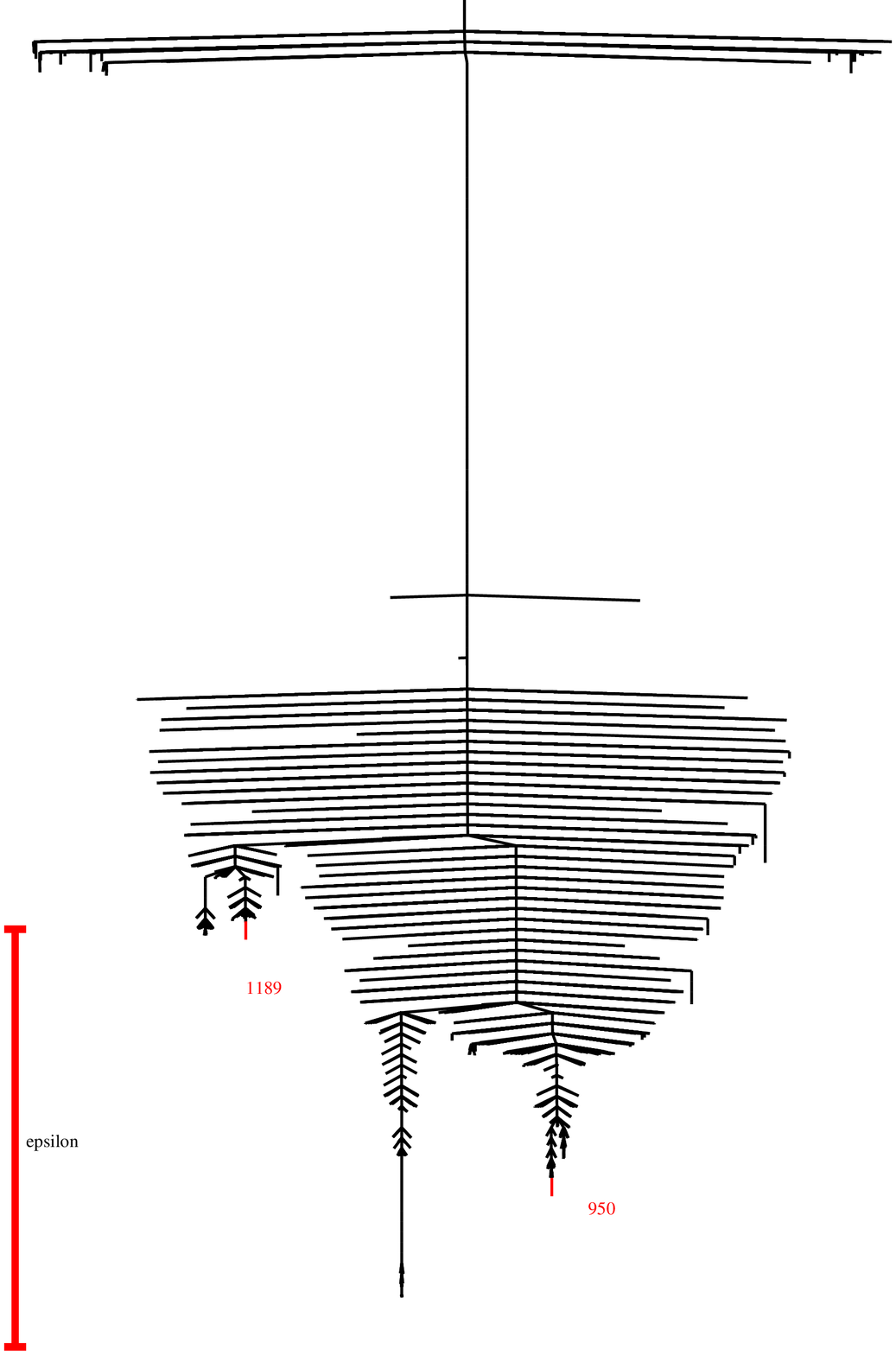}
        \caption{Original}
    \end{subfigure}%
    \begin{subfigure}[t]{0.33\textwidth}
        \centering
        \psfrag{epsilon}{\small{}}
        \psfrag{1421}{\scriptsize{1}}
        \psfrag{74}{\scriptsize{2}}
        \psfrag{257}{\scriptsize{3}}
        \includegraphics[width=1.0\textwidth]{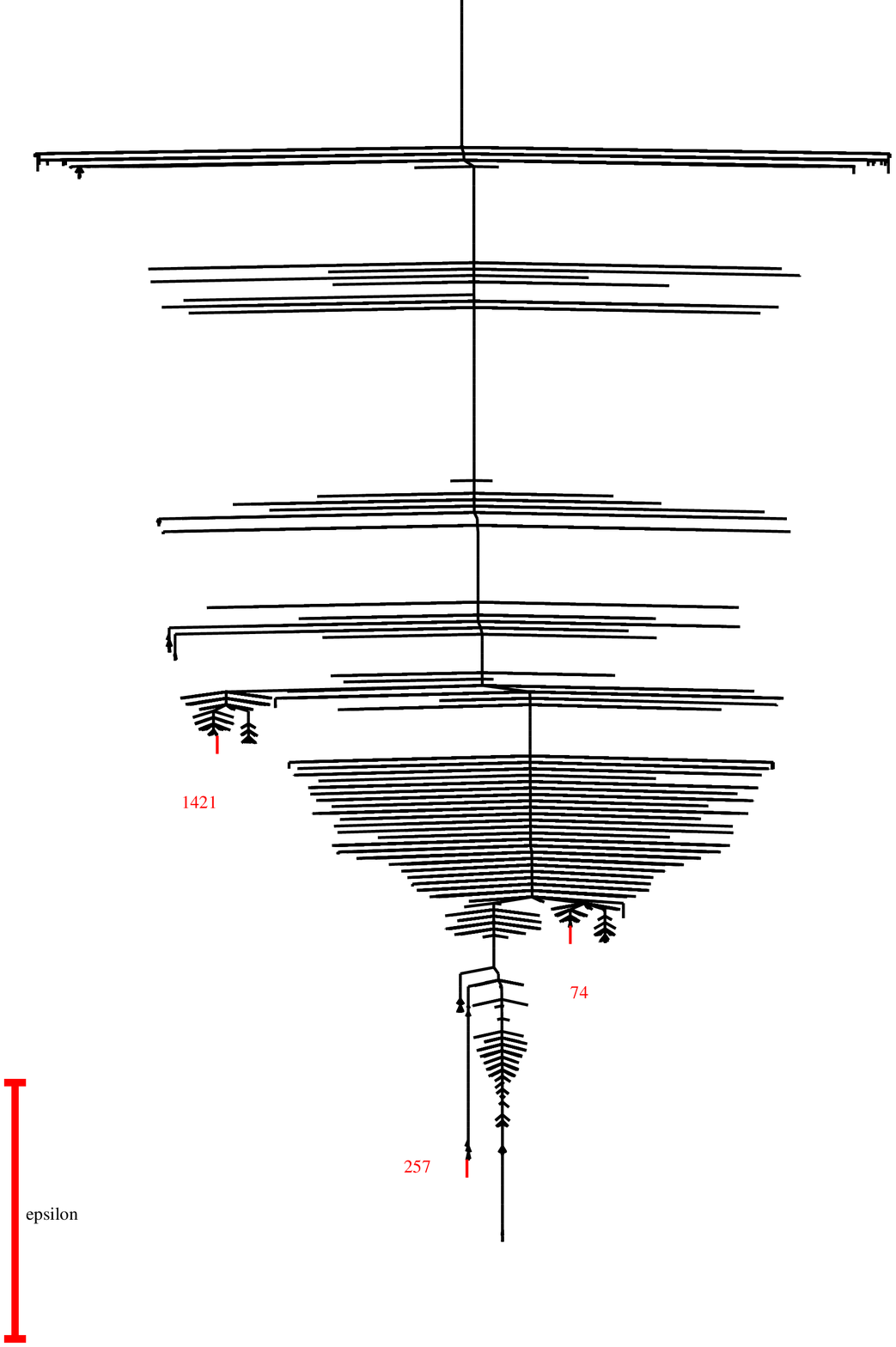}
        \caption{Outlier 1}
    \end{subfigure}%
    \begin{subfigure}[t]{0.33\textwidth}
        \centering
        \psfrag{epsilon}{\small{}}
        \psfrag{318}{\scriptsize{3}}
        \psfrag{675}{\scriptsize{1}}
        \psfrag{634}{\scriptsize{2}}
        \includegraphics[width=1.0\textwidth]{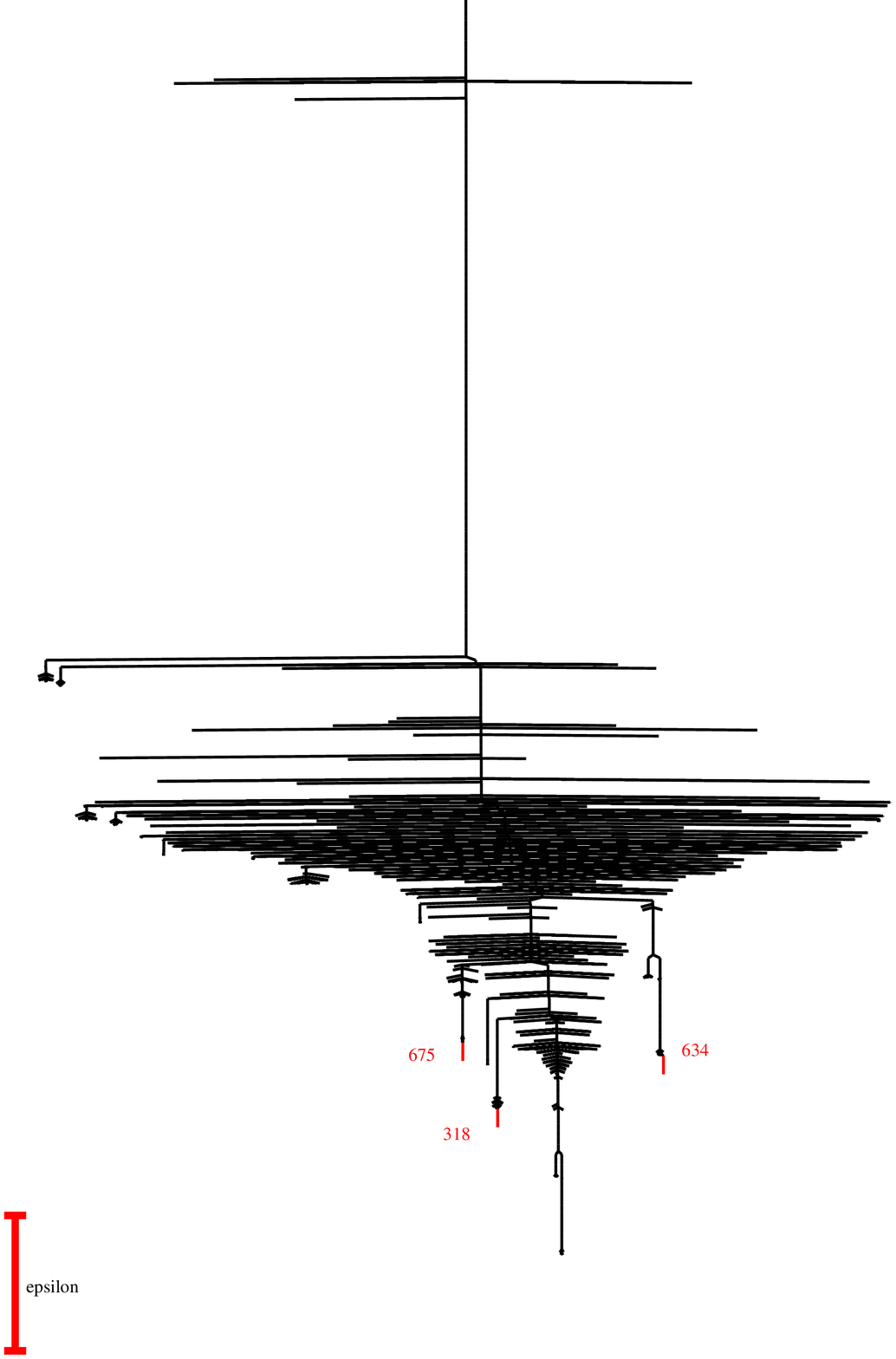}
        \caption{Outlier 2}
    \end{subfigure} \\
    \begin{subfigure}[t]{0.33\textwidth}
        \centering
        \psfrag{epsilon}{\small{}}
        \psfrag{74}{\scriptsize{1}}
        \psfrag{541}{\scriptsize{3}}
        \psfrag{111}{\scriptsize{2}}
        \includegraphics[width=1.0\textwidth]{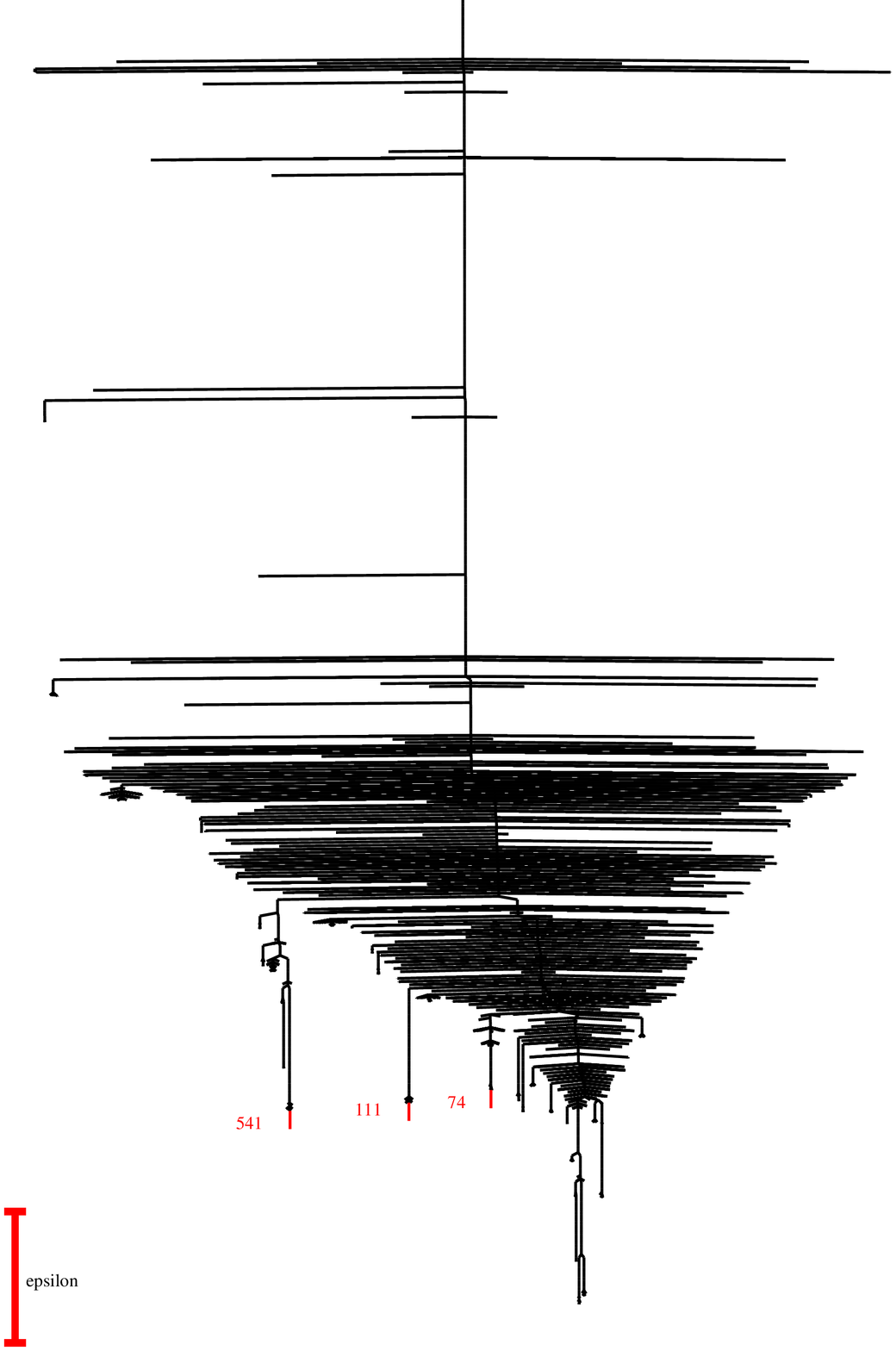}
        \caption{Outlier 3}
    \end{subfigure}%
    \begin{subfigure}[t]{0.33\textwidth}
        \centering
        \psfrag{epsilon}{\small{}}
        \psfrag{782}{\scriptsize{3}}
        \psfrag{174}{\scriptsize{1}}
        \psfrag{1363}{\scriptsize{2}}
        \includegraphics[width=1.0\textwidth]{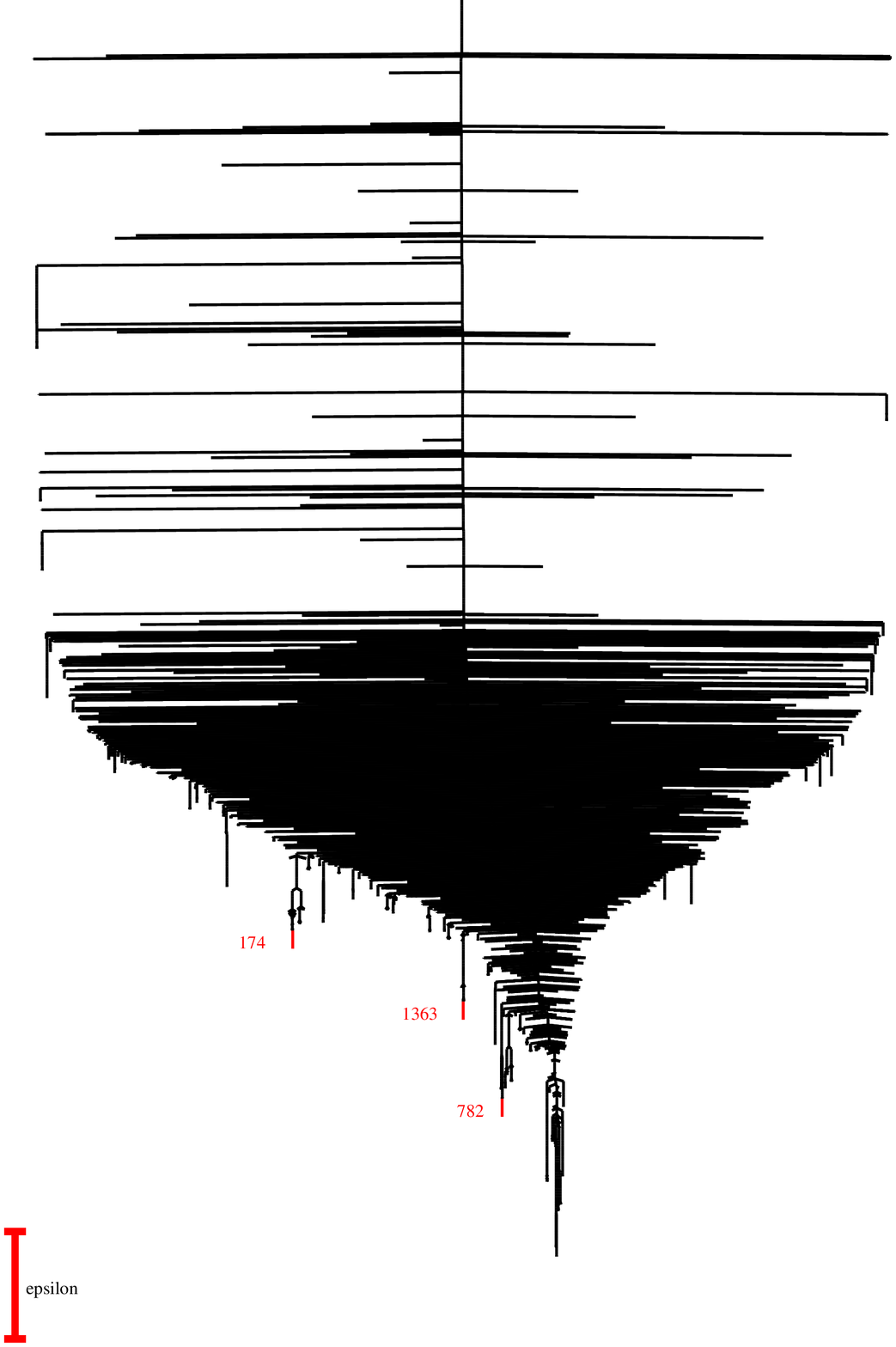}
        \caption{Outlier 4}
    \end{subfigure} \\
    \caption{$K$-means landscapes for Fisher's \textit{Iris} dataset with $K=6$, highlighting kinetic traps. The lowest minimum from each kinetic trap is coloured red, and the associated numerical label is given. The numbering corresponds to that used in Sec.~A.3 of the main text.}
    \label{OutliersIris6KTDGs}
\end{figure}

\begin{figure}[!htb]
    \centering
    \begin{subfigure}[t]{0.4\textwidth}
        \centering
        \psfrag{epsilon}{\small{80}}
        \psfrag{3371}{\scriptsize{1}}
        \psfrag{1309}{\scriptsize{2}}
        \includegraphics[width=1.0\textwidth]{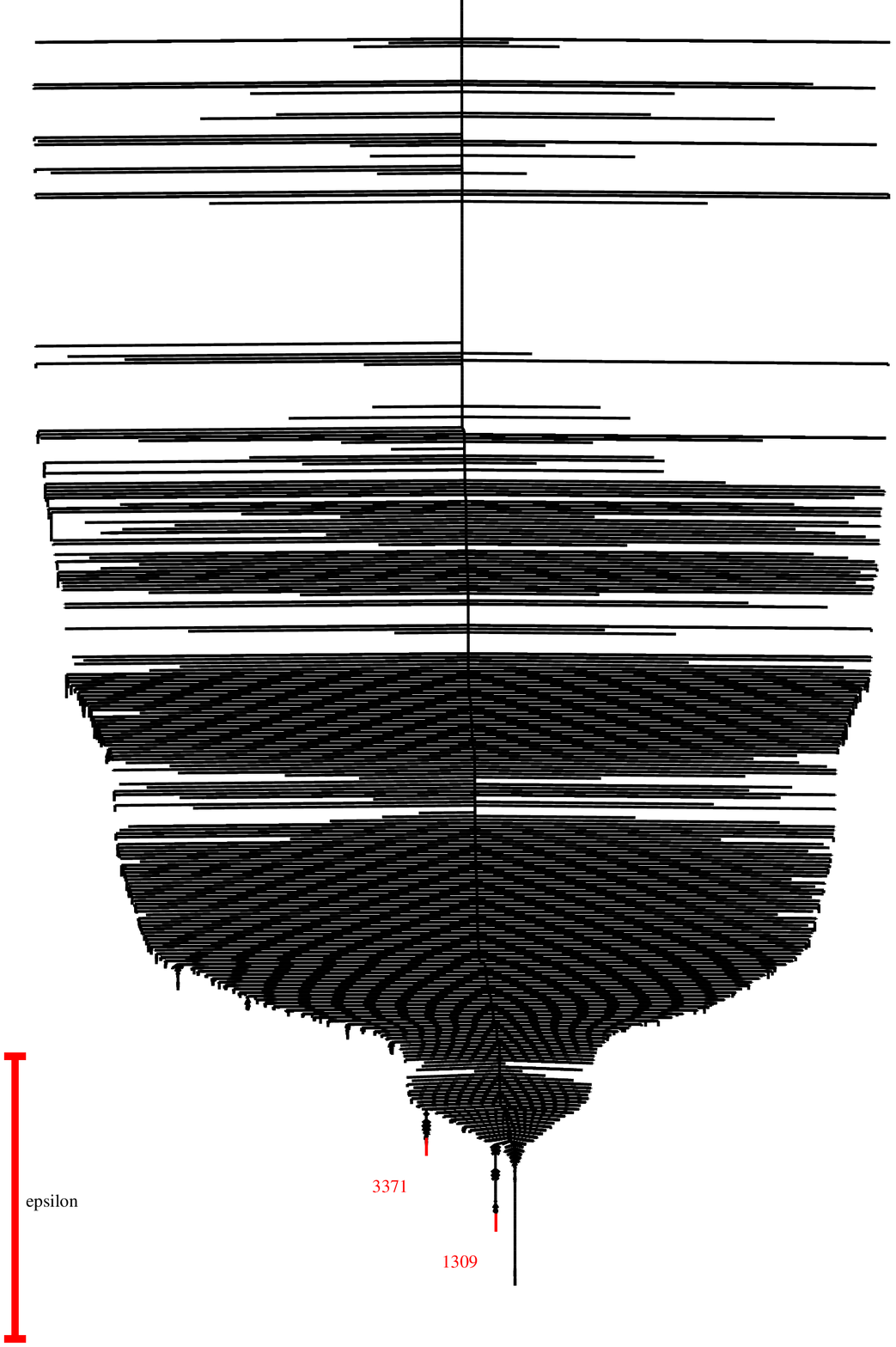}
        \caption{Original}
    \end{subfigure}%
    \begin{subfigure}[t]{0.4\textwidth}
        \centering
        \psfrag{epsilon}{\small{}}
        \psfrag{595}{\scriptsize{1}}
        \psfrag{9}{\scriptsize{2}}
        \includegraphics[width=1.0\textwidth]{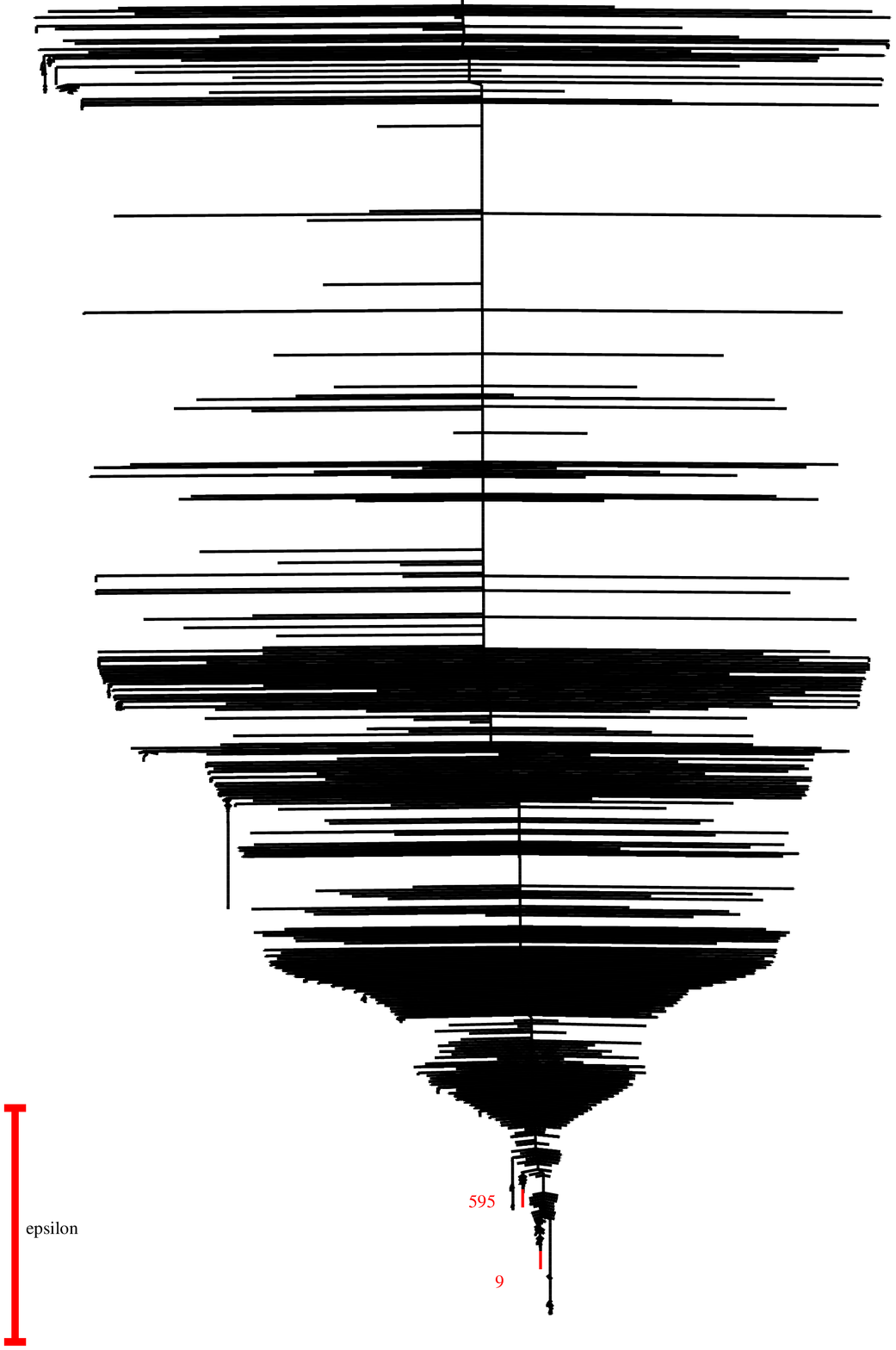}
        \caption{Outlier 1}
    \end{subfigure} \\
    \begin{subfigure}[t]{0.4\textwidth}
        \centering
        \psfrag{epsilon}{\small{}}
        \psfrag{503}{\scriptsize{1}}
        \psfrag{578}{\scriptsize{2}}
        \includegraphics[width=1.0\textwidth]{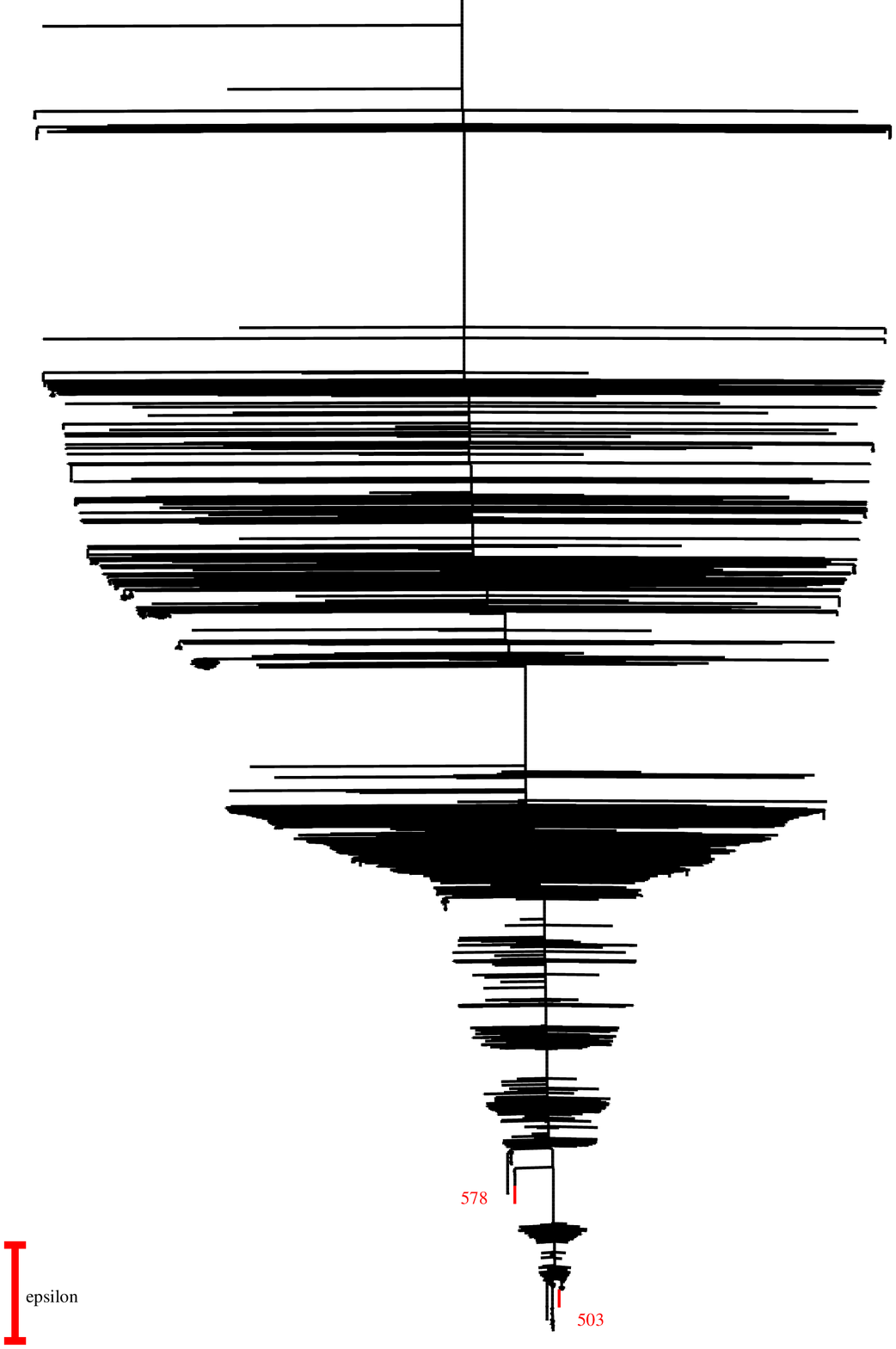}
        \caption{Outlier 2}
    \end{subfigure}%
    \begin{subfigure}[t]{0.4\textwidth}
        \centering
        \psfrag{epsilon}{\small{}}
        \psfrag{830}{\scriptsize{1}}
        \psfrag{239}{\scriptsize{2}}
        \includegraphics[width=1.0\textwidth]{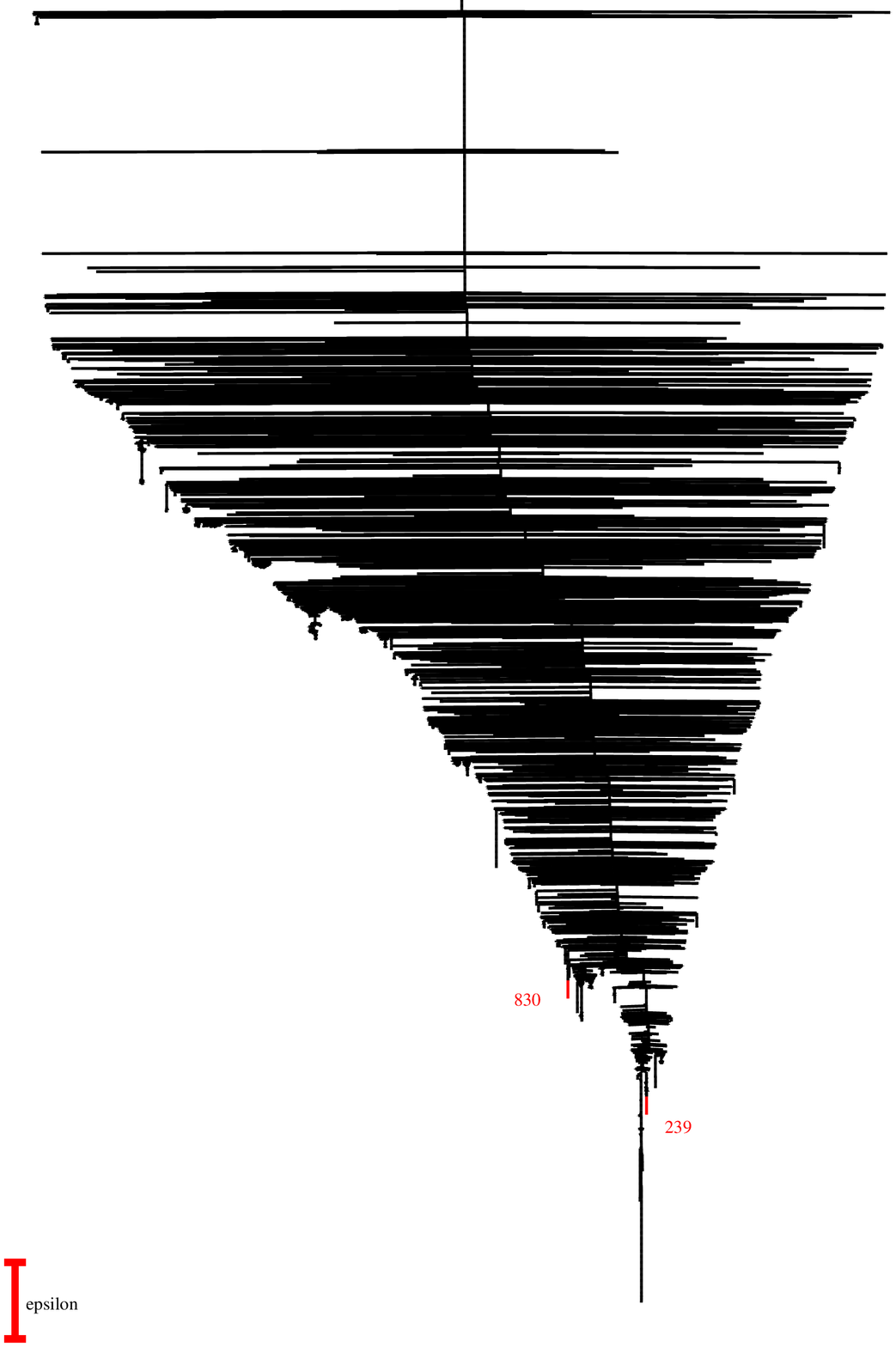}
        \caption{Outlier 3}
    \end{subfigure}%
    \caption{$K$-means landscapes for the glass datasets with varying outlier number. Selected kinetic traps are identified by red minima in the disconnectivity graph, with the associated label. The label corresponds to the kinetic traps given in Sec.~A.3 of the main text.}
    \label{OutliersGlassKTDGs}
\end{figure}

\clearpage

\subsection{Clustering solution properties}

\begin{figure}[h!]
    \centering
    \begin{subfigure}[t]{0.2\textwidth}
        \centering
        \psfrag{epsilon}{\small{25}}
        \psfrag{a}{\scriptsize{0}}
        \psfrag{b}{\scriptsize{1}}
        \includegraphics[width=1.0\textwidth]{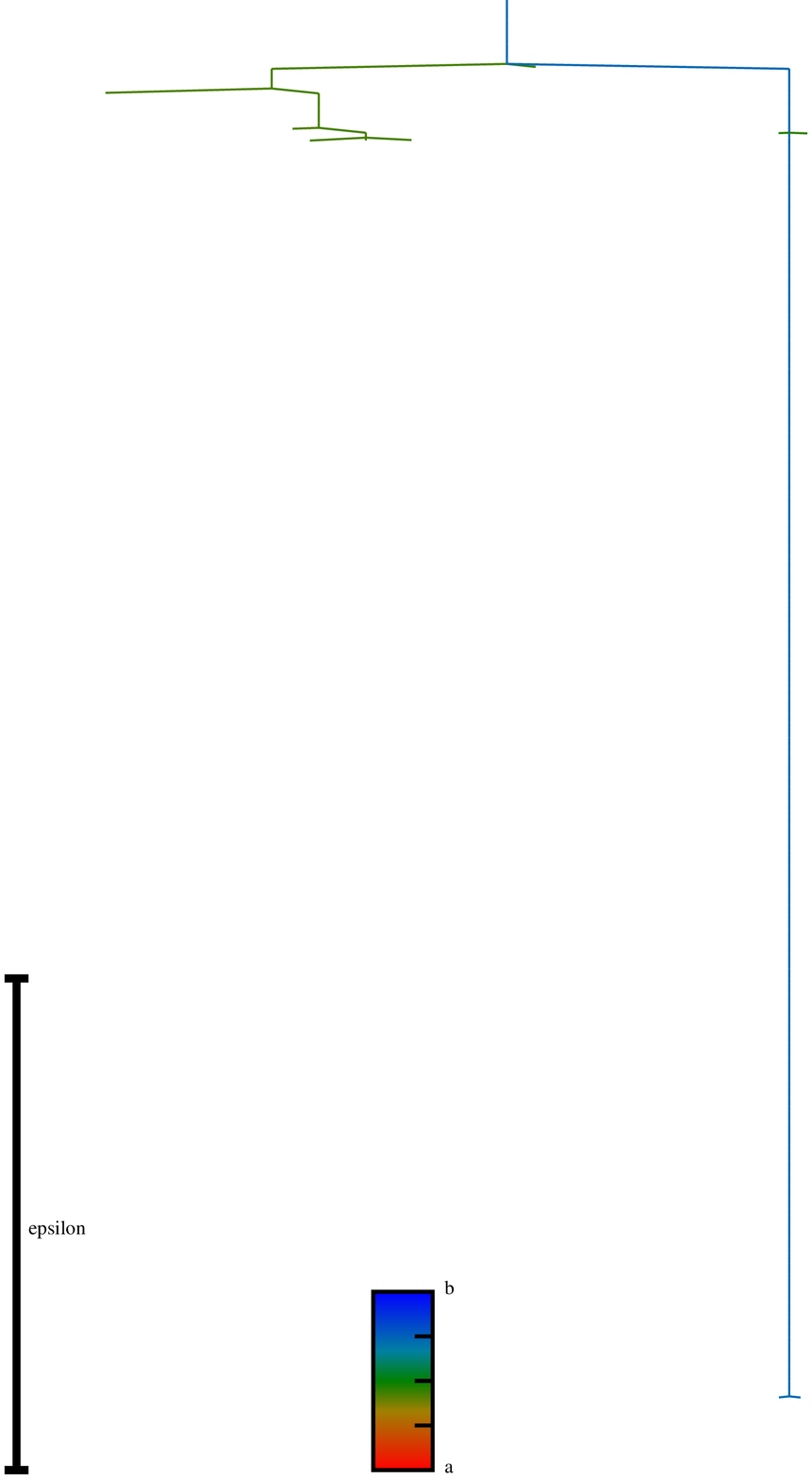}
        \caption{Original}
    \end{subfigure}%
    \begin{subfigure}[t]{0.2\textwidth}
        \centering
        \psfrag{epsilon}{\small{}}
        \psfrag{a}{\scriptsize{0}}
        \psfrag{b}{\scriptsize{1}}
        \includegraphics[width=1.0\textwidth]{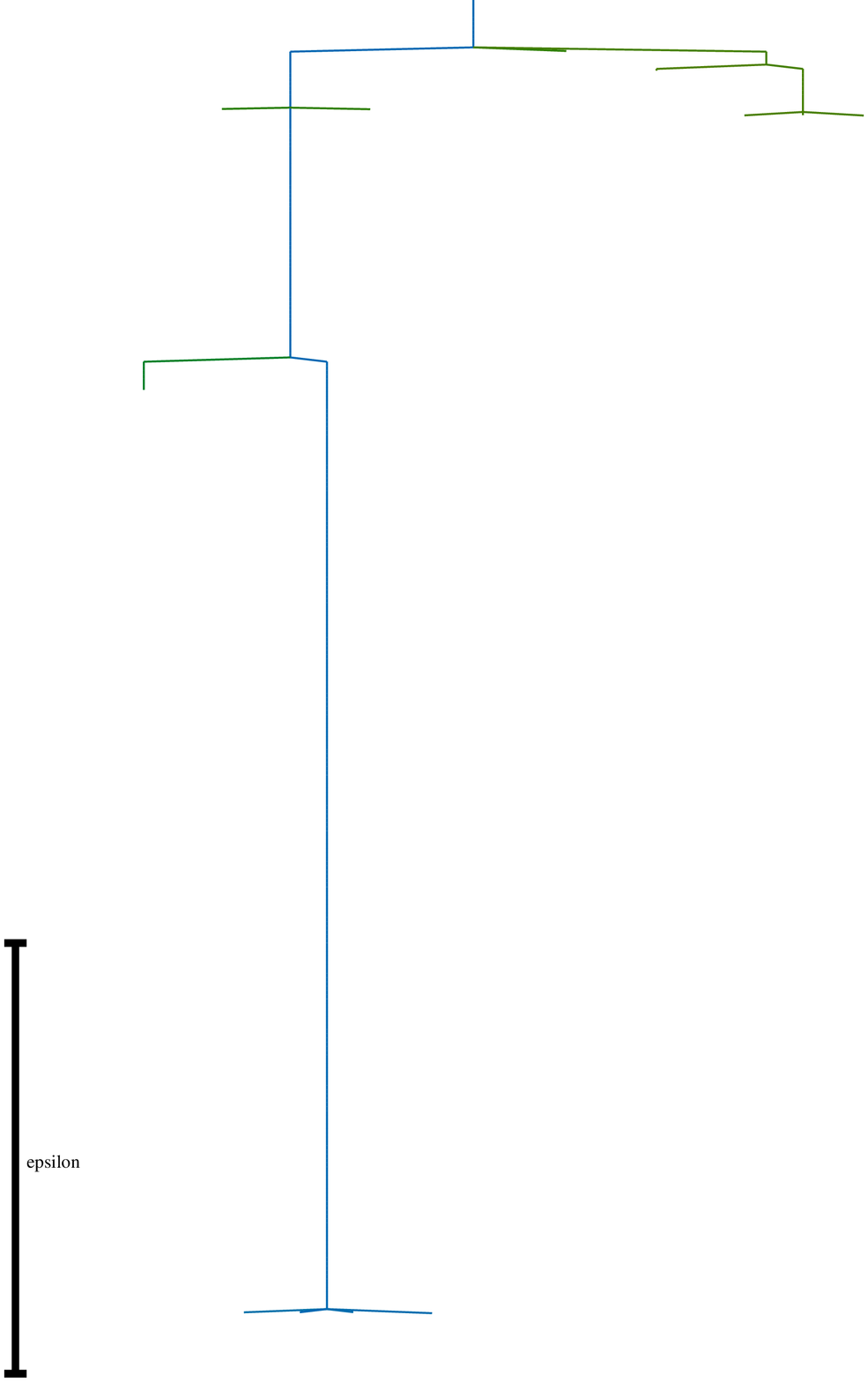}
        \caption{Outlier 1}
    \end{subfigure}%
    \begin{subfigure}[t]{0.2\textwidth}
        \centering
        \psfrag{epsilon}{\small{}}
        \psfrag{a}{\scriptsize{0}}
        \psfrag{b}{\scriptsize{1}}
        \includegraphics[width=1.0\textwidth]{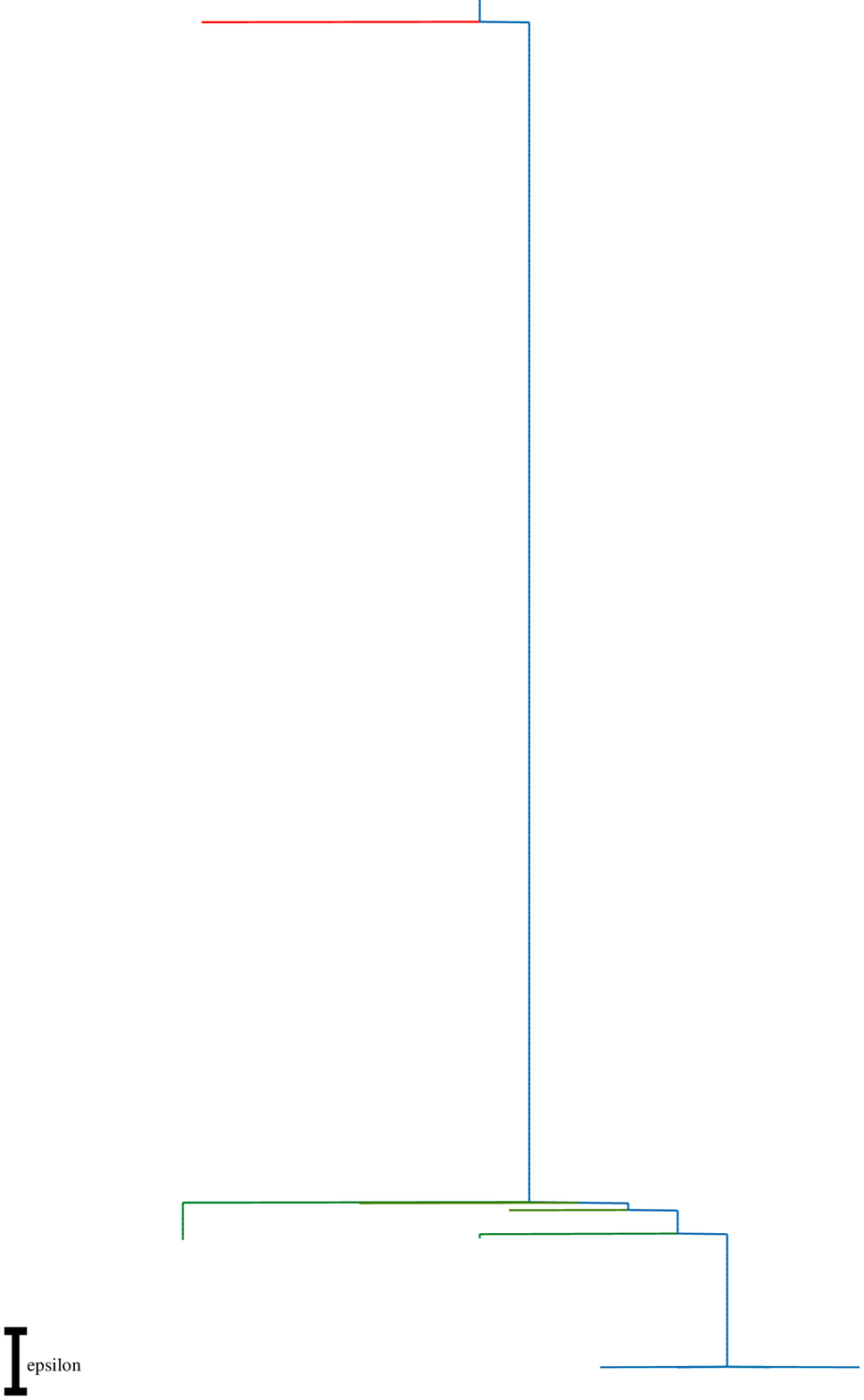}
        \caption{Outlier 2}
    \end{subfigure}%
    \begin{subfigure}[t]{0.2\textwidth}
        \centering
        \psfrag{epsilon}{\small{}}
        \psfrag{a}{\scriptsize{0}}
        \psfrag{b}{\scriptsize{1}}
        \includegraphics[width=1.0\textwidth]{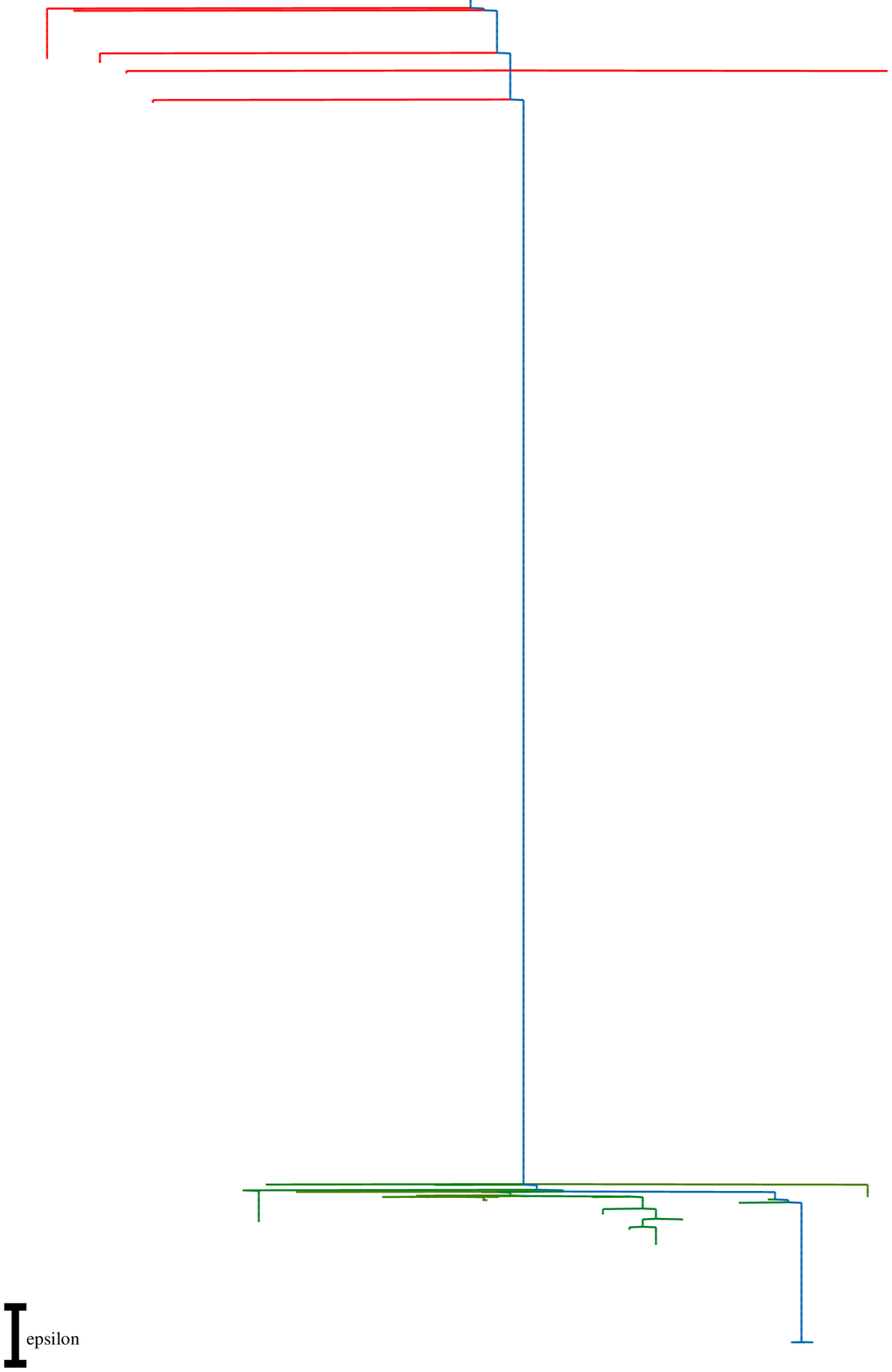}
        \caption{Outlier 3}
    \end{subfigure}%
    \begin{subfigure}[t]{0.2\textwidth}
        \centering
        \psfrag{epsilon}{\small{}}
        \psfrag{a}{\scriptsize{0}}
        \psfrag{b}{\scriptsize{1}}
        \includegraphics[width=1.0\textwidth]{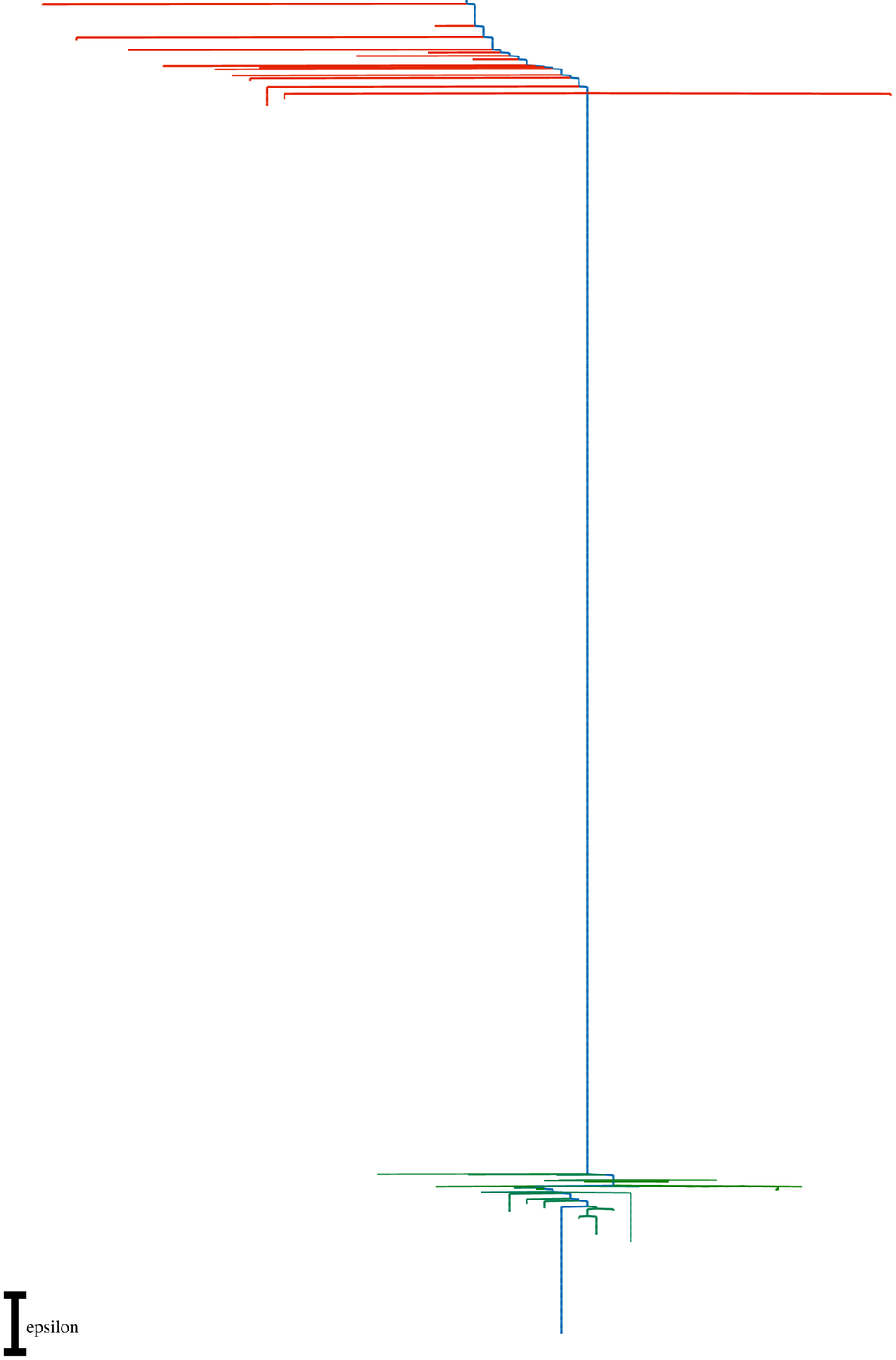}
        \caption{Outlier 4}
    \end{subfigure} \\
    \caption{Disconnectivity graphs for $K$-means landscapes of Fisher's
\textit{Iris} dataset with $K=3$ for the original dataset, and datasets with
outliers. The minima are coloured according to the ARI relative to the ground
truth labels. The scalebar and colour scale are the same for all panels in the
same row.}
    \label{OutliersIris3PropertyDGs}
\end{figure}

\begin{figure}[h!]
    \centering
    \begin{subfigure}[t]{0.4\textwidth}
        \centering
        \psfrag{epsilon}{\small{80}}
        \psfrag{a}{\scriptsize{0}}
        \psfrag{b}{\scriptsize{0.35}}
        \includegraphics[width=1.0\textwidth]{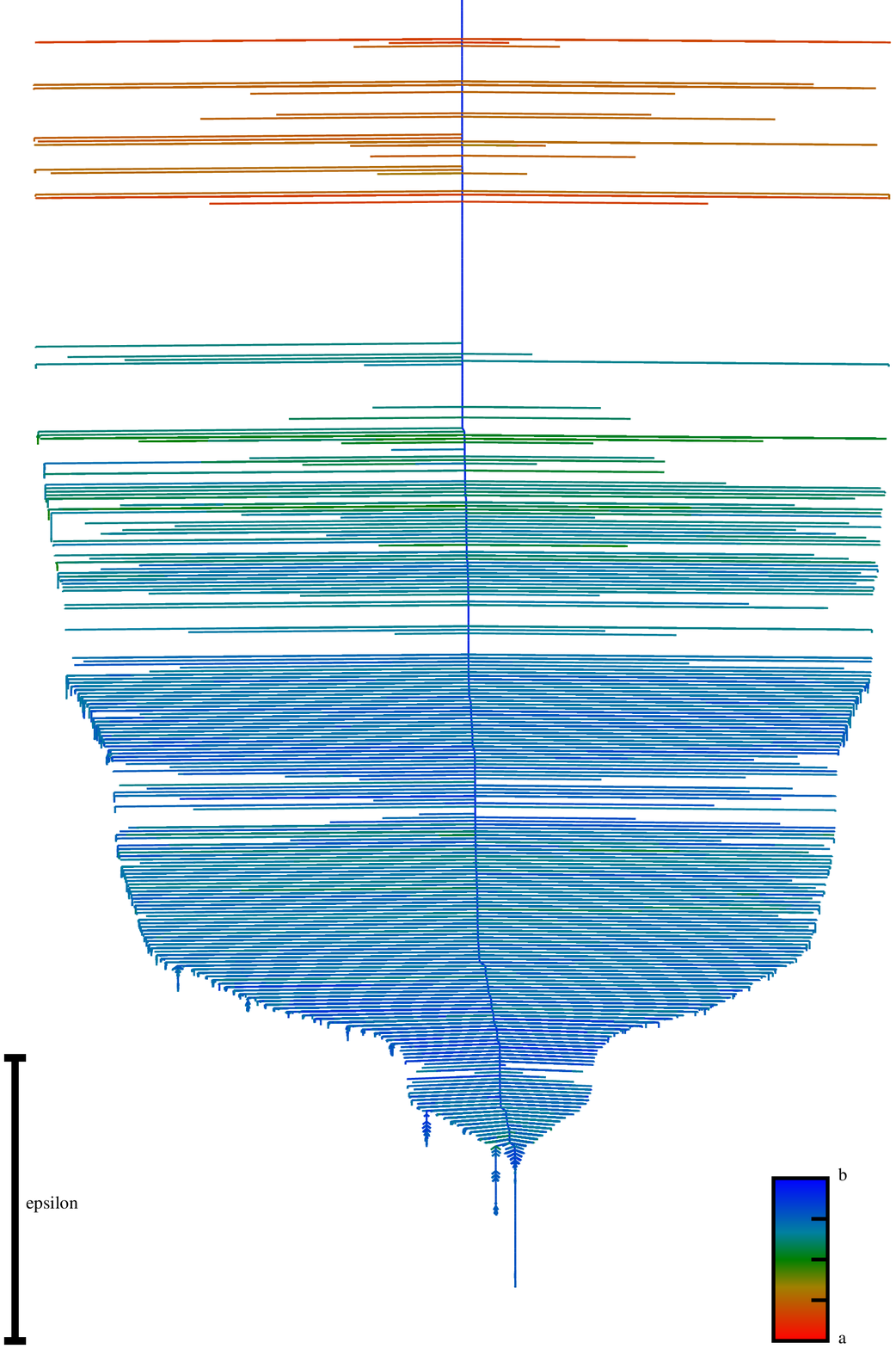}
        \caption{Original}
    \end{subfigure}%
    \begin{subfigure}[t]{0.4\textwidth}
        \centering
        \psfrag{epsilon}{\small{}}
        \psfrag{a}{\scriptsize{0}}
        \psfrag{b}{\scriptsize{0.35}}
        \includegraphics[width=1.0\textwidth]{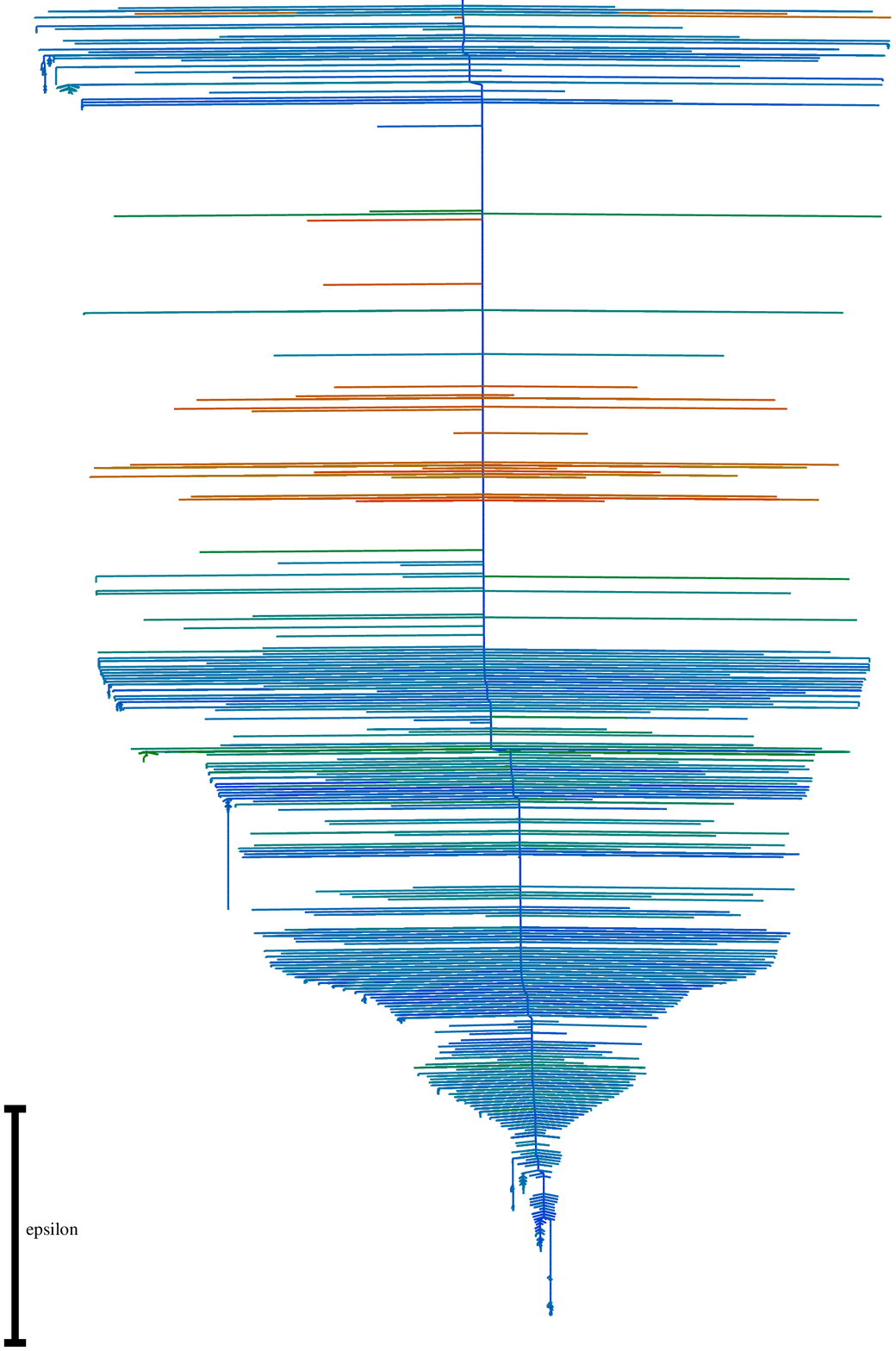}
        \caption{Outlier 1}
    \end{subfigure} \\
    \begin{subfigure}[t]{0.4\textwidth}
        \centering
        \psfrag{epsilon}{\small{}}
        \psfrag{a}{\scriptsize{0}}
        \psfrag{b}{\scriptsize{0.35}}
        \includegraphics[width=1.0\textwidth]{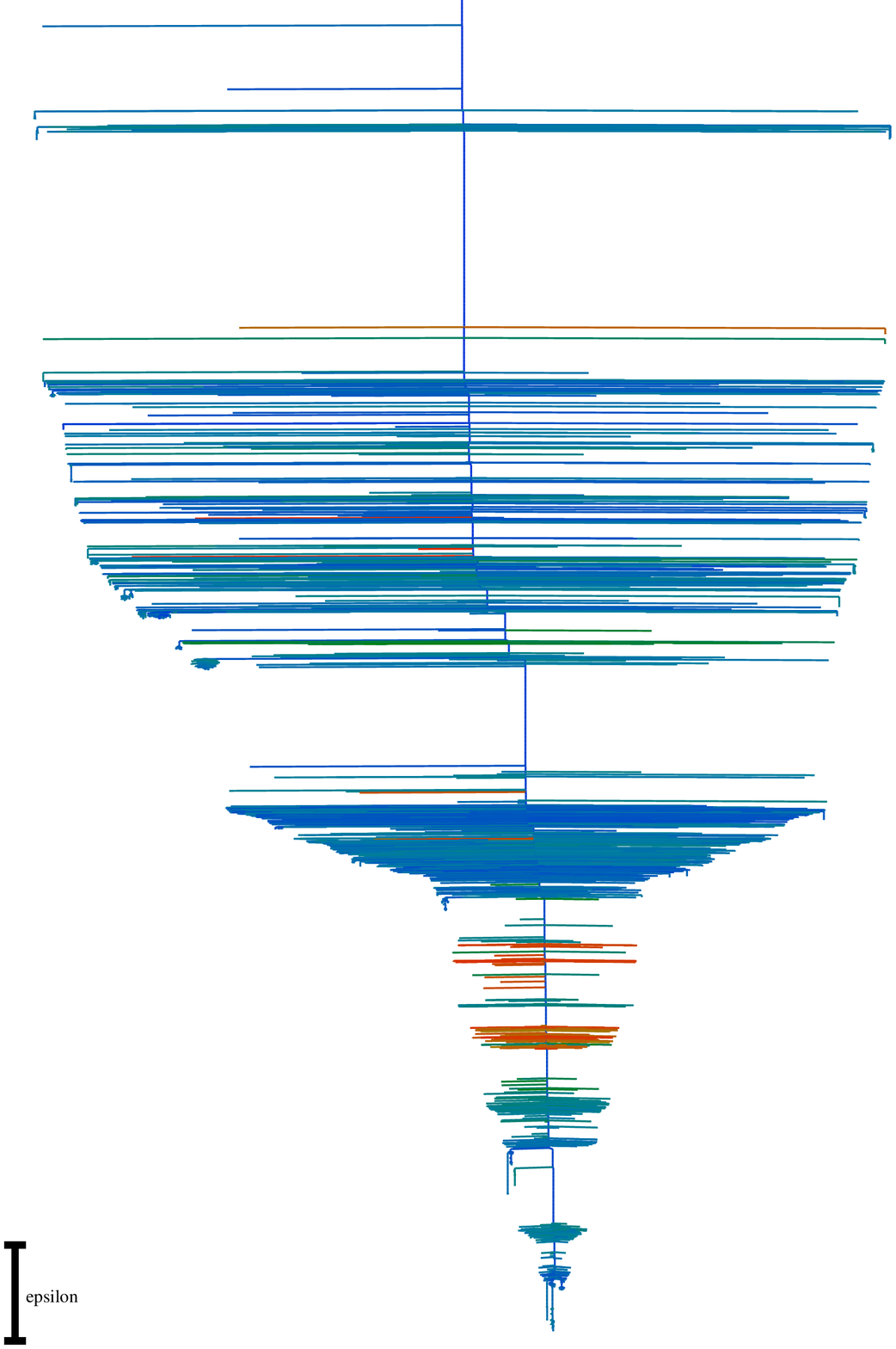}
        \caption{Outlier 2}
    \end{subfigure}%
    \begin{subfigure}[t]{0.4\textwidth}
        \centering
        \psfrag{epsilon}{\small{}}
        \psfrag{a}{\scriptsize{0}}
        \psfrag{b}{\scriptsize{0.35}}
        \includegraphics[width=1.0\textwidth]{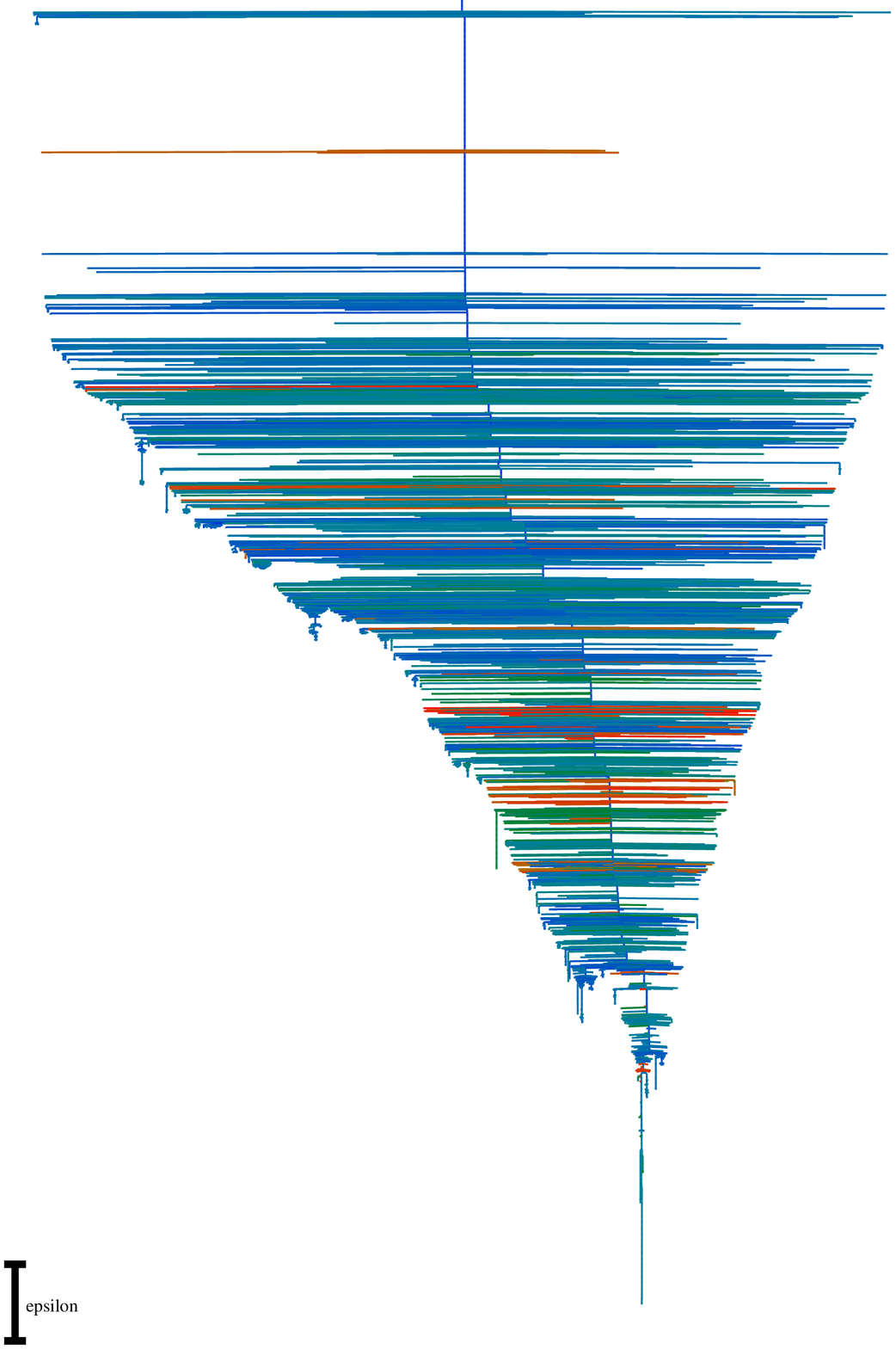}
        \caption{Outlier 3}
    \end{subfigure} \\
    \caption{Disconnectivity graphs showing the $K$-means landscapes for the
glass datasets with an increasing number of outliers. The scale bar represents
the same range in all plots. The colouring of minima is defined by the ARI
between the cluster labels of each minimum and the ground truth labels. The
colour range is given in (a) and remains the same in all plots.}
    \label{OutliersGlassAccuracyDGs}
\end{figure}

\begin{figure}[h!]
    \centering
    \begin{subfigure}[t]{0.33\textwidth}
        \centering
        \includegraphics[width=1.00\textwidth]{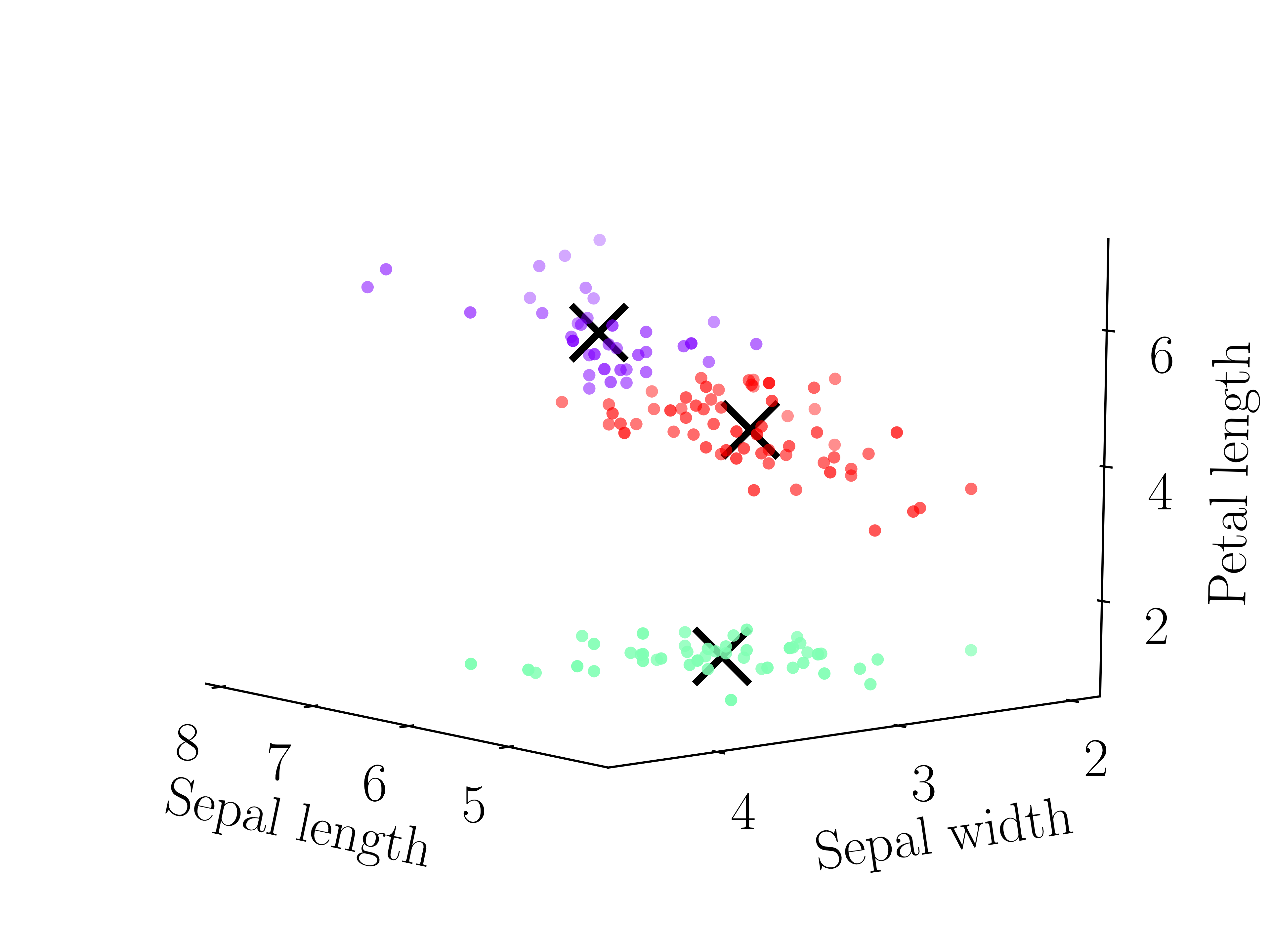}
        \caption{Original}
    \end{subfigure}%
    \begin{subfigure}[t]{0.33\textwidth}
        \centering
        \includegraphics[width=1.00\textwidth]{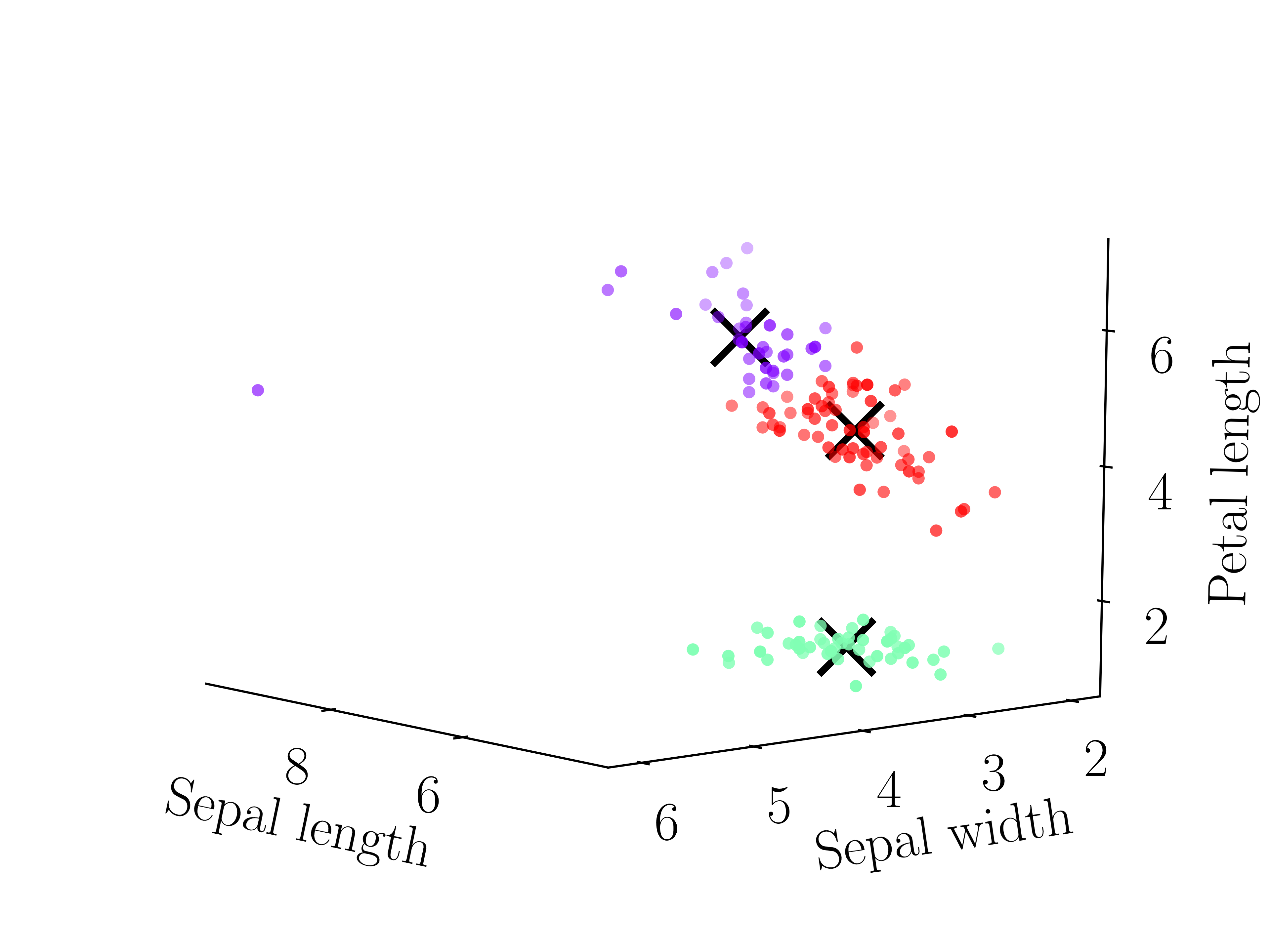}
        \caption{Outlier 1}
    \end{subfigure}%
    \begin{subfigure}[t]{0.33\textwidth}
        \centering
        \includegraphics[width=1.00\textwidth]{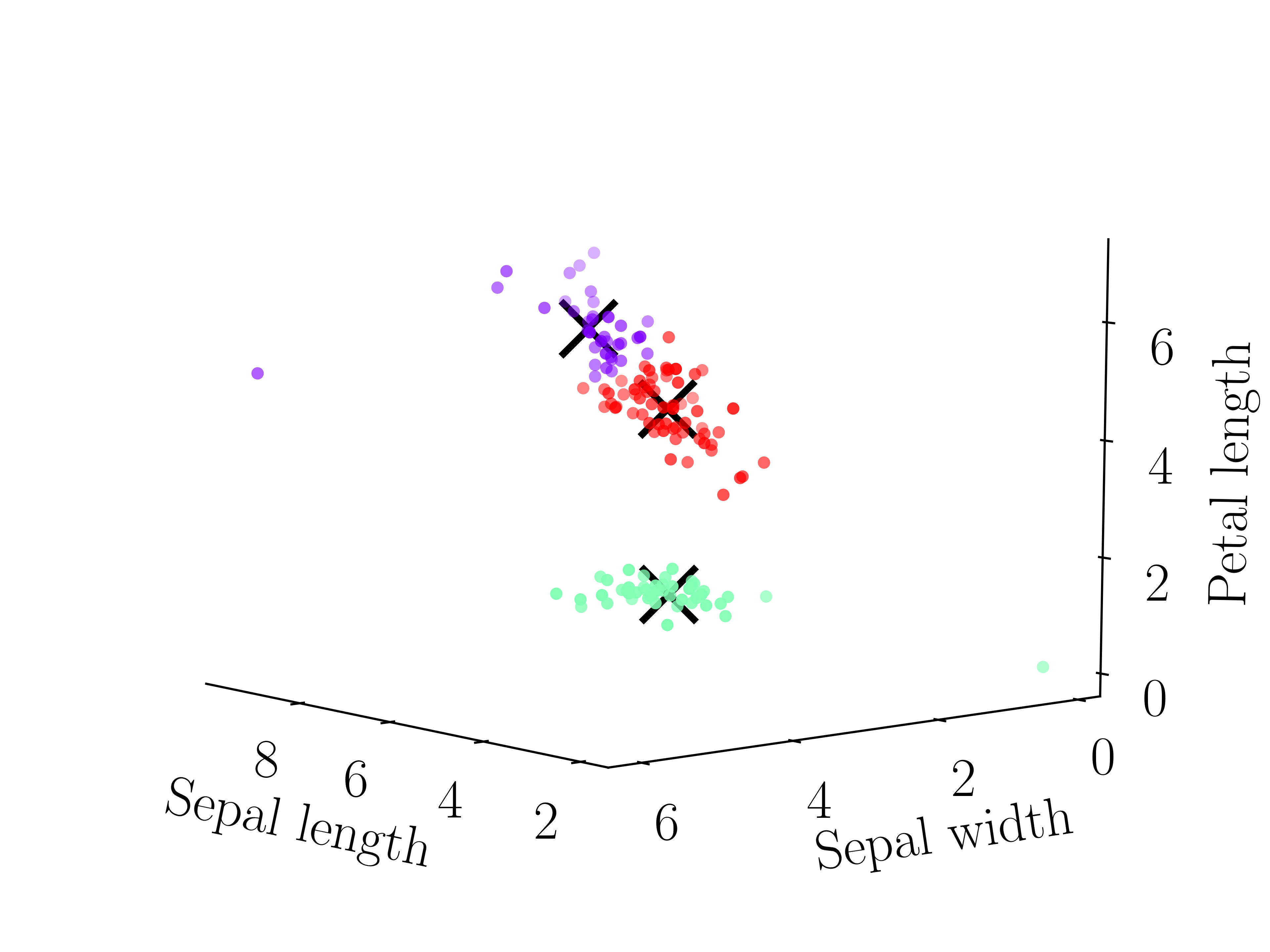}
        \caption{Outlier 2}
    \end{subfigure} \\
    \begin{subfigure}[t]{0.35\textwidth}
        \centering
        \includegraphics[width=1.00\textwidth]{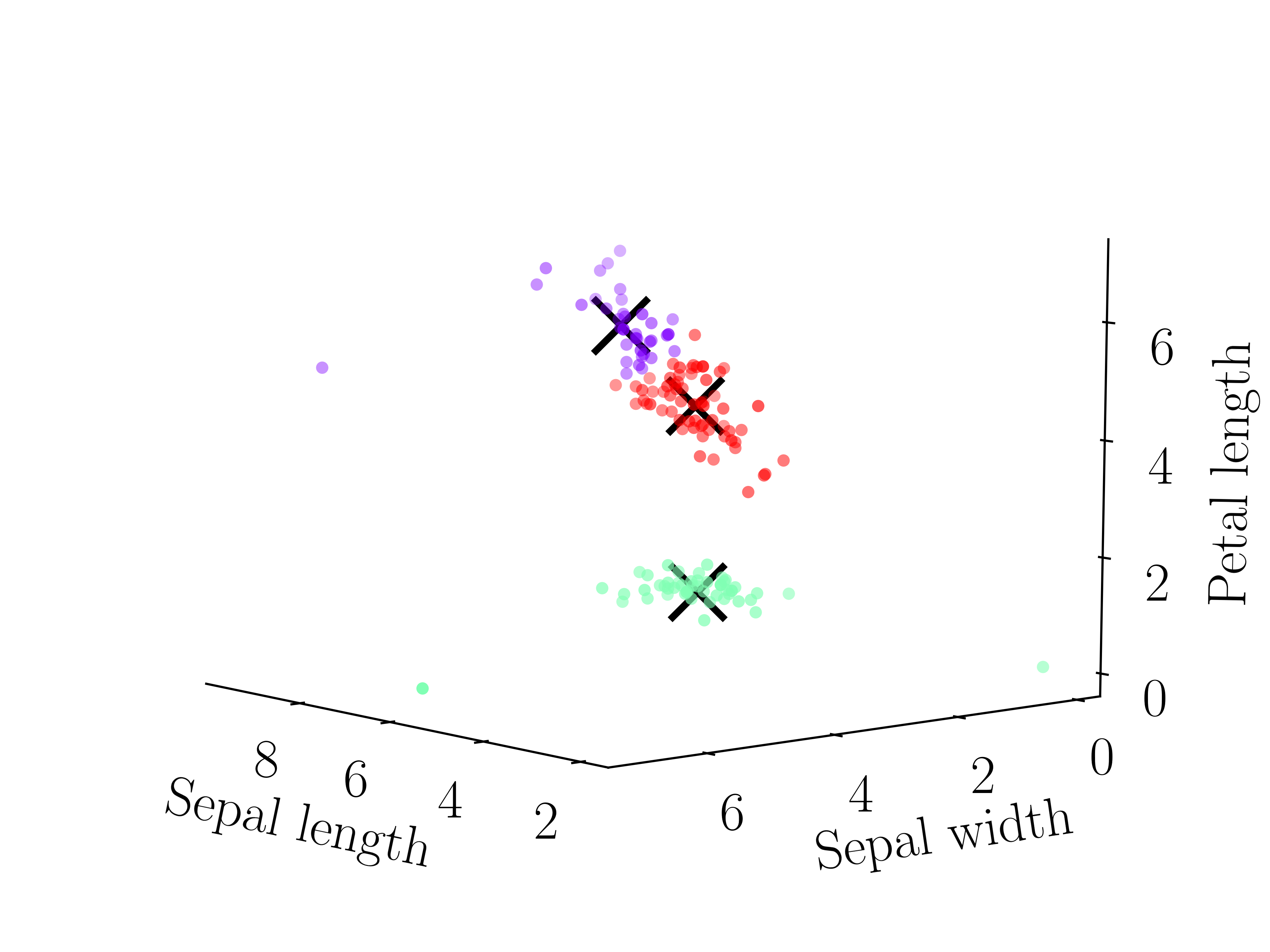}
        \caption{Outlier 3}
    \end{subfigure}%
    \begin{subfigure}[t]{0.35\textwidth}
        \centering
        \includegraphics[width=1.00\textwidth]{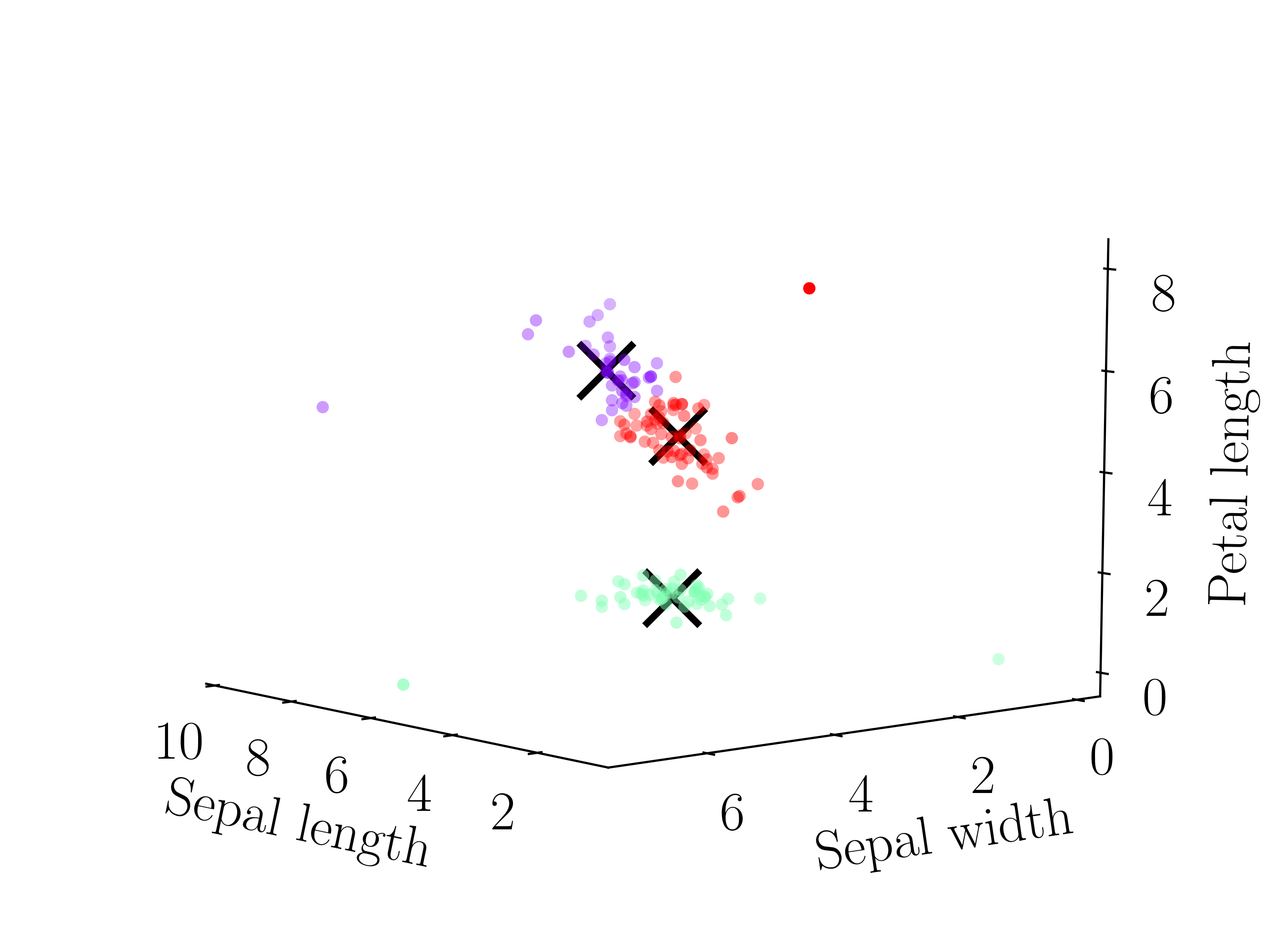}
        \caption{Outlier 4}
    \end{subfigure}%
    \caption{Global minima for the \textit{Iris} landscapes with $K=3$. Three
of the four features are presented to allow visualisation, given in
centimetres. Petal width is excluded as it contains little information. Cluster
centres are denoted by large black crosses, and the assignment of the data
points to each of the three clusters is given by their colour.}
    \label{OutliersIris3Minima}
\end{figure}

\begin{figure}[h!]
    \centering
    \begin{subfigure}[t]{0.33\textwidth}
        \centering
        \psfrag{epsilon}{\small{10}}
        \psfrag{a}{\scriptsize{1}}
        \psfrag{b}{\scriptsize{4}}
        \includegraphics[width=1.0\textwidth]{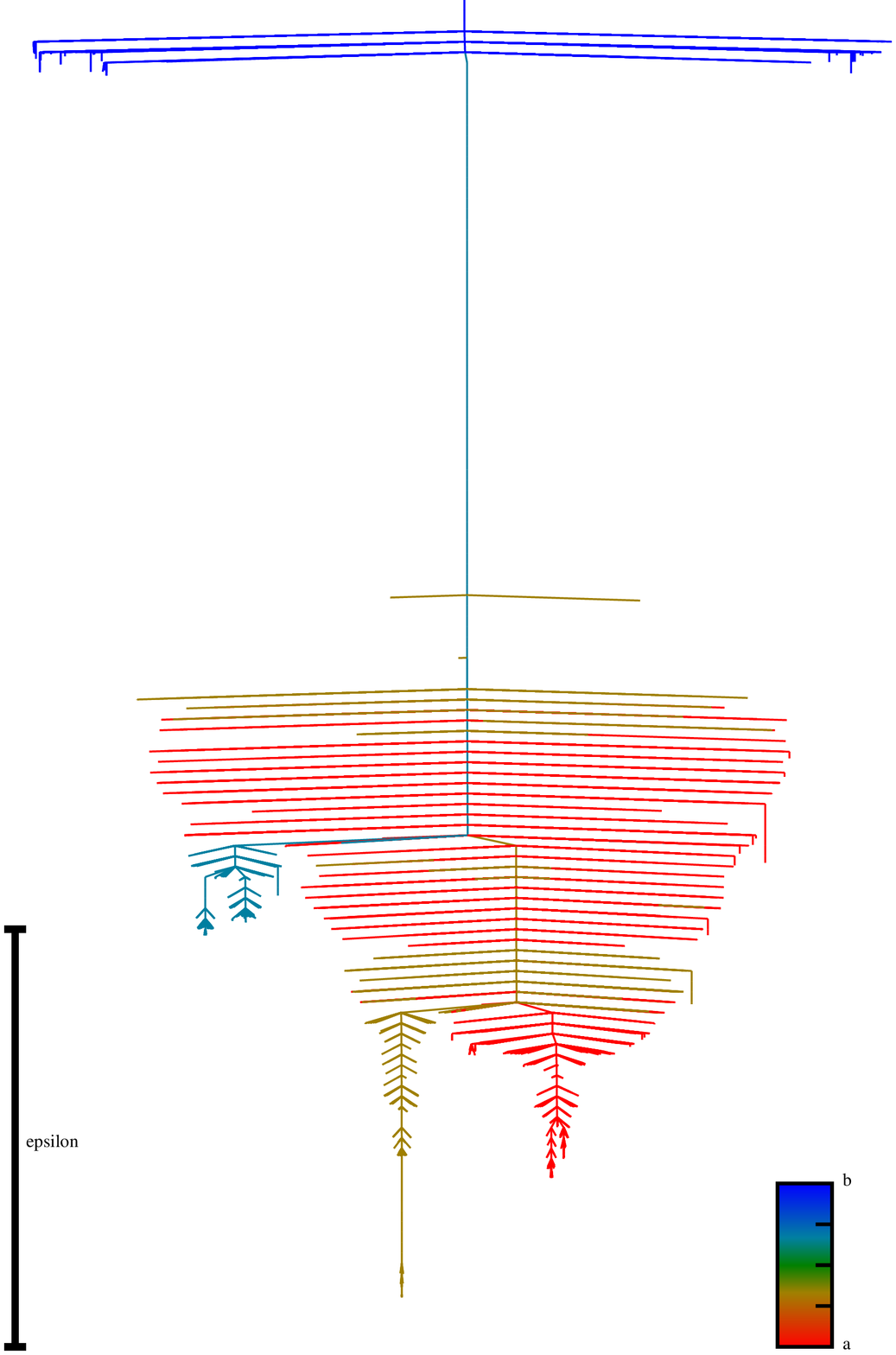}
        \caption{Original}
    \end{subfigure}%
    \begin{subfigure}[t]{0.33\textwidth}
        \centering
        \psfrag{epsilon}{\small{}}
        \psfrag{a}{\scriptsize{0}}
        \psfrag{b}{\scriptsize{1}}
        \includegraphics[width=1.0\textwidth]{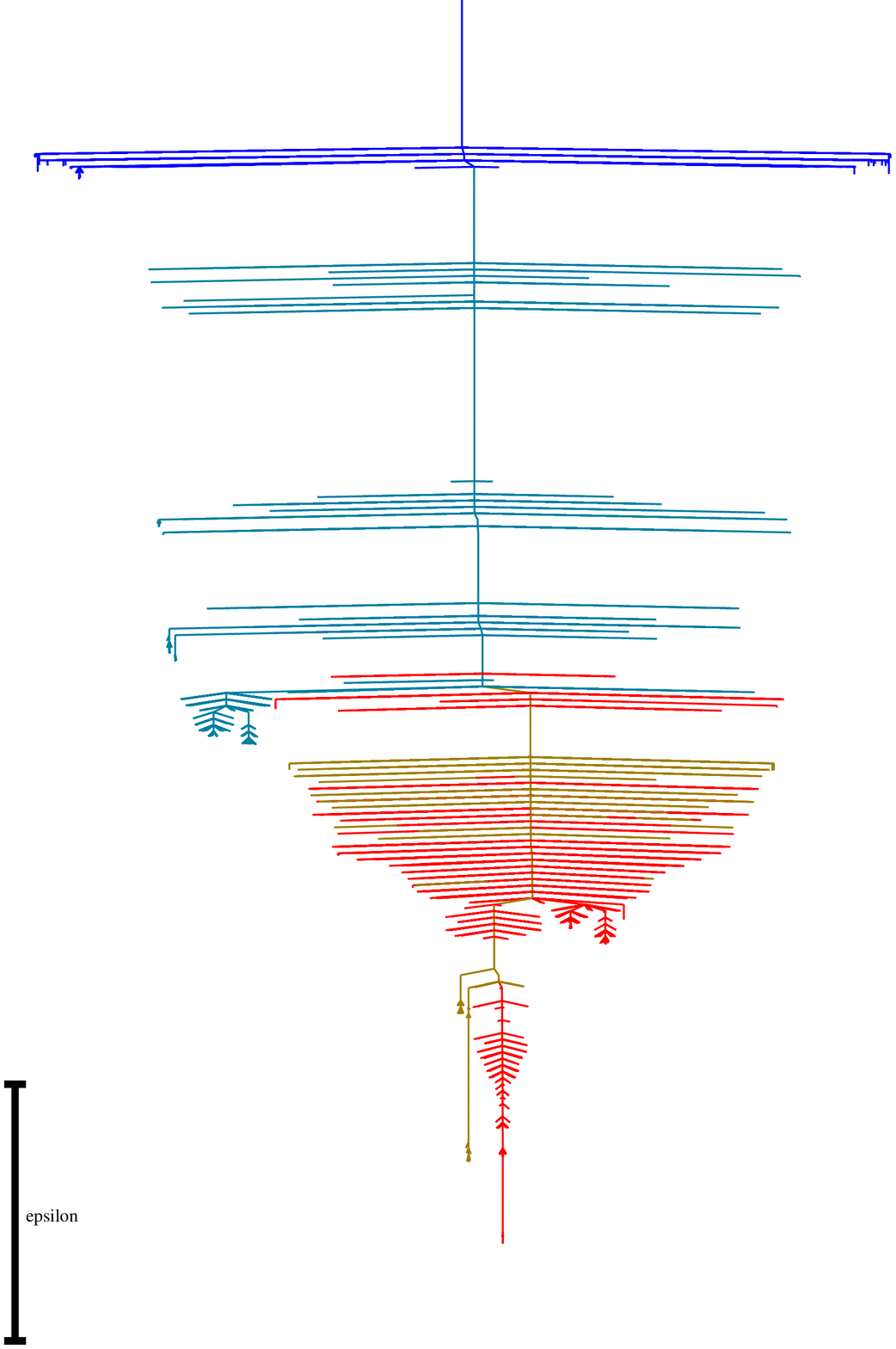}
        \caption{Outlier 1}
    \end{subfigure}%
    \begin{subfigure}[t]{0.33\textwidth}
        \centering
        \psfrag{epsilon}{\small{}}
        \psfrag{a}{\scriptsize{0}}
        \psfrag{b}{\scriptsize{1}}
        \includegraphics[width=1.0\textwidth]{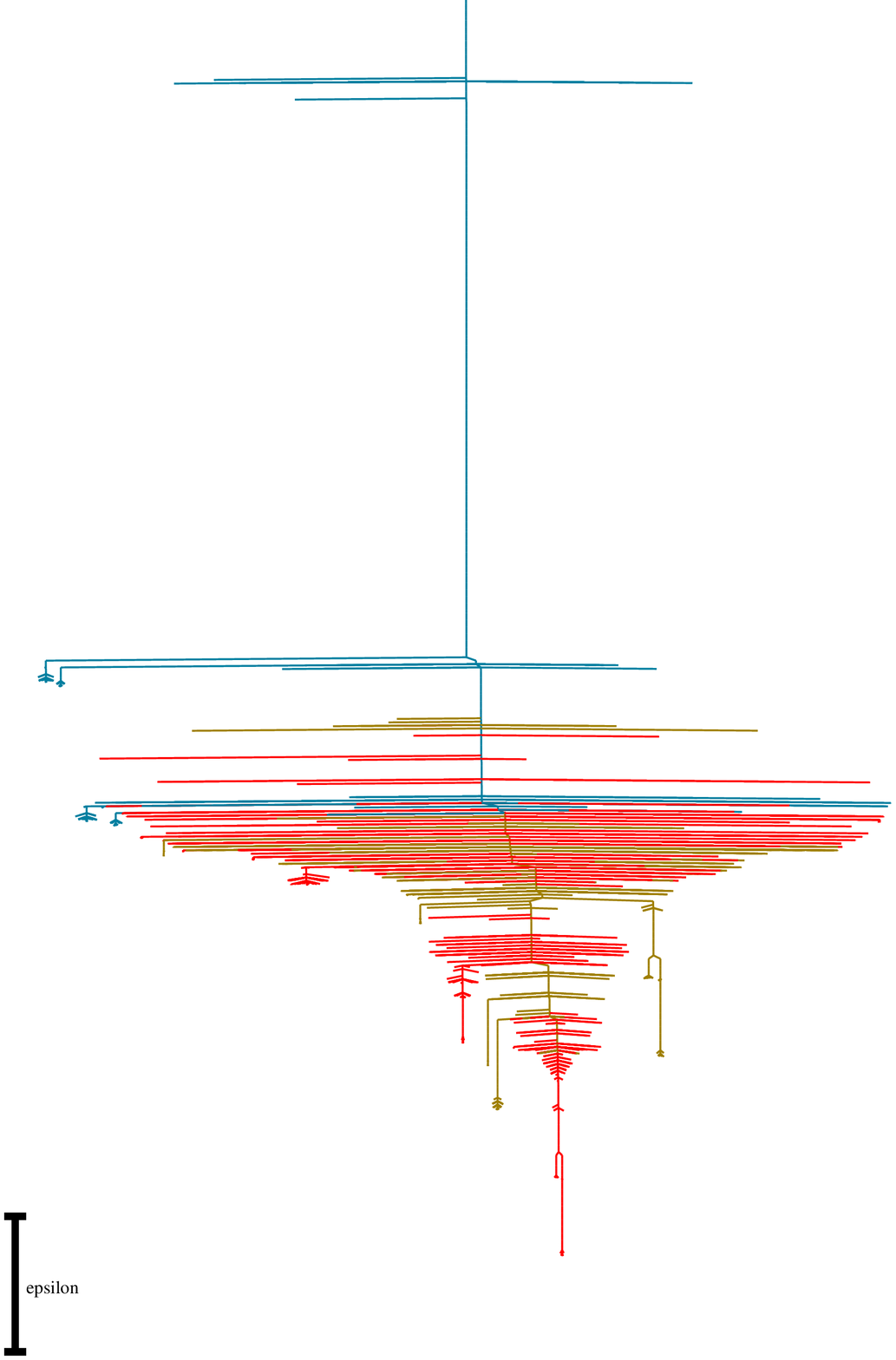}
        \caption{Outlier 2}
    \end{subfigure} \\
    \begin{subfigure}[t]{0.33\textwidth}
        \centering
        \psfrag{epsilon}{\small{}}
        \psfrag{a}{\scriptsize{0}}
        \psfrag{b}{\scriptsize{1}}
        \includegraphics[width=1.0\textwidth]{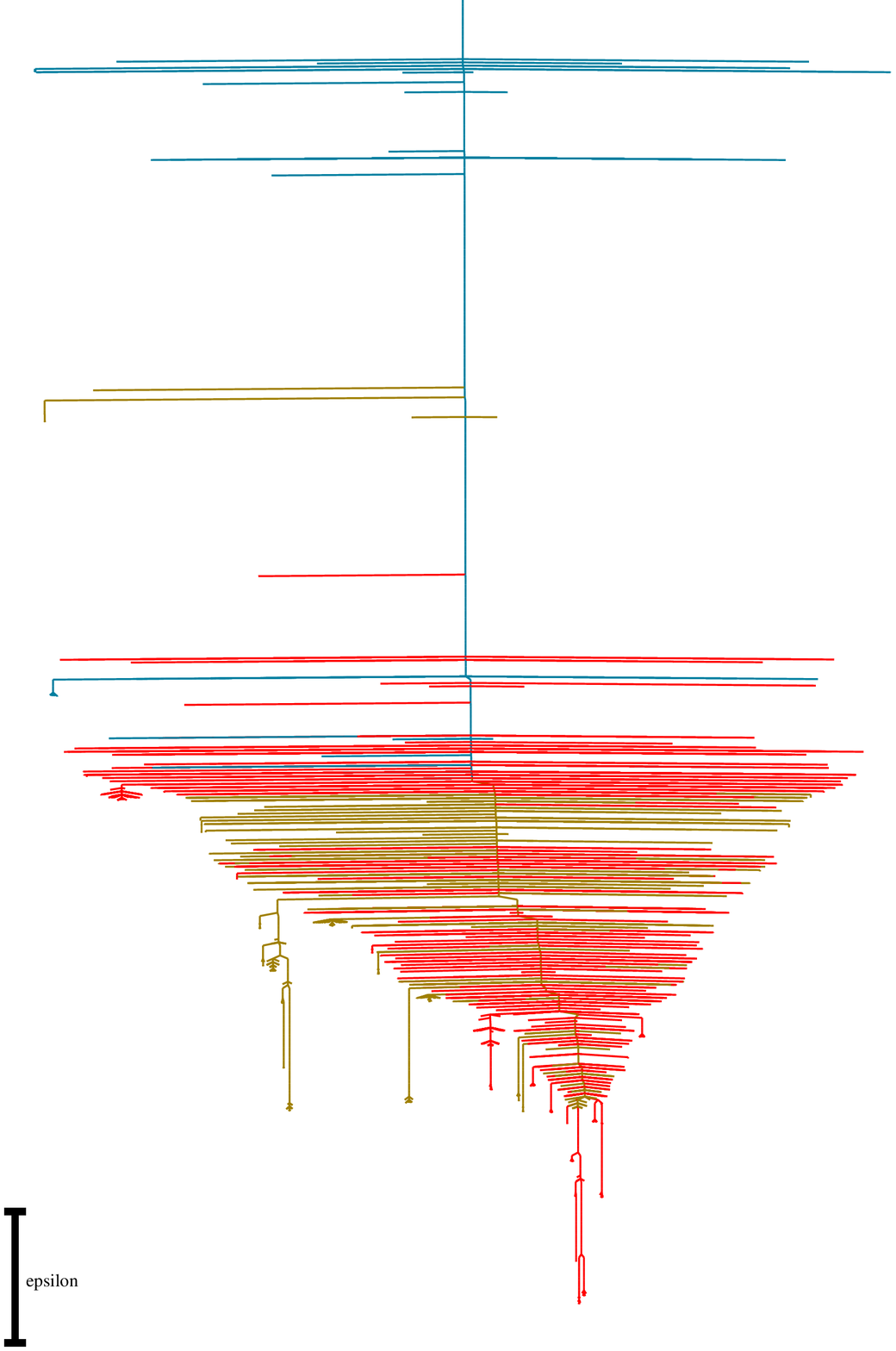}
        \caption{Outlier 3}
    \end{subfigure}%
    \begin{subfigure}[t]{0.33\textwidth}
        \centering
        \psfrag{epsilon}{\small{}}
        \psfrag{a}{\scriptsize{0}}
        \psfrag{b}{\scriptsize{1}}
        \includegraphics[width=1.0\textwidth]{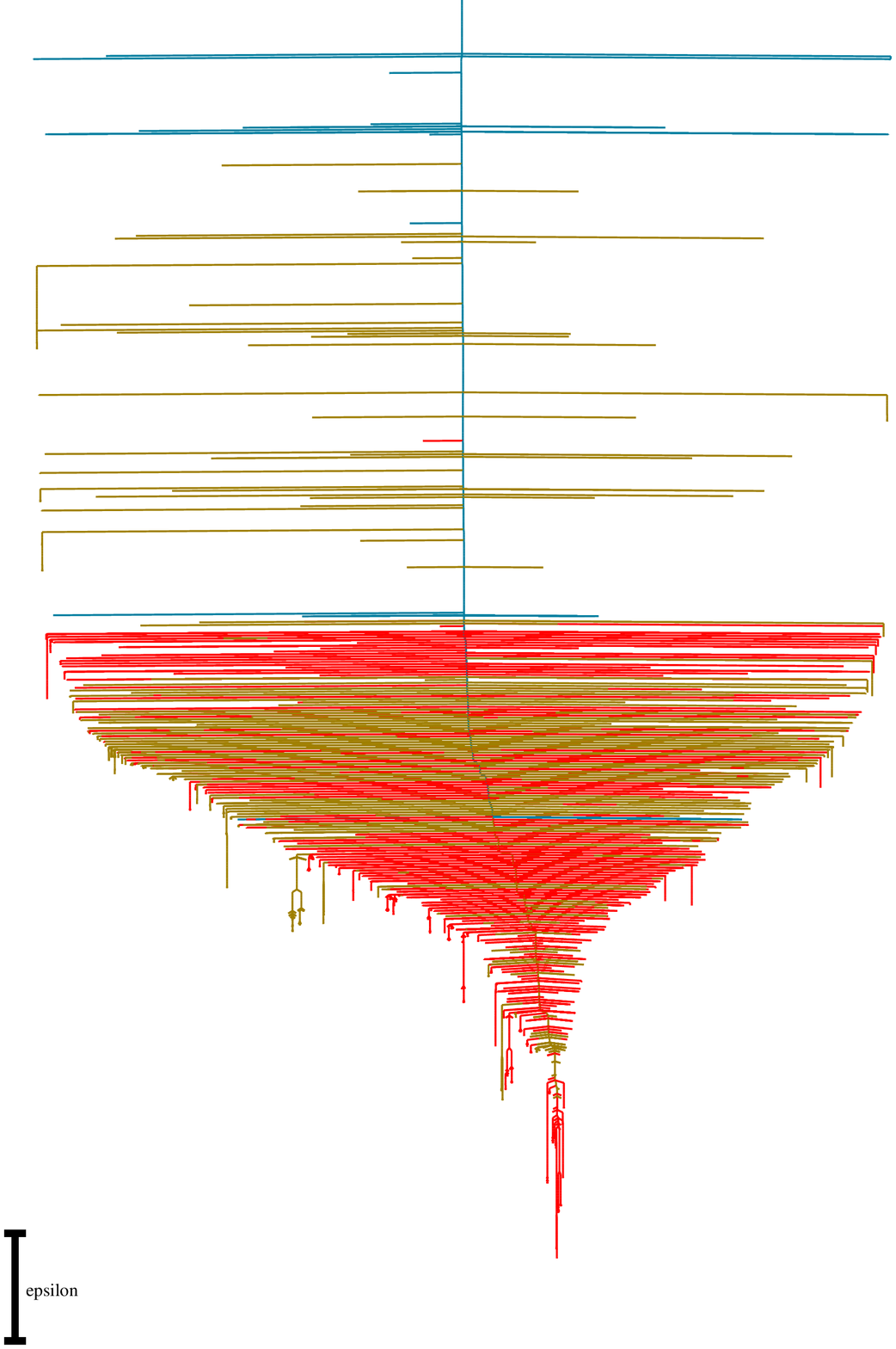}
        \caption{Outlier 4}
    \end{subfigure} \\
    \caption{Disconnectivity graphs for $K$-means landscapes of Fisher's \textit{Iris} dataset with six clusters. Minima are coloured according to the number of clusters  assigned to the \textit{Setosa} data points. The cost function and colour scale are the same in all plots.}
    \label{OutliersIris6PartitionDGs}
\end{figure}

\begin{figure}[h!]
    \centering
    \begin{subfigure}[t]{0.33\textwidth}
        \centering
        \includegraphics[width=1.00\textwidth]{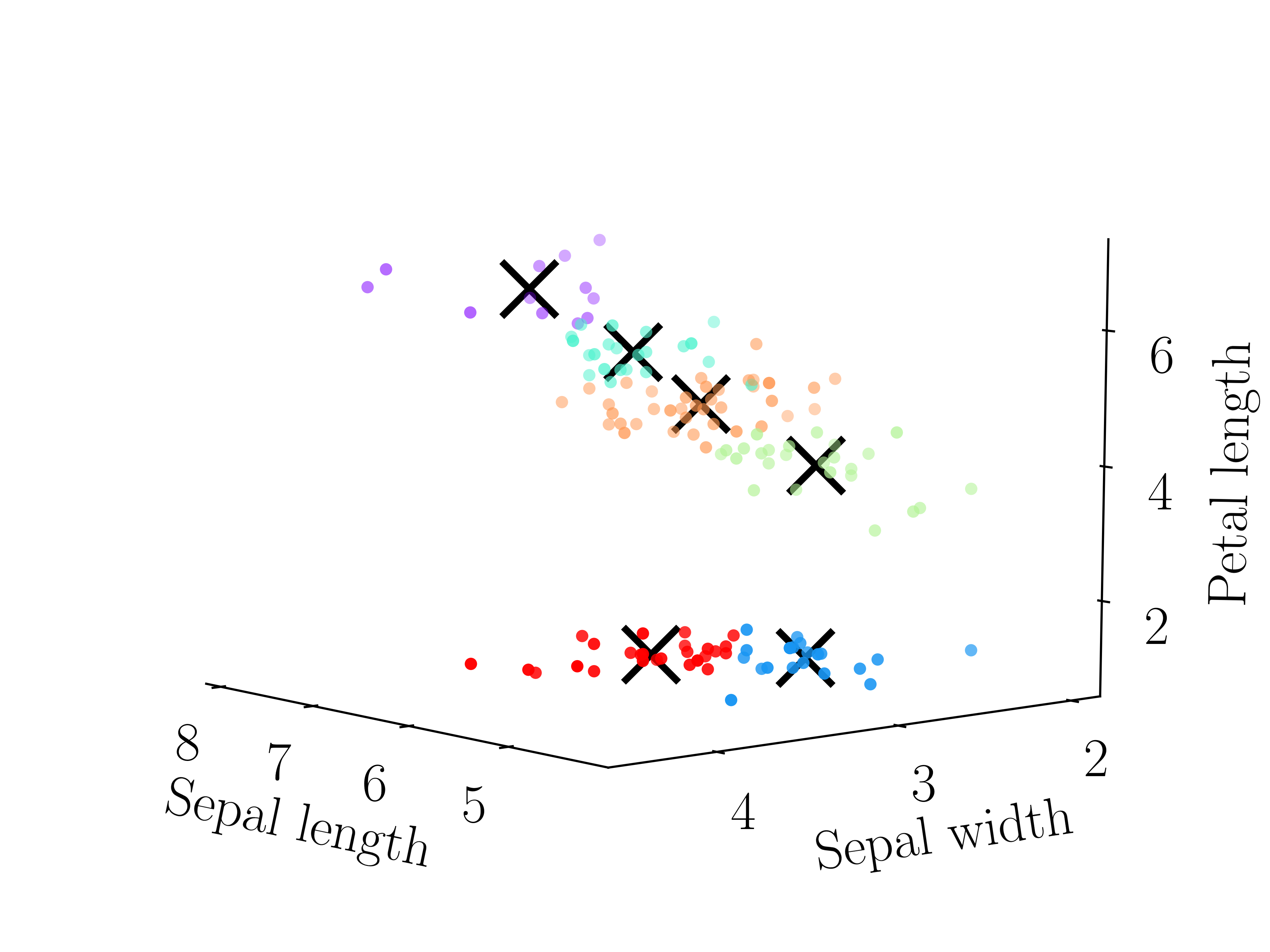}
        \caption{Original}
    \end{subfigure}%
    \begin{subfigure}[t]{0.33\textwidth}
        \centering
        \includegraphics[width=1.00\textwidth]{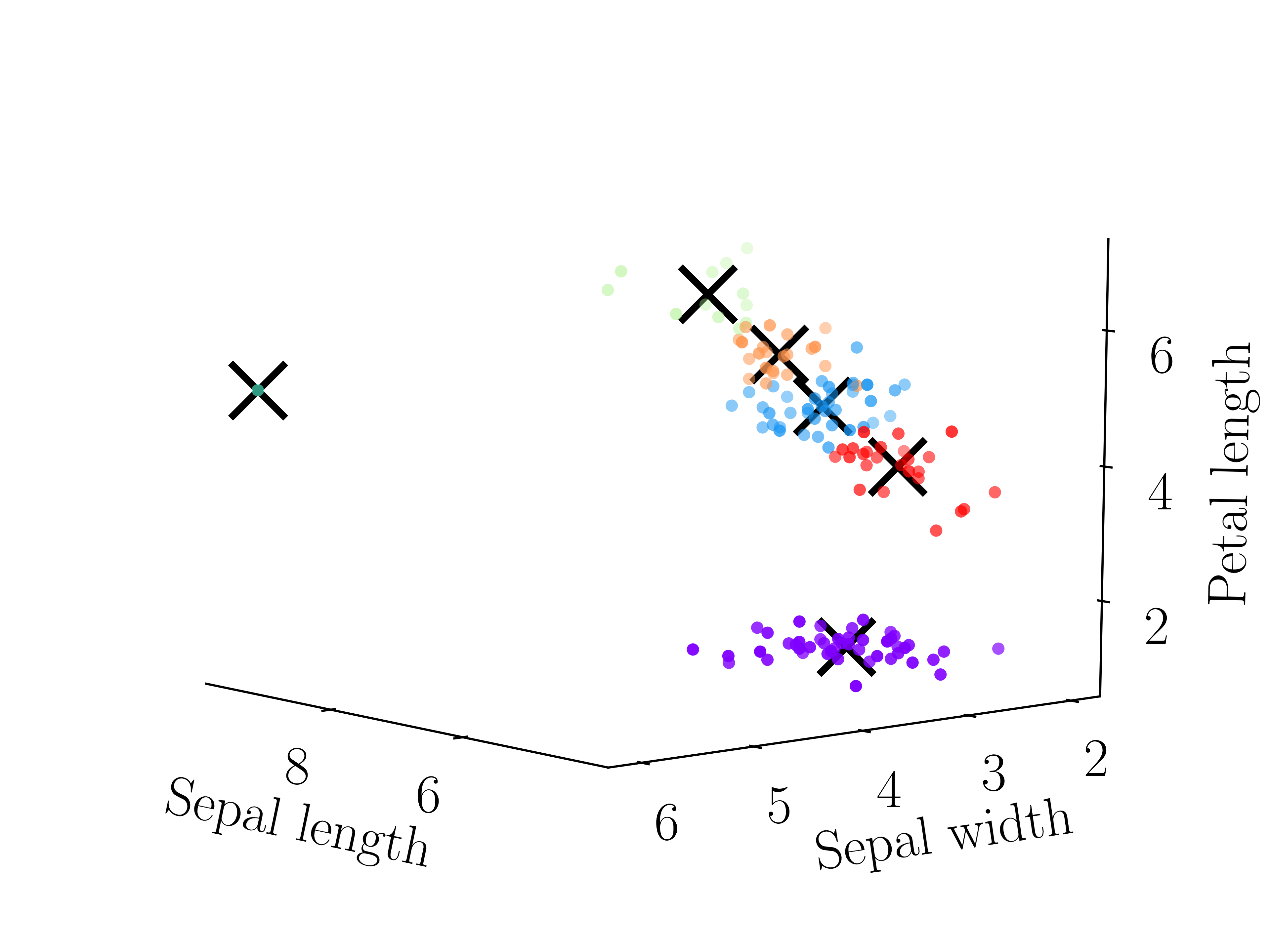}
        \caption{Outlier 1}
    \end{subfigure}%
    \begin{subfigure}[t]{0.33\textwidth}
        \centering
        \includegraphics[width=1.00\textwidth]{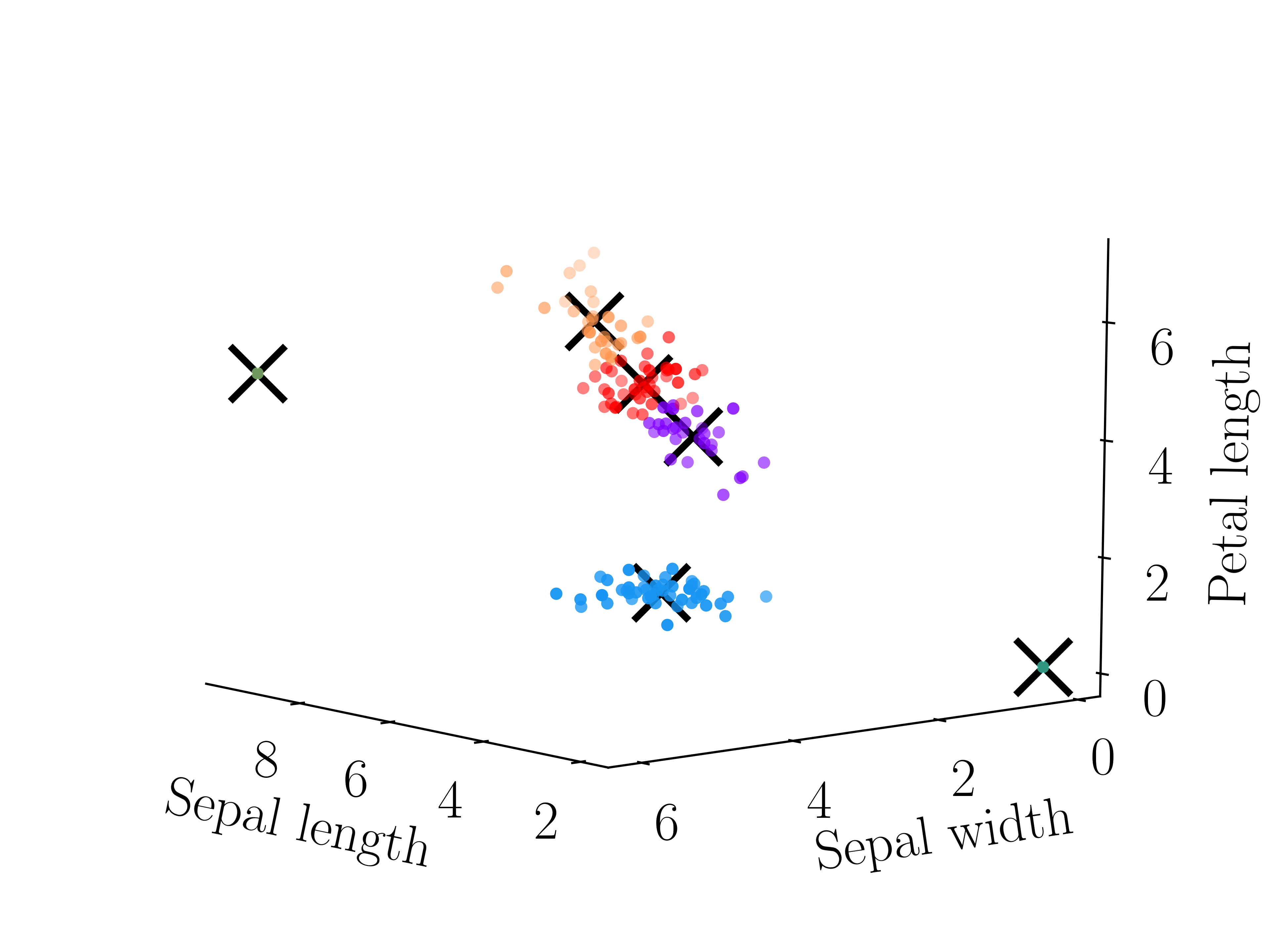}
        \caption{Outlier 2}
    \end{subfigure} \\
    \begin{subfigure}[t]{0.35\textwidth}
        \centering
        \includegraphics[width=1.00\textwidth]{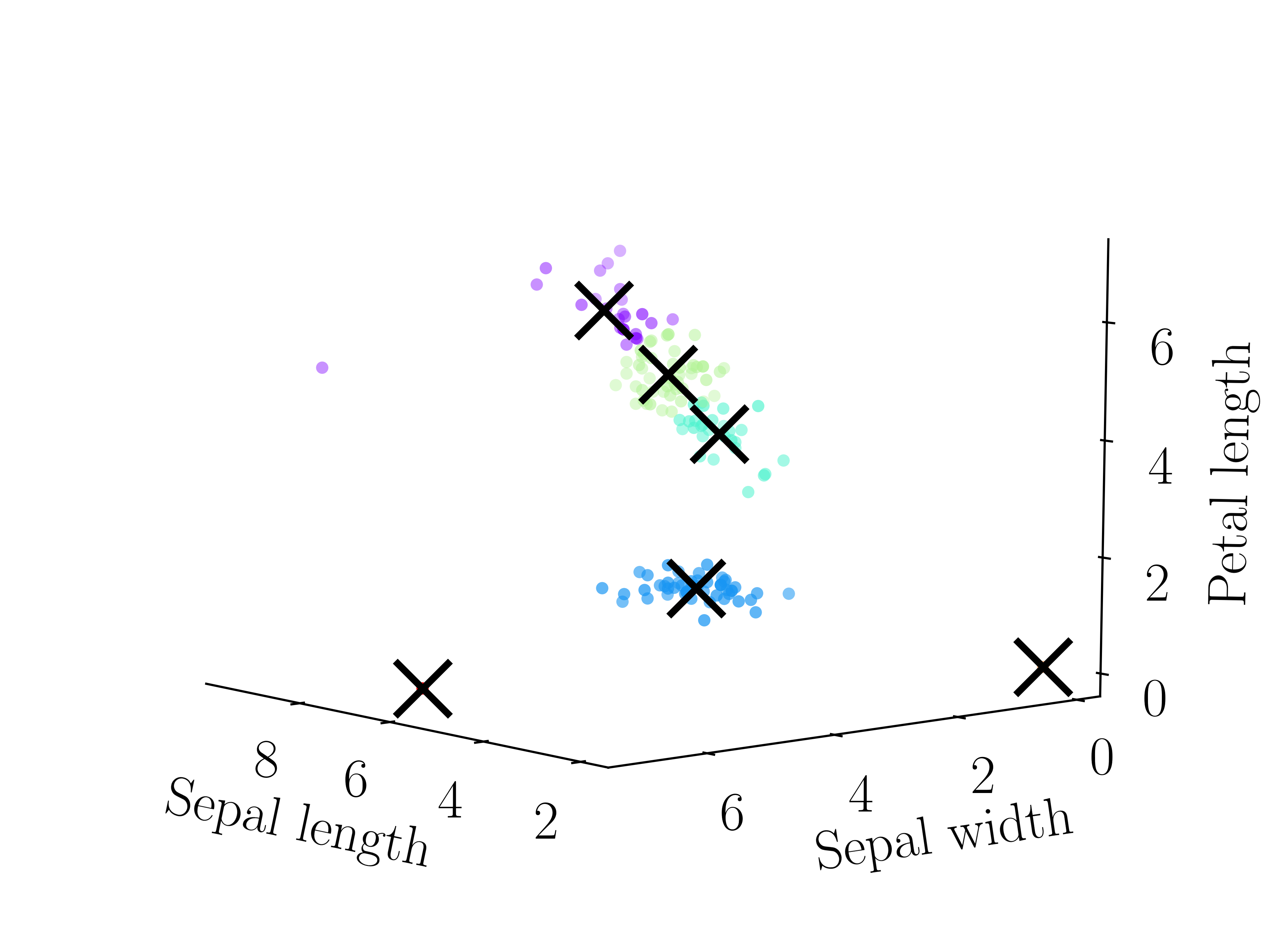}
        \caption{Outlier 3}
    \end{subfigure}%
    \begin{subfigure}[t]{0.35\textwidth}
        \centering
        \includegraphics[width=1.00\textwidth]{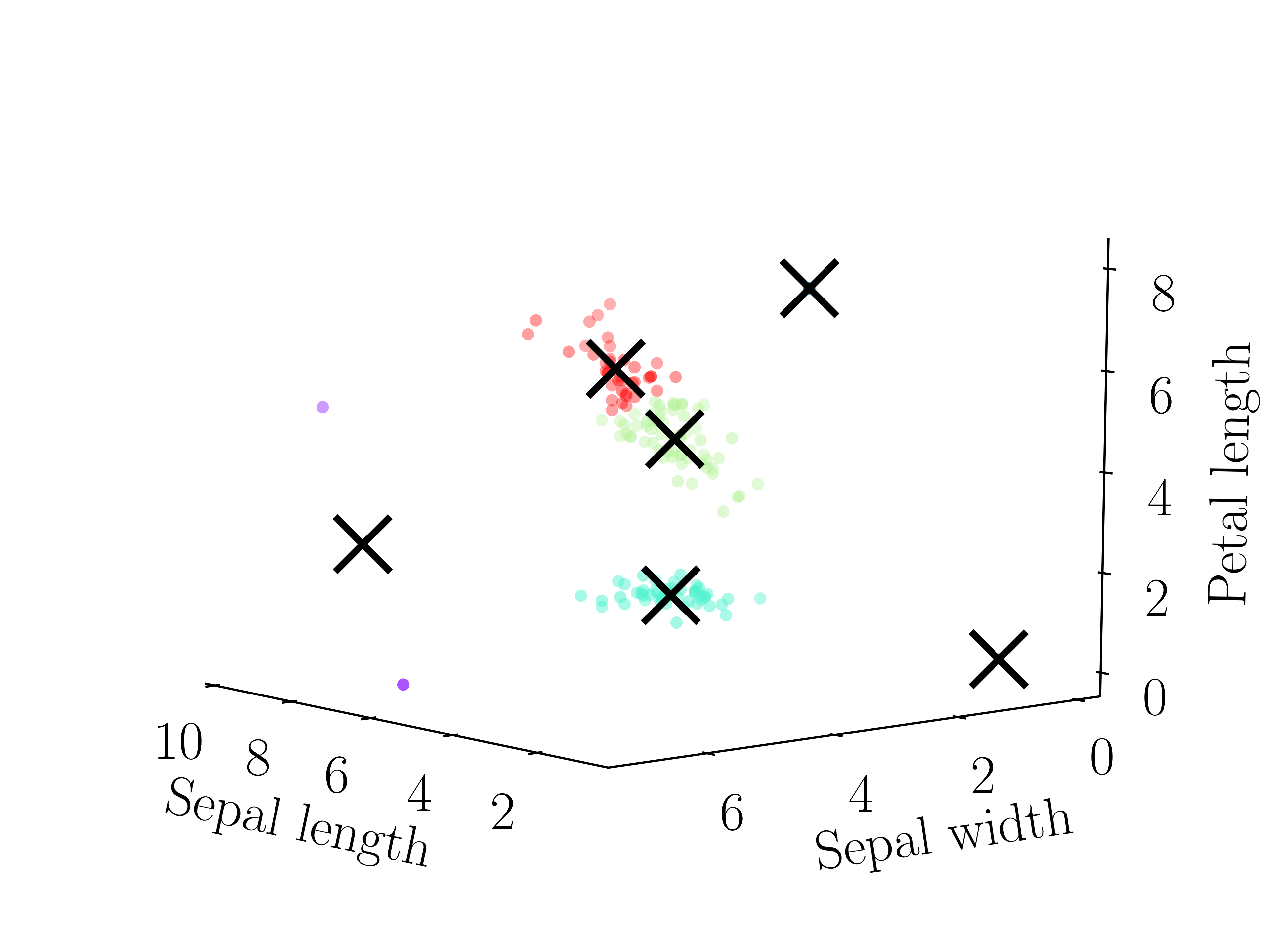}
        \caption{Outlier 4}
    \end{subfigure}%
    \caption{Global minima for the \textit{Iris} landscapes with $K=6$, and a
growing number of outliers. Minima are presented in three of the four features
to allow visualisation, given in centimetres. Petal width is excluded as it
contains the least information. The colours denote assignment to the
cluster centres, which are denoted by black crosses.}
    \label{OutliersIris6Minima}
\end{figure}

\begin{table}[h!]
\caption{The accuracy of clustering solutions in the glass landscapes. The
highest achievable accuracy is given, along with the cost function gap between
this clustering and the global minimum, and its structure type. The accuracy of
the global minimum (GM) is also given, along with its structure type.}
\label{OutlierGlassAccuracy}
\centering
\begin{tabularx}{\textwidth}{Y|Y|Y|Y|Y|Y}
      &   Best accuracy & $\Delta J$ & Type & Accuracy (G) & Type (GM) \\ \hline
Original  & 0.313  & 126.80 & - & 0.255 & -   \\
Outlier 1 & 0.303  & 24.04 / 180.66 & 1 / 0 & 0.252 & 1 \\
Outlier 2 & 0.303  & 595.26 & 1 & 0.252 & 2 \\
Outlier 3 & 0.299  & 811.15 & 3 & 0.226 & 4 \\
\end{tabularx}
\end{table}

\clearpage

\section{Rate-based clustering comparison}

\begin{figure}
    \centering
    \includegraphics[width=1.00\textwidth]{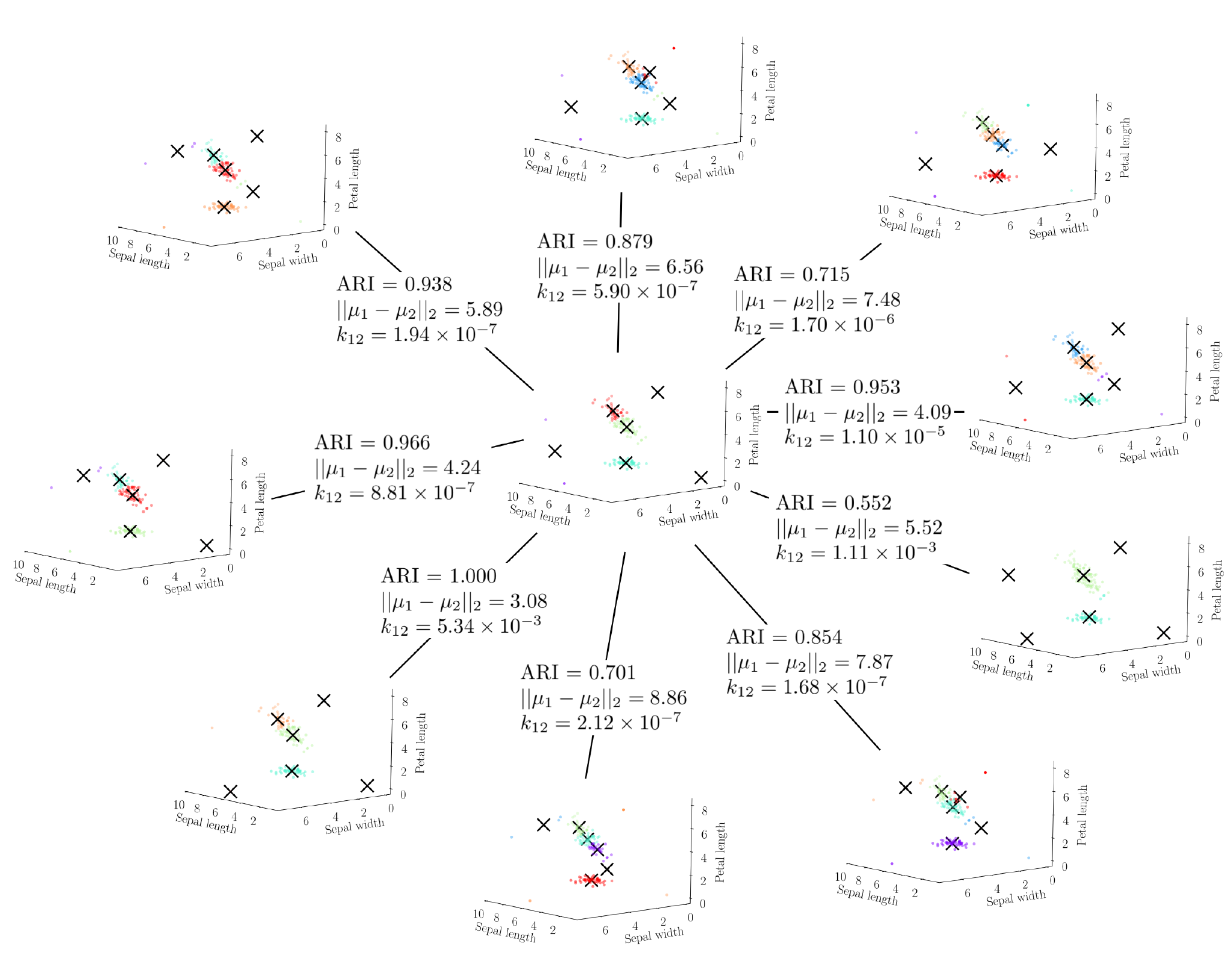}
    \caption{The different structure types of clustering solution for the
\textit{Iris} dataset with four outliers. The value of the cluster comparison
metrics are given for each minimum to the global minimum in the centre.}
    \label{ComparativeClusterings}
\end{figure}

\begin{table}[]
\caption{Transition rates (reduced units) between the set of minima with a given structure
type to the global minimum for the glass dataset, and variations with
additional outliers. Rates are calculated at a fictitious temperature $T=3.0$
in reduced units.
The dot indicates the structure type of the global minimum.}
\label{OutliersGlassRates}
\centering
\begin{tabularx}{\textwidth}{Y|Y|Y|Y}
Type      &   Outlier 1 & Outlier 2 & Outlier 3 \\ \hline
0 & $2.54 \times 10^{-4}$ & $8.13 \times 10^{-7}$ & $2.17 \times 10^{-10}$ \\
1 & $\cdot$ & $5.45 \times 10^{-7}$ & $2.45 \times 10^{-10}$ \\
2 &  & $\cdot$ & $4.84 \times 10^{-10}$ \\
3 &  &  & $2.92 \times 10^{-11}$ \\
4 &  &  & $\cdot$ \\
5 &  &  & $9.41 \times 10^{-11}$ \\
\end{tabularx}
\end{table}

\begin{table}[]
\caption{Transition rates (reduced units) from minima at the bottom of kinetic traps to the
global minimum for various $K$-means landscapes of the glass dataset. The rates
are calculated for selected kinetic traps, which are shown in
Fig.~S\ref{OutliersGlassKTDGs}, at a fictitious temperature of $T=3.0$. The
geometric mean rate for each landscape is given in the final row.}
\label{OutliersEscapeRatesGlass}
\centering
\begin{tabularx}{\textwidth}{Y|Y|Y|Y|Y}
Kinetic trap & Original     &   Outlier 1 & Outlier 2 & Outlier 3 \\ \hline
1 & $1.77 \times 10^{-5}$ & $7.83 \times 10^{-4}$ & $2.49 \times 10^{-6}$ & $1.30 \times 10^{-6}$ \\
2 & $8.25 \times 10^{-6}$ & $4.86 \times 10^{-4}$ & $3.55 \times 10^{-6}$ & $4.84 \times 10^{-10}$ \\ \hline
Mean & $1.21 \times 10^{-5}$ & $6.17 \times 10^{-4}$ & $2.97 \times 10^{-6}$ & $2.51 \times 10^{-8}$ \\
\end{tabularx}
\end{table}

\clearpage

\bibliographystyle{unsrt}
\bibliography{./Outliers_References.bib}